\documentclass[10pt,twocolumn,letterpaper]{article}
\usepackage[pagenumbers]{cvpr}
\usepackage[T1]{fontenc}
\usepackage{multirow}
\newcommand{\method}{\textsc{STAGE}}

\DeclareMathOperator*{\argmin}{arg\,min}
\DeclareMathOperator*{\argmax}{arg\,max}
\DeclareMathOperator{\diag}{diag}
\DeclareMathOperator{\med}{med}

\usepackage{placeins}

\definecolor{cvprblue}{rgb}{0.21,0.49,0.74}
\usepackage[pagebackref,breaklinks,colorlinks,allcolors=cvprblue,hypertexnames=false]{hyperref}
\title{STAGE: Structured Test-Time Adaptation with Global Entropic Soft-coupling for Cross-Subject Brain-to-Image Retrieval}
\author{%
    \large
    Yihang Zou$^{1}$\thanks{Equal contribution.} \quad
    Ruixi Liang$^{1}$\footnotemark[1] \quad
    Chunmei Qing$^{1}$\thanks{Corresponding author.} \quad
    Junpeng Tan$^{2}$ \quad
    Haichuan Wei$^{1}$\\[3pt]
    \large
    Wanxiang Luo$^{3}$ \quad
    Gao Wu$^{1}$ \quad
    Haodong Li$^{4}$ \quad
    Xiangmin Xu$^{1}$ \quad
    Yefeng Zheng$^{2}$\\[6pt]
    \normalsize $^{1}$South China University of Technology \quad
    $^{2}$Westlake University\\
    \normalsize $^{3}$Harbin Institute of Technology \quad
    $^{4}$The Chinese University of Hong Kong
}
\hypersetup{
    pdftitle={STAGE: Structured Test-Time Adaptation with Global Entropic Soft-coupling for Cross-Subject Brain-to-Image Retrieval},
    pdfauthor={Yihang Zou, Ruixi Liang, Chunmei Qing, Junpeng Tan, Haichuan Wei, Wanxiang Luo, Gao Wu, Haodong Li, Xiangmin Xu, Yefeng Zheng},
    pdfsubject={Cross-subject brain-to-image retrieval},
    pdfkeywords={brain-to-image retrieval, test-time adaptation, optimal transport}
}

\begin{document}
\maketitle
\begin{abstract}
Cross-subject brain-to-image retrieval aims to identify viewed images from unseen subjects' brain signals. However, inter-subject variability hinders alignment of brain--image representations, while independent query ranking tends to concentrate predictions on a few candidates, making accurate retrieval difficult without target labels.
We propose STAGE (Structured Test-Time Adaptation with Global Entropic Soft-coupling), a structured test-time adaptation framework for recovering cross-subject alignment from uncertain brain--image correspondences. Moment calibration reduces statistical discrepancies between features, while coupling-weighted initialization establishes initial alignment. Coupling-guided adaptation further refines residual feature mismatches, and soft-coupling ranking coordinates candidate competition across queries to prevent overly concentrated matching assignments. Using only unlabeled target batches, STAGE maintains consistent matching from feature calibration to image retrieval while keeping both encoders frozen.
Experiments on three public datasets validate STAGE's retrieval performance. On THINGS-EEG2, STAGE achieves 68.25\% Top-1 and 90.47\% Top-5, improving Top-1 by 18.52 percentage points over SCORE recovery and CSLS on identical frozen NICE features. Calibrated on 200 queries and a 200-image gallery covering their identities, STAGE retains 42.05\% Top-1 when ranking 1,854 images with the transform frozen.
\end{abstract}

\begin{figure}[t]
\centering
\includegraphics[width=\linewidth]{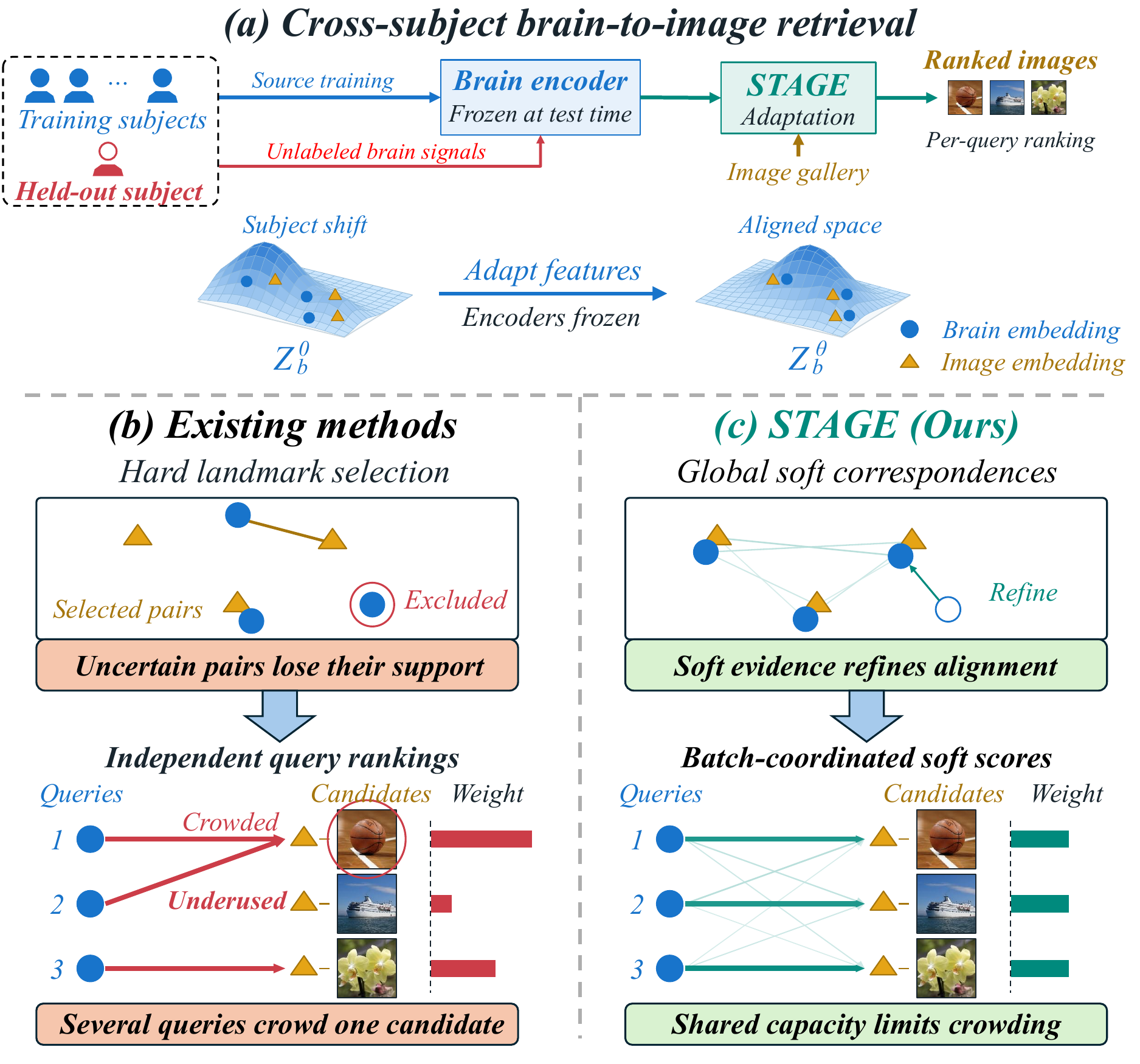}
\caption{\textbf{Cross-subject retrieval with global soft correspondences.} (a) A brain encoder trained on remaining subjects retrieves images for a held-out subject using unlabeled brain signals and a gallery, with both encoders frozen. (b) Hard landmark selection discards uncertain pairs, while independent ranking crowds a few candidates. (c) \method{} preserves soft correspondences for alignment and coordinates retrieval via shared candidate capacities. Blue circles and gold triangles denote brain and image embeddings; link widths and bars illustrate pairwise support and candidate weights.}
\label{fig:teaser}
\end{figure}

\section{Introduction}
\label{sec:introduction}

Brain-to-image retrieval aims to identify viewed images from non-invasive brain signals, including electroencephalography (EEG), magnetoencephalography (MEG), and functional magnetic resonance imaging (fMRI), supporting visual neuroscience and brain--computer interfaces~\cite{kay2008identifying,horikawa2017generic,du2023bravl,song2024nice}. Related work reconstructs images from EEG~\cite{kavasidis2017brain2image,bai2024dreamdiffusion} and fMRI~\cite{miyawaki2009mindeye,shen2019deep}, with diffusion models providing strong image priors~\cite{chen2023mindvis,takagi2023high,ozcelik2023brain}. Existing contrastive learning methods align brain signals with pretrained image representations, enabling retrieval of unseen concepts and image reconstruction~\cite{song2024nice,jiang2026samga,atm2024guided,scotti2023mindeye}. However, subject-specific models rely on extensive paired data, limiting their scalability. To reduce calibration costs for new users, recent studies have explored cross-subject transfer~\cite{wang2024mindbridge,scotti2024mindeye2} and test-time adaptation, which adapts pretrained models using unlabeled target data~\cite{wang2021tent}. Further efforts have investigated adaptation strategies based on embeddings and retrieval scores~\cite{huang2026sattc,cui2026score}. Nevertheless, inter-subject distribution shifts hinder brain--image representation alignment, while independent query ranking tends to concentrate retrieval results on a small subset of candidates.

Fig.~\ref{fig:teaser}(a) illustrates the leave-one-subject-out (LOSO)
evaluation setting, where a brain encoder trained on the remaining subjects
retrieves images for a held-out subject. During test-time adaptation only
unlabeled target batches and the image gallery are available; both encoders
remain frozen, and source data are inaccessible. Robust cross-subject
representation alignment and global retrieval without source data or target
labels is therefore a key challenge.

Inter-subject distribution shifts increase the uncertainty of the initial
cross-modal correspondences~\cite{cui2026score}, and existing methods use
mutual nearest neighbor (MNN) matching to select landmarks~\cite{cui2026score,conneau2020csls}. As shown in the upper part of Fig.~\ref{fig:teaser}(b), MNN excludes non-mutual pairs that may become informative as alignment improves. When geometric estimation relies on a fixed initial landmark set, subsequent correspondence evidence also remains unused. Thus, preserving soft correspondence evidence across all pairs and iteratively updating both the correspondences and the geometric transform are important for reliable cross-modal alignment.

Independent query ranking concentrates multiple queries on a few candidates, leading to hubness~\cite{radovanovic2010hubness}; as shown in the lower part of Fig.~\ref{fig:teaser}(b), this leaves other candidates underutilized. Density correction adjusts scores without constraining total candidate mass~\cite{conneau2020csls,huang2026sattc,cui2026score}, and query-aware Sinkhorn normalization balances assignments using uniform marginals~\cite{pan2025dbsn} but may over-allocate weights to unsupported candidates in expanded galleries. Thus, retrieval requires global query coordination and adaptive assignment based on batch--gallery composition.

To address these challenges, we propose STAGE (Structured Test-Time Adaptation with Global Entropic Soft-coupling), which uses global entropic soft-coupling (GES) to recover cross-subject alignment from uncertain correspondences and coordinate candidate competition. As outlined in Fig.~\ref{fig:teaser}(c), moment calibration first aligns feature means and scales. Coupling-weighted initialization then alternates soft correspondence estimation and geometric alignment, incorporating initially excluded pairs as alignment improves. Building on this geometry, coupling-guided adaptation uses refreshed soft teacher targets to refine residual mismatch through a constrained transform while limiting feature drift. Finally, soft-coupling ranking coordinates candidate competition at fixed geometry, allowing weakly supported candidates to remain below capacity in expanded galleries. Both encoders and image embeddings remain frozen throughout. Our main
contributions are:
\begin{itemize}
\setlength{\itemsep}{2pt}\setlength{\parsep}{0pt}\setlength{\topsep}{2pt}
\item STAGE, a structured test-time adaptation framework using global
entropic soft-coupling to recover cross-subject alignment and coordinate
candidate competition, enabling unseen-category image retrieval without
target labels, encoder retraining, or source data access.
\item A coupling-guided alignment strategy that alternates soft correspondence
estimation and geometric recovery, then refines the recovered features with
refreshed soft teacher targets. Joint ranking coordinates candidate competition
to limit matching-mass concentration.
\item Evaluation on three EEG and MEG benchmarks with multiple encoders and
varying gallery sizes, retrieving images for unseen subjects from unlabeled
target data and frozen encoders.
\end{itemize}

\begin{figure*}[t]
\centering
\includegraphics[width=0.95\textwidth]{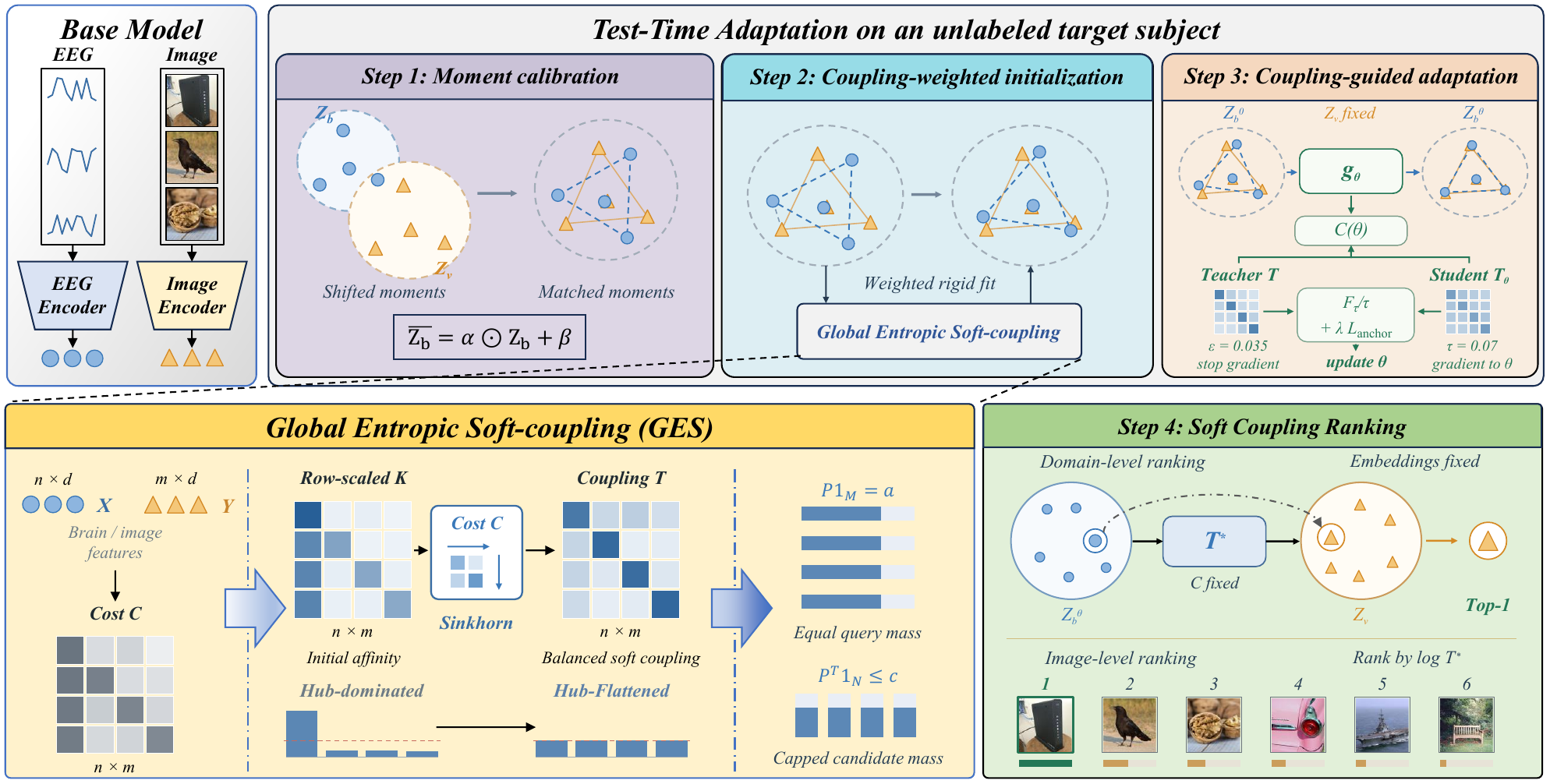}
\caption{\textbf{The \method{} test-time adaptation pipeline.}
Frozen encoders map an unlabeled target batch and the gallery to neural and
image embeddings. After moment calibration, global entropic soft-coupling (GES) supplies
initialization weights, teacher targets for feature adaptation, and final
ranking scores; only initialization and adaptation modify neural features.
The coupling-matching loss guides feature alignment, while the feature anchor
limits deviation from the initialized embeddings.}
\label{fig:architecture}
\end{figure*}

\section{Method}
\label{sec:method}

\subsection{Problem Setup}
\label{sec:setup}

At test time, cross-subject brain--image retrieval receives an unlabeled batch
of neural responses from one target subject, a known image gallery, and two
encoders trained on source subjects and not modified afterwards; the task is to
rank the gallery for every query of that subject. Let
$Z_b\in\mathbb{R}^{N\times d}$ be the brain embeddings of the $N$ queries of
that subject and $Z_v\in\mathbb{R}^{M\times d}$ the image embeddings of the
$M$ gallery candidates, where $d$ is the embedding dimension. The embeddings
$Z_b$ are re-estimated; the final score matrix $S\in\mathbb{R}^{N\times M}$
is computed from the re-estimated $Z_b$ and $Z_v$, its row $i$ giving the
ranks of the $M$ candidates for query $i$.

\subsection{Framework Overview}
\label{sec:pipeline}

\method{} recovers alignment from uncertain correspondences before
coordinating candidate competition, through four consecutive test-time
steps in Fig.~\ref{fig:architecture}. Moment calibration (Step~1) corrects
feature means and scales to make the two embedding sets comparable.
Coupling-weighted initialization (Step~2) alternates soft correspondence
estimation and rigid alignment, allowing pairs excluded by an initial
landmark selection to contribute as geometry improves.
Coupling-guided adaptation (Step~3) refines the initialized features
through a constrained similarity transform, using refreshed soft targets
to preserve correspondence uncertainty during refinement.
Soft-coupling ranking (Step~4) coordinates the final candidate scores
across queries at fixed geometry.
Steps 2--4 use global entropic soft-coupling (GES) for distinct purposes:
pair weights for recovery, teacher and student targets for refinement,
and joint ranking weights. Its query-mass and candidate-capacity
constraints carry batch-level competition through these steps.

\subsection{Moment Calibration}
\label{sec:moment}

Moment calibration removes the systematic difference between the two embedding
sets in per-coordinate mean and scale. Write $\mu(Z),\sigma(Z)\in\mathbb{R}^d$
for the per-coordinate mean and population standard deviation, and take
\begin{gather}
\alpha=\frac{\sigma(Z_v)+\varepsilon_m}{\sigma(Z_b)+\varepsilon_m}{,}\quad
\beta=\mu(Z_v)-\alpha\odot\mu(Z_b),\label{eq:momentalpha}\\
\bar Z_b=Z_b\odot\alpha+\beta .\label{eq:momentapply}
\end{gather}
Each coordinate of the neural embeddings is shifted and scaled into the
coordinate system of the gallery: $\alpha$ corrects the scale, $\beta$ the
mean, $\varepsilon_m>0$ prevents division by zero on low-variance coordinates,
and $\odot$ and division are elementwise with vectors broadcast over rows. The
gallery is kept as the reference, and $\bar Z_b$ is the input of geometry
alignment.

\subsection{The Global Entropic Soft-Coupling}
\label{sec:transport}

\paragraph{Feasible set.}
The matching must be represented by a matrix that lets the weights spread
over the candidates while capping the mass each candidate may receive. Let
$P\in\mathbb{R}_{+}^{N\times M}$ be a query--candidate weight matrix, $P_{ij}$ the weight
assigned by query $i$ to candidate $j$, and
$\mathbf1_n$ the all-ones vector of length $n$. The feasible set is
\begin{equation}
\Pi(a,c)=\left\{P\geq 0:\ P\mathbf1_M=a,\ P^\top\mathbf1_N\leq c\right\},
\label{eq:polytope}
\end{equation}
with $a=\frac{\mathbf1_N}{N}$ and $c=\frac{\mathbf1_M}{\min(N,M)}$. The row constraint fixes the total mass released by each query and the column
constraint caps the mass each candidate may receive. When $N=M$ the total
capacity $\sum_jc_j$ equals the total mass $\sum_ia_i$, so all column caps are
saturated and the column marginal is uniform; when $M>N$ each column may stay
below its cap.

\paragraph{Representation and cost.}
Differences in the norms of the raw embeddings would dominate the whole cost
matrix, so the pairwise cost first limits the role of norm differences and then
unifies the distance scale. Write $\widehat n_z=\max(\lVert z\rVert_2,\delta)$ for the floored norm,
where $\delta>0$ prevents division by zero. Let
$m_Z=\med_{z\in Z}\widehat n_z$ be the lower median of $\widehat n_z$ over
the set $Z$. The row-wise representation
\begin{equation}
\phi(z;m_Z)=\frac{z}{\widehat n_z}\,\kappa_r\exp\!\left[\kappa_p
\operatorname{clip}\!\left(\log\frac{\widehat n_z}{m_Z},-\kappa_c,\kappa_c\right)\right],
\label{eq:radial}
\end{equation}
keeps the direction $z/\widehat n_z$ and clips the log norm ratio to
$[-\kappa_c,\kappa_c]$, bounding the radius through gain $\kappa_p$
and base radius $\kappa_r$;
$\Phi(Z)$ applies it row-wise with a separate median per set. For the represented
query matrix $X$ and gallery matrix $Y$, the pairwise cost is
\begin{equation}
C_{ij}(X,Y)=\frac{\lVert x_i-y_j\rVert_2^2}
{\max\!\left(\operatorname{med}_{p,q}\lVert y_p-y_q\rVert_2^2,\delta_C\right)} ,
\label{eq:cost}
\end{equation}
where $x_i$ and $y_j$ are rows of $X$ and $Y$, $p,q$ run over all gallery
candidate pairs, and $\delta_C>0$ is the cost scale floor; the median distance
over all gallery pairs is the reference, so cost values from different batches
fall on the same scale.

\paragraph{Objective and coupling.}
The feasible set only fixes how the mass may be distributed; a criterion is
still needed to allocate it inside the set. Let
$H(P)=\sum_{i,j}P_{ij}(\log P_{ij}-1)$ with $0\log0=0$. For a cost $C$ and a
coefficient $\eta>0$, the scalar optimum is
$\mathrm{val}_\eta(C)=\min_{P\in\Pi(a,c)}\langle C,P\rangle+\eta H(P)$ and
the optimal coupling, i.e. the transport plan, is the matrix
$T_\eta(C)=\argmin_{P\in\Pi(a,c)}\langle C,P\rangle+\eta H(P)$, where
$\langle C,P\rangle=\sum_{i,j}C_{ij}P_{ij}$. The cost term prefers compatible
pairs and the negative entropy prefers spread-out soft weights; the larger $\eta$
is, the smoother the coupling. Initialization, the teacher, and ranking use $\eta=\varepsilon$;
the student uses $\eta=\tau>\varepsilon$.

\paragraph{Solver.}
The optimal coupling can be written as a product of a Gibbs kernel and positive
factors, so the constraints are satisfied by alternating scaling
iterations~\cite{cuturi2013sinkhorn}.
For $G_{ij}=\exp(-C_{ij}/\eta)$ the coupling has the form
$T=\diag(u)G\diag(\nu)$, with $u\in\mathbb{R}^N$ and
$\nu\in\mathbb{R}^M$ positive row and column factors, and capacity-aware
scaling alternates
\begin{equation}
u\leftarrow a\oslash(G\nu),\qquad
\nu\leftarrow\min\{\mathbf1_M,\ c\oslash(G^\top u)\},
\label{eq:sinkhorn}
\end{equation}
where $\oslash$ and $\min$ are elementwise: the first update pulls the row
mass of every query back to $a$, the second clips the received mass of
overloaded columns to $c$ and leaves the factor of an inactive column at one.

\subsection{Geometry Alignment}
\label{sec:align}

Geometry alignment removes the coordinate misalignment that survives
calibration. Initialization estimates a rigid transform over all pairs, and
adaptation learns a constrained similarity transform on top of that result.

\paragraph{Coupling-weighted initialization.}
Initialization estimates an orthogonal map and centroid translation
without committing to an initial set of landmark pairs.
Alternating transport and orthogonal fitting is an established alignment
strategy~\cite{grave2019wasserstein}. Here, all-pair coupling weights
retain alternative matches, query-confidence weights reduce the influence
of diffuse rows, and an identity penalty limits unstable rotations.
Each round recomputes correspondences from the current geometry and
refits the transform from the calibrated features.

Let $Q^{(0)}=\bar Z_b$ be the brain-side features at round~0. At round
$k=0,\ldots,K-1$, the coupling of the current features is computed first
and turned into query weights,
\begin{gather}
T^{(k)}=T_\varepsilon\!\left(C\!\left(\Phi(Q^{(k)}),\Phi(Z_v)\right)\right),\label{eq:initplan}\\
w_i=\frac{\gamma_0+\gamma_1\max\!\left(0,1-\frac{e_i}{\log M}\right)}
{\sum_{i'}\left(\gamma_0+\gamma_1\max\!\left(0,1-\frac{e_{i'}}{\log M}\right)\right)},\label{eq:initweights}
\end{gather}
where $T_\varepsilon(\cdot)$ is the optimal coupling at entropy
coefficient $\varepsilon$, $T^{(k)}_{ij}$ is the weight of the pair (query
$i$, candidate $j$) over all $N\times M$ pairs, and
$e_i=-\sum_jp_{ij}\log p_{ij}$ is the entropy of the row
$p_{ij}=T^{(k)}_{ij}/\sum_{j'}T^{(k)}_{ij'}$ of the coupling, so that a small
$e_i$ means that the matches of that query concentrate on few candidates. Here
$\gamma_0>0$ is the weight floor, $\gamma_1\geq0$ the gain, and $\log M$ the
entropy of the uniform distribution over $M$ elements, so that every query
enters the fit with a non-zero weight.

Multiplying the weights back into the coupling gives the confidence-weighted pair matrix
$A^{(k)}_{ij}=w_iT^{(k)}_{ij}$ and the total mass
$\omega=\sum_{i,j}A^{(k)}_{ij}$; the weighted centroids of the two sides, the
centered features and their cross moment are
\begin{gather}
x_c=\frac{\bar Z_b^\top A^{(k)}\mathbf{1}_M}{\omega},
\quad
y_c=\frac{Z_v^\top(A^{(k)})^\top\mathbf{1}_N}{\omega},
\label{eq:initcentroids}\\
X_c=\bar Z_b-\mathbf{1}_N x_c^\top,
\quad
Y_c=Z_v-\mathbf{1}_M y_c^\top,
\label{eq:initcentered}\\
D^{(k)}=X_c^\top A^{(k)}Y_c
\in\mathbb{R}^{d\times d},
\label{eq:initcov}\\
R^{(k)}=\operatorname*{arg\,max}_{R^\top R=I_d}
\bigl\{\operatorname{tr}(R^\top D^{(k)})
-\frac{\rho}{2}\lVert D^{(k)}\rVert_2
\lVert R-I_d\rVert_F^2\bigr\},
\label{eq:initrigid}
\end{gather}
where $A^{(k)}$ carries the weight of every pair and $D^{(k)}$ gathers its
directional information. The rigid transform $R^{(k)}$ solves a regularized, weighted orthogonal
Procrustes problem~\cite{kabsch1998}, with pair weights from the coupling
instead of a selected correspondence set;
its solution is obtained in closed form through a singular value decomposition,
and the ridge coefficient is set to $\rho\lVert D^{(k)}\rVert_2$, so that its
strength is independent of the scale of the weights.

At the end of each round the transform is applied to the calibrated features,
\begin{gather}
Q^{(k+1)}=X_cR^{(k)}+\mathbf{1}_N y_c^\top,
\label{eq:initapply}\\
Z_b^0=\Phi\!\bigl(Q^{(K)}\bigr){,}
\quad
Z_v^0=\Phi(Z_v).
\label{eq:initout}
\end{gather}
Each round updates the correspondences using the refined geometry;
$K$ controls the number of alternating rounds.
The outputs $Z_b^0$ and $Z_v^0$ are the represented neural and gallery sides
that form the input of adaptation.

\paragraph{Coupling-guided adaptation.}
Adaptation refines the remaining coordinate mismatch after initialization.
The initialized geometry supplies soft targets, while a constrained
similarity transform and feature anchor limit the displacement induced
by uncertain target matches. Only this transform is optimized; the
encoders and image features remain fixed. Take the first $r$ right
singular vectors of $[Z_b^0;Z_v^0]$ as a fixed orthonormal basis
$B\in\mathbb{R}^{d\times r}$, and let the neural subspace coordinates be
$E_0=Z_b^0B\in\mathbb{R}^{N\times r}$. The parameters are
$\theta=(W,\ell,h)$, with $W\in\mathbb{R}^{r\times r}$,
$\ell\in\mathbb{R}$ and $h\in\mathbb{R}^d$, and the transform has three
parts,
\begin{align}
R_\theta&=\exp(W-W^\top),\label{eq:rot}\\
s_\theta&=\exp(\sigma_s\tanh\ell),\label{eq:scale}\\
t_\theta&=\sigma_t\tanh h,\label{eq:shift}
\end{align}
where $R_\theta=\exp(W-W^\top)$ gives an orthogonal rotation inside the
subspace, $s_\theta$ is a bounded shared scale and $t_\theta$ a bounded
translation, with $\sigma_s$ and $\sigma_t$ bounding their ranges; the
feature map is
\begin{equation}
Z_b^\theta=s_\theta\left[Z_b^0-E_0B^\top+E_0R_\theta B^\top\right]
+\mathbf1_Nt_\theta^\top .
\label{eq:featmap}
\end{equation}
Only subspace coordinates rotate; the complement is preserved before uniform
scaling. Zero parameters give the identity map, without shear or anisotropic
stretch. $B$, image features and encoders stay fixed. Under fixed row and column
marginals, translation preserves the optimal coupling; the feature anchor
penalizes the full displacement.

Write $C_\theta=C(Z_b^\theta,Z_v^0)$. Every $K_{\mathrm{ref}}\geq1$ student
updates, recompute and detach the teacher coupling at the current reference
state $\theta_{\mathrm{ref}}$,
\begin{align}
T&=T_\varepsilon\!\left(C_{\theta_{\mathrm{ref}}}\right),\label{eq:teacher}\\
\widehat T&=\frac{T}{\sum_{i,j}T_{ij}},\label{eq:teachercopy}
\end{align}
where $T$ is the teacher coupling and $\widehat T$ its unit-mass copy, used
for numerical stability. Holding $\widehat T$ fixed between refreshes
provides a stable correspondence target for several feature updates;
refreshing it incorporates the resulting geometric changes.
The student recomputes $T_\theta=T_\tau(C_\theta)$ at the current
parameters, with $\tau>\varepsilon$ encouraging broader support.
Matching it to the sharper soft teacher refines alignment while retaining
alternative correspondences.

The Fenchel--Young gap measures the excess regularized cost of the
fixed teacher over the current optimal student coupling,
\begin{gather}
F_\tau=\langle C_\theta,\widehat T\rangle+\tau H(\widehat T)
-\mathrm{val}_\tau(C_\theta)\geq0,\label{eq:gap}\\
\frac{\partial(F_\tau/\tau)}{\partial C_\theta}
=\frac{\widehat T-T_\theta}{\tau},\label{eq:gapgrad}
\end{gather}
and the envelope gradient $(\widehat T-T_\theta)/\tau$ is backpropagated
through $C_\theta$ to $\theta$; neither the teacher nor the solver iterations
enter the differentiation path. With Adam we minimize
\begin{gather}
\mathcal{L}=\frac{F_\tau}{\tau}+\lambda L_{\mathrm{anchor}},\label{eq:loss}\\
L_{\mathrm{anchor}}=
\frac{\lVert Z_b^\theta-Z_b^0\rVert_F^2/N}
{\max\!\left(\lVert Z_b^0\rVert_F^2/N,\delta\right)},\label{eq:anchor}
\end{gather}
where $L_{\mathrm{anchor}}$ measures the displacement of the features relative
to $Z_b^0$ with the Frobenius norm and $\lambda$ controls the anchor strength. The final neural
features are $Z_b^\star$.

\subsection{Soft-Coupling Ranking}
\label{sec:ranking}

Aligned features can still make several queries prefer the same
candidate. Soft-coupling ranking uses competition across queries to
redistribute matching mass at fixed geometry. It fixes the final features
and the represented gallery, $Y=Z_v^0$ in the closed set and the extended
gallery after the same representation for gallery expansion, computes
$C^\star=C(Z_b^\star,Y)$ and $T^\star=T_\varepsilon(C^\star)$, and obtains
the final score matrix $S$, whose element $S_{ij}$ is the score of query $i$
for candidate $j$,
\begin{equation}
S_{ij}=\log T^\star_{ij}.
\label{eq:ranking}
\end{equation}

\section{Experiments}
\label{sec:experiments}

\subsection{Setup}
\label{sec:exp-setup}

\paragraph{Protocol.}
We evaluate 200-way brain-to-image retrieval on
THINGS-EEG2~\cite{gifford2022things}, THINGS-MEG from
THINGS-data~\cite{hebart2023things} and Alljoined-1.6M~\cite{xu2025alljoined},
using their released splits and the standard leave-one-subject-out protocol:
each fold holds out one subject, the base is trained on the remaining
subjects, and at test time the pipeline adapts to the held-out subject from
unlabelled responses and the known candidate set, with both encoders frozen.
THINGS-EEG2's ten subjects, THINGS-MEG's four held-out subjects and
Alljoined-1.6M's twenty held-out subjects each run three seeds
(seed 0/1/2). Top-1 is primary and Top-5 secondary.
We average seeds within subjects, then across subjects.
We jointly fit and rank closed-set batches with one query per candidate identity;
preprocessing and trial averaging are detailed in Supplementary Section~A.

\paragraph{Training and deployment.}
After concept-disjoint epoch selection, we retrain on all source training
data and report the final checkpoint. Two EEG encoders are used, NICE and
TSConv~\cite{song2024nice}, with the training protocol of prior
work~\cite{jiang2026samga}; the image side uses frozen features from
InternViT-6B-448px-V2.5~\cite{chen2024internvl25}, and the brain side is
projected to $d=512$. A target batch is processed by the four steps of
Section~\ref{sec:pipeline}; the final score is the coupling ranking, which
orders candidates by $\log T^\star_{ij}$, with $N=M=200$.
The reported initialization/adaptation budgets are fixed at 12/40 on THINGS-EEG2, 2/8 on
THINGS-MEG and 5/13 on Alljoined-1.6M, shared across subjects and source
seeds within each dataset. The supplementary material defines a source-only
protocol for fold-specific budget selection and lists the remaining hyperparameters.

\subsection{Main results}
\label{sec:main-results}

\begin{table*}[t]
\centering
\small
\caption{\textbf{Cross-subject retrieval accuracy (\%).} Above the divider, EEG2
and Alljoined use published SAMGA-based baselines from
SCORE~\cite{cui2026score}; MEG uses four held-out subjects. Below it,
SCORE recovery + CSLS and STAGE share frozen NICE features, queries,
galleries and source seeds (0/1/2). Subject-paired intervals are reported
in Supplementary Table~S40.}
\label{tab:main}
\begin{tabular*}{\textwidth}{@{\extracolsep{\fill}}lrrrrrr@{}}
\toprule
\multirow{2}{*}{Method}
& \multicolumn{2}{c}{THINGS-EEG2~\cite{gifford2022things}}
& \multicolumn{2}{c}{THINGS-MEG~\cite{hebart2023things}}
& \multicolumn{2}{c}{Alljoined-1.6M~\cite{xu2025alljoined}}\\
\cmidrule(lr){2-3}\cmidrule(lr){4-5}\cmidrule(lr){6-7}
& Top-1 & Top-5 & Top-1 & Top-5 & Top-1 & Top-5\\
\midrule
SAMGA base (Unadapted)~\cite{jiang2026samga} & 26.22 & 57.98 & 3.88 & 14.00 & 6.08 & 20.55\\
EA~\cite{he2020euclidean} & 27.45 & 59.37 & 4.00 & 15.88 & 6.31 & 20.47\\
SAW~\cite{huang2026sattc} & 30.98 & 60.85 & 4.50 & 14.88 & 8.93 & 27.54\\
SATTC~\cite{huang2026sattc} & 26.12 & 52.73 & 5.13 & 13.38 & 6.78 & 20.70\\
CSLS~\cite{conneau2020csls} & 35.78 & 67.85 & 4.75 & 16.25 & 8.04 & 25.39\\
SCORE~\cite{cui2026score} & 53.23 & 83.55 & 5.13 & 17.88 & 12.01 & 32.16\\
\midrule
SCORE recovery + CSLS (NICE)~\cite{cui2026score} & 49.73 & 81.15 & 5.00 & 17.67 & 10.18 & 30.45\\
\method{} (ours) & \textbf{68.25} & \textbf{90.47} & \textbf{6.25} & \textbf{20.17}
& \textbf{12.89} & \textbf{34.74}\\
\bottomrule
\end{tabular*}

\end{table*}

On identical frozen NICE features, STAGE improves Top-1 over SCORE
recovery plus CSLS by 18.52 percentage points on EEG2, 1.25 on MEG and 2.72 on
Alljoined. With the same final Sinkhorn ranking, STAGE initialization improves
EEG2 Top-1 from 58.20\% to 65.68\%, and adaptation further increases it to
68.25\% (Supplementary Table~S39).

\begin{figure}[t]
\centering
\includegraphics[width=\columnwidth]{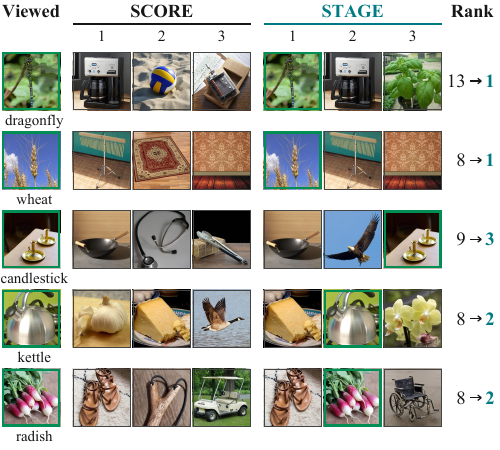}
\caption{\textbf{Qualitative retrieval comparison with SCORE.} THINGS-EEG2, subject 01, seed 0,
200-image gallery. SCORE (reproduced coordinate recovery with CSLS) and
\method{} share one frozen base. Each shows its top three; green borders mark
the viewed image. Right: its SCORE-to-\method{} rank.}
\label{fig:examples}
\end{figure}

\paragraph{Retrieval examples.}
Fig.~\ref{fig:examples} illustrates how STAGE improves the true image's
rank relative to SCORE for five queries from one held-out subject.

\begin{figure}[t]
\centering
\includegraphics[width=\columnwidth]{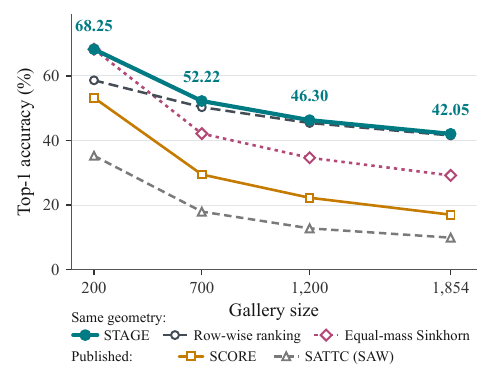}
\caption{\textbf{Retrieval performance with expanded galleries.}
Top-1 over 30 subject--seed cases. Geometry is fitted without pair labels
on 200 queries and 200 candidates covering their identities, then frozen
for gallery expansion.
Coupling, row-wise, and equal-mass Sinkhorn rankings share the fitted
geometry. SCORE and SATTC curves use published values; full numerical
results are reported in Supplementary Table~S28.}
\label{fig:gallery}
\end{figure}

\paragraph{When the gallery outgrows the batch.}
Following SCORE~\cite{cui2026score}, we fit 200 queries to a 200-image
gallery covering their identities without pair labels, then freeze
geometry and add ranking distractors (Fig.~\ref{fig:gallery}).
The supplement separately tests fitting with the full expanded gallery.

\paragraph{Swapping the EEG encoder.}
Table~\ref{tab:encoder} gives the Top-1 of each stage of the pipeline on four
EEG encoders. Across these encoder-specific evaluations,
the adaptation and ranking stages raise Top-1 to about three times its former value on
every encoder ($3.0$--$4.0\times$); the chance level of a 200-way retrieval is
0.5\%.

\begin{table}[t]
\centering
\small
\caption{\textbf{Retrieval performance across EEG encoders.} Top-1 (\%) of each stage of the pipeline on
four EEG encoders.
}
\label{tab:encoder}
\begin{tabular*}{\columnwidth}{@{\extracolsep{\fill}}lcccc@{}}
\toprule
Stage & \shortstack{SAMGA-\\TSConv\\\cite{jiang2026samga,song2024nice}}
& \shortstack{EEGNet\\\cite{lawhern2018eegnet}}
& \shortstack{Shallow-\\FBCSP\\\cite{schirrmeister2017deep}}
& \shortstack{UBP\\\cite{wu2025ubp}}\\
\midrule
None (raw) & 18.87 & 16.65 & 8.22 & 20.60\\
Calibration & 44.25 & 29.70 & 17.43 & 36.75\\
Init.+adapt. & 56.98 & 47.50 & 27.20 & 51.05\\
Ranking & \textbf{66.30} & \textbf{56.25} & \textbf{33.10} & \textbf{61.65}\\
\bottomrule
\end{tabular*}

\end{table}

\subsection{Ablations}
\label{sec:ablation}

\paragraph{Progressive ablation.}
Table~\ref{tab:progressive} traces feature alignment and ranking:
initialization and adaptation improve row-wise Top-1 across datasets,
and coupling ranking adds gains at the resulting geometry.

\begin{table*}[t]
\centering
\small
\caption{\textbf{Progressive component ablation}
(Top-1/Top-5, \%). The unadapted baseline and subsequent feature stages
use row-wise radial-distance ranking; the last
row switches to coupling ranking.}
\label{tab:progressive}
\begin{tabular*}{\textwidth}{@{\extracolsep{\fill}}ccccrrrrrr@{}}
\toprule
\multirow{2}{*}{Moment} & \multirow{2}{*}{Coupled init.} & \multirow{2}{*}{Coupled adapt.}
& \multirow{2}{*}{Coupling ranking}
& \multicolumn{2}{c}{THINGS-EEG2} & \multicolumn{2}{c}{THINGS-MEG}
& \multicolumn{2}{c}{Alljoined-1.6M}\\
\cmidrule(lr){5-6}\cmidrule(lr){7-8}\cmidrule(lr){9-10}
& & & & Top-1 & Top-5 & Top-1 & Top-5 & Top-1 & Top-5\\
\midrule
\(\times\) & \(\times\) & \(\times\) & \(\times\)
& 26.80 & 58.45 & 3.13 & 14.00 & 6.20 & 20.60\\
\(\checkmark\) & \(\times\) & \(\times\) & \(\times\)
& 30.80 & 63.10 & 3.50 & 13.71 & 8.23 & 25.50\\
\(\checkmark\) & \(\checkmark\) & \(\times\) & \(\times\)
& 53.03 & 81.90 & 5.00 & 17.42 & 9.95 & 29.92\\
\(\checkmark\) & \(\checkmark\) & \(\checkmark\) & \(\times\)
& 58.63 & 85.18 & 5.17 & 17.13 & 10.28 & 30.03\\
\midrule
\(\checkmark\) & \(\checkmark\) & \(\checkmark\) & \(\checkmark\)
& \textbf{68.25} & \textbf{90.47} & \textbf{6.25} & \textbf{20.17} & \textbf{12.89} & \textbf{34.74}\\
\bottomrule
\end{tabular*}

\end{table*}

\paragraph{Ranking at fixed features.}
Within each row of Table~\ref{tab:singlestep}, both rankings share features
and costs. Initialization and adaptation improve both rankings; coupling
ranking further improves Top-1 at every stage. Feature refinement and
joint ranking thus make complementary contributions.

\begin{table}[t]
\centering
\small
\setlength{\tabcolsep}{4pt}
\caption{\textbf{Ranking at fixed features} (THINGS-EEG2, Top-1, 30 cases). Each
row adds one stage to the row above it; the two ranking columns share the
same features and the same cost matrix, so \(\Delta\) is the gain of the
coupling ranking over the row-wise ranking at that stage set.}
\label{tab:singlestep}
\begin{tabular*}{\columnwidth}{@{\extracolsep{\fill}}lrrr@{}}
\toprule
Enabled stages & Row-wise & Coupled & \(\Delta\)\\
\midrule
Moment calibration & 30.80 & 48.08 & +17.28\\
+ coupling-weighted init. & 53.03 & 65.68 & +12.65\\
+ coupling-guided adapt. & 58.63 & \textbf{68.25} & +9.62\\
\bottomrule
\end{tabular*}

\end{table}

\paragraph{Correspondences for coordinate recovery.}
SCORE~\cite{cui2026score} fits a rigid transform to landmark pairs.
Fig.~\ref{fig:weightgrid} compares MNN and dense support, weights and
rounds with the same regularized fit and coupling ranking.
Margin and coupling weights perform similarly and favor dense support at
both round counts; uniform dense weights dilute the fit with incorrect pairs.
With coupling weights, all-pair Top-1 rises from 57.83\% at one round to
65.68\% at $K=12$, versus 63.55\% for MNN under the same weighting.
With uniform query multipliers and matched budgets, estimator and final
SN ranking, soft recovery improves EEG2 Top-1 over iterative CSLS MNN
by 2.27 points without adaptation and 1.50 with identical global
adaptation (Table~\ref{tab:matched_recovery}).
Among queries whose true pairs are excluded by the common initial CSLS MNN
selection, soft recovery reaches 52.83\% Top-1 after twelve rounds, versus
48.83\% for iterative CSLS MNN (Supplementary Fig.~S11).
With the estimator, update budget and final ranking matched, soft
correspondences recover more initially excluded matches even when hard
correspondences are repeatedly updated.

\begin{table}[t]
\centering
\small
\caption{\textbf{Matched correspondence recovery.}
EEG2 Top-1 (\%), 30 subject--seed cases. Initializers share the
regularized estimator, twelve rounds, uniform query multipliers and
GES ranking; adaptation uses forty identical global updates.
Cross-dataset results appear in
Supplementary Table~S41.}
\label{tab:matched_recovery}
\begin{tabular*}{\columnwidth}{@{\extracolsep{\fill}}lrr@{}}
\toprule
Initialization & Init. only & Init. + adapt.\\
\midrule
Radial MNN & 58.40 & 63.07\\
CSLS MNN & 62.93 & 66.52\\
Soft correspondence & \textbf{65.20} & \textbf{68.02}\\
\bottomrule
\end{tabular*}
\end{table}

\subsection{Mechanism analysis}
\label{sec:mechanism}

\paragraph{A capacity bound against forced uniform mass.}
The gallery-expansion results test how joint ranking handles additional
candidates. Both rules use alternating scaling~\cite{cuturi2013sinkhorn},
but standard Sinkhorn requires equal column mass, whereas capacities allow
columns to remain below their bounds.
For $M>N$, capacities avoid forcing equal mass onto unsupported candidates,
explaining their advantage over equal-mass ranking in
Fig.~\ref{fig:gallery}.

\begin{figure}[!htbp]
\centering
\includegraphics[width=\columnwidth]{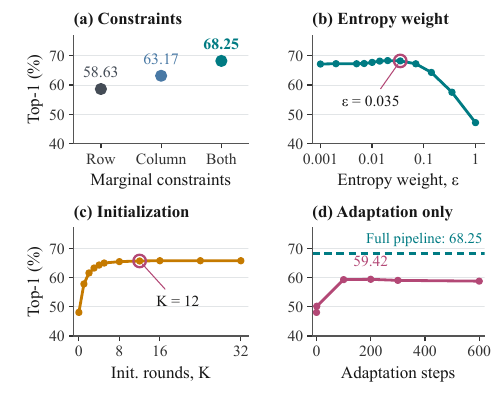}
\caption{\textbf{Effects of coupling constraints and optimization settings.} THINGS-EEG2 Top-1.
(a) Row-only, column-only and joint constraints at fixed calibrated and
aligned features. (b) Ranking entropy $\varepsilon$.
(c) Initialization rounds. (d) Adaptation steps without initialization.
Rings mark the adopted $\varepsilon=0.035$ and $K=12$;
the dashed line in (d) marks the full pipeline.}
\label{fig:mechanism}
\end{figure}

\paragraph{Candidate competition at fixed geometry.}
To explain the within-row gains in Table~\ref{tab:singlestep},
Fig.~\ref{fig:mechanism}(a) varies only the coupling constraints on one
frozen final cost matrix. The row constraint alone reproduces direct ranking
(58.63\% Top-1); the column constraint alone increases Top-1 to 63.17\%; with
both, the coupling reaches 68.25\%. Row constraints fix each query's total
mass; column constraints limit the mass accumulated by each candidate.
Their combination improves retrieval through coordinated candidate
competition at unchanged geometry.

\paragraph{Entropy keeps the coupling soft.}
Minimizing transport cost favors low-cost pairs, but low cost need not
indicate a correct match (Supplementary Table~S44). Entropy retains weight on
near-optimal alternatives; excessive smoothing spreads it across all pairs
and reduces accuracy (Fig.~\ref{fig:mechanism}(b)).
The adopted $\varepsilon=0.035$ gives 68.25\% Top-1, close to
68.37\% at $\varepsilon=0.02$, while excessive smoothing at
$\varepsilon=1$ reduces accuracy to 47.12\%.

\paragraph{Initialization prepares the geometry for adaptation.}
At $K=12$, initialization reaches 65.68\% Top-1 (plateau: 65.78\%).
Adaptation without it peaks at 59.42\%, versus 68.25\% overall
(Fig.~\ref{fig:mechanism}(c,d); Table~\ref{tab:progressive}),
supporting recovery before refinement.
At K12/40 with GES ranking, STAGE exceeds row-only teacher--student
coupling by 2.00 points (95\% interval [0.93, 3.05]).
Replacing only the soft teacher with a hard assignment reduces Top-1
from 68.25\% to 67.02\%, a paired gain of 1.23 points
([0.12, 2.32]) for retaining soft targets.
These controls isolate competition during feature updates and
teacher uncertainty.
Soft teacher targets give higher Top-1 than direct OT under matched settings
(Table~\ref{tab:adaptation_objectives}).
Entropic OT--Procrustes continuation reaches 68.08\%
(Supplementary Table~S43).

\begin{table}[!t]
\centering
\small
\caption{\textbf{Matched adaptation objectives} (Top-1, \%).
All other settings are shared (Supplementary Table~S45).}
\label{tab:adaptation_objectives}
\begin{tabular*}{\columnwidth}{@{\extracolsep{\fill}}lrrr@{}}
\toprule
Objective & EEG2 & MEG & Alljoined\\
\midrule
Direct OT (\(\eta=\varepsilon\)) & 61.48 & 5.29 & 10.93\\
Direct OT (\(\eta=\tau\)) & 52.53 & 5.58 & 10.43\\
STAGE (soft teacher) & \textbf{68.25} & \textbf{6.25} & \textbf{12.89}\\
\bottomrule
\end{tabular*}
\end{table}

\begin{figure}[t]
\centering
\includegraphics[width=\columnwidth]{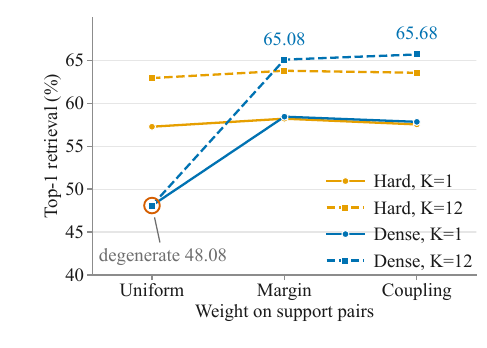}
\caption{\textbf{Effects of correspondence support, weights, and initialization rounds.}
THINGS-EEG2, 30 subject--seed cases, coupling ranking, no adaptation.
MNN and all-pair support are compared under uniform, margin and coupling
weights at $K=1$ and $K=12$. The hollow circle marks the degenerate
all-pair fit with uniform weights.}
\label{fig:weightgrid}
\end{figure}

Supplementary Fig.~S12 visualizes the support in
Fig.~\ref{fig:weightgrid}: MNN admits few pairs; the coupling retains
alternatives across all 200 candidates.

\begin{figure}[t]
\centering
\includegraphics[width=\columnwidth]{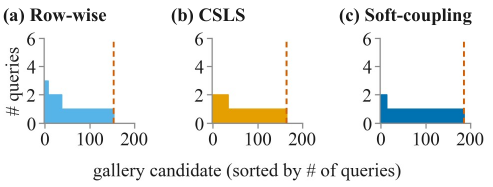}
\caption{\textbf{Distribution of top-1 predictions across candidates.}
The distributions compare (a) row-wise ranking, (b) CSLS and (c) coupling
ranking on one cost matrix. Candidates are sorted by their prediction
counts. Dashed line: gallery coverage;
leftmost bar: the most-cited candidate.
THINGS-EEG2, subject 01, seed 0.}
\label{fig:hub}
\end{figure}

\paragraph{The coupling ranking mitigates hubness.}
Candidate-level predictions illustrate the effect of joint ranking.
On one fixed cost matrix, 42.00\% of queries share their top-ranked candidate
with at least one other query. Row-wise ranking selects 154 distinct
candidates as Top-1, compared with 165 for CSLS and 186 for coupling
ranking (Fig.~\ref{fig:hub}).

\paragraph{Computation time.}
With four CPU threads and 200 queries/candidates, frozen-feature
calibration and ranking take 14.09/1.13/2.01\,s on EEG2/MEG/Alljoined,
excluding acquisition and encoder inference (Supplementary Table~S3).
EEG2 post-initialization fitting takes 6.91\,s, versus 39.22\,s for
confidence-weighted, regularized OT--Procrustes (Supplementary Table~S46).

\subsection{Transfer to Disjoint Queries}
\label{sec:boundary}

The frozen complete map reaches 75.47\% Top-1 on 100 disjoint identities
of the same subject, supporting reuse of calibrated geometry.
Ranking remains joint.

\section{Conclusion}
\label{sec:conclusion}

STAGE uses global entropic soft-coupling to recover cross-subject geometry
through alternating correspondence and transform estimation.
Constrained adaptation then refines residual mismatch, while
capacity-constrained ranking coordinates candidate competition.
With both encoders frozen and no target labels, STAGE reaches
68.25\%/90.47\% Top-1/Top-5 on THINGS-EEG2.
Calibrated on 200 queries and 200 candidates covering their identities,
it retains 42.05\% Top-1 when ranking 1,854 images with frozen geometry.
Together with matched-feature comparisons and disjoint-query transfer,
these results support soft correspondence recovery and batch-level
retrieval for unseen subjects.

\section*{LLM Usage Statement}
OpenAI ChatGPT and Codex assisted with code development and with drafting and
revising the initial manuscript. The authors take responsibility for the final
content, including all scientific claims, experimental results, and references.

{
    \small
    \bibliographystyle{ieeenat_fullname}
    \bibliography{main}
}

\clearpage
\raggedbottom
\makeatletter
\setlength{\@fptop}{0pt}
\setlength{\@fpsep}{8pt plus 2pt}
\setlength{\@fpbot}{0pt plus 1fil}
\setlength{\@dblfptop}{0pt}
\setlength{\@dblfpsep}{8pt plus 2pt}
\setlength{\@dblfpbot}{0pt plus 1fil}
\makeatother
\maketitlesupplementary
\appendix
\setcounter{figure}{0}
\setcounter{table}{0}
\setcounter{equation}{0}
\renewcommand{\thefigure}{S\arabic{figure}}
\renewcommand{\thetable}{S\arabic{table}}
\renewcommand{\theequation}{S\arabic{equation}}

\newcommand{\StageDatasetTable}{%
\begin{table}[!htbp]
\centering
\small
\setlength{\tabcolsep}{2pt}
\caption{\textbf{Ranking comparisons across datasets with the NICE encoder.} Top-1/Top-5 (\%). Ranking rules in a row use exactly the same final features and costs. Initialization/update budgets match the main results: EEG2 12/40, MEG 2/8 and Alljoined 5/13.}
\label{tab:supp_datasets}
\begin{tabular*}{\columnwidth}{@{\extracolsep{\fill}}lrrr@{}}
\toprule
Dataset & Direct & CSLS & GES\\
\midrule
EEG2 & 58.63/85.18 & 63.15/87.67 & \textbf{68.25}/\textbf{90.47}\\
MEG & 5.00/16.75 & 6.13/19.63 & \textbf{7.25}/\textbf{20.38}\\
AJ & 10.65/30.18 & 11.80/32.83 & \textbf{13.25}/\textbf{35.00}\\
\bottomrule
\end{tabular*}

\end{table}}

\newcommand{\StageSubjectTable}{%
\begin{table}[!htbp]
\centering
\small
\setlength{\tabcolsep}{2pt}
\caption{\textbf{Per-subject NICE EEG2 Top-1.} Three source seeds are averaged within each subject; accuracy (\%).}
\label{tab:supp_subject}
\begin{tabular*}{\columnwidth}{@{\extracolsep{\fill}}lrrr@{}}
\toprule
Subject & Direct & CSLS & Coupling\\
\midrule
01 & 66.67 & 72.67 & \textbf{77.67}\\
02 & 64.33 & 68.83 & \textbf{76.67}\\
03 & 56.17 & 57.67 & \textbf{59.33}\\
04 & 49.00 & 52.00 & \textbf{53.83}\\
05 & 54.33 & 61.33 & \textbf{65.83}\\
06 & 63.83 & 71.00 & \textbf{80.33}\\
07 & 57.83 & 63.33 & \textbf{69.33}\\
08 & 48.67 & 50.17 & \textbf{53.17}\\
09 & 56.00 & 60.33 & \textbf{64.67}\\
10 & 69.50 & 74.17 & \textbf{81.67}\\
\bottomrule
\end{tabular*}

\end{table}}

\newcommand{\StageSeedTable}{%
\begin{table}[!htbp]
\centering
\small
\setlength{\tabcolsep}{2pt}
\caption{\textbf{Source-model seed variation.} Mean NICE EEG2 Top-1 (\%) across ten held-out subjects.}
\label{tab:supp_seed}
\begin{tabular*}{\columnwidth}{@{\extracolsep{\fill}}lrrr@{}}
\toprule
Seed & Direct & CSLS & Coupling\\
\midrule
0 & 58.40 & 62.60 & 68.40\\
1 & 58.70 & 63.45 & 67.80\\
2 & 58.80 & 63.40 & 68.55\\
\bottomrule
\end{tabular*}

\end{table}}

\newcommand{\StageProgressiveTable}{%
\begin{table}[!htbp]
\centering
\small
\setlength{\tabcolsep}{2pt}
\caption{\textbf{Cumulative feature stages and final ranking.} Top-1/Top-5 (\%). The first two rows are unadapted score references. The radial baseline and subsequent feature stages use Direct ranking of the same radial squared-distance definition. Only the final row switches to coupling ranking. Bold marks the best entry per dataset and metric.}
\label{tab:supp_ladder}
\begin{tabular*}{\columnwidth}{@{\extracolsep{\fill}}lrrr@{}}
\toprule
Stage & EEG2 & MEG & AJ\\
\midrule
Encoder cosine & 27.65/59.52 & 3.88/14.00 & 5.90/20.90\\
Encoder dot & 18.70/49.37 & 2.13/8.63 & 2.95/11.20\\
Radial baseline & 26.80/58.45 & 3.13/14.00 & 6.20/20.60\\
Moment calib. & 30.80/63.10 & 3.50/13.63 & 8.25/25.95\\
Soft init. & 53.03/81.90 & 4.75/16.75 & 10.18/30.00\\
Adaptation & 58.63/85.18 & 5.00/16.75 & 10.65/30.18\\
GES ranking & \textbf{68.25}/\textbf{90.47} & \textbf{7.25}/\textbf{20.38} & \textbf{13.25}/\textbf{35.00}\\
\bottomrule
\end{tabular*}

\end{table}}

\newcommand{\StageMegSubjectTable}{%
\begin{table}[!htbp]
\centering
\small
\setlength{\tabcolsep}{2pt}
\caption{\textbf{Per-subject MEG Top-1.} Frozen NICE encoder, source seed 0; accuracy (\%).}
\label{tab:supp_subject_meg}
\begin{tabular*}{\columnwidth}{@{\extracolsep{\fill}}lrrr@{}}
\toprule
Subject & Direct & CSLS & Coupling\\
\midrule
01 & 2.50 & 5.00 & \textbf{7.00}\\
02 & \textbf{9.00} & \textbf{9.00} & 8.50\\
03 & 6.00 & 7.50 & \textbf{8.00}\\
04 & 2.50 & 3.00 & \textbf{5.50}\\
\bottomrule
\end{tabular*}

\end{table}}

\newcommand{\StageAlljoinedSubjectTable}{%
\begin{table}[!htbp]
\centering
\small
\setlength{\tabcolsep}{2pt}
\caption{\textbf{Per-subject AJ Top-1.} Frozen NICE encoder, source seed 0; accuracy (\%).}
\label{tab:supp_subject_alljoined}
\begin{tabular*}{\columnwidth}{@{\extracolsep{\fill}}lrrr@{}}
\toprule
Subject & Direct & CSLS & Coupling\\
\midrule
01 & 12.00 & 13.00 & \textbf{13.50}\\
02 & 17.00 & 21.00 & \textbf{27.50}\\
03 & 16.50 & 20.00 & \textbf{25.00}\\
04 & 14.00 & \textbf{16.50} & 12.00\\
05 & 15.50 & 16.00 & \textbf{18.00}\\
06 & 13.50 & \textbf{15.00} & \textbf{15.00}\\
07 & 2.00 & 2.50 & \textbf{3.00}\\
08 & 2.00 & 3.50 & \textbf{5.00}\\
09 & 4.50 & 4.50 & \textbf{5.00}\\
10 & 12.50 & \textbf{13.00} & 12.50\\
11 & 10.50 & 11.50 & \textbf{12.00}\\
12 & 17.00 & 19.00 & \textbf{21.50}\\
13 & 11.00 & 12.50 & \textbf{14.50}\\
14 & 17.00 & 16.00 & \textbf{18.50}\\
15 & 10.50 & 13.00 & \textbf{13.50}\\
16 & 5.00 & 5.50 & \textbf{7.50}\\
17 & 2.50 & \textbf{3.00} & 2.50\\
18 & 14.50 & 15.00 & \textbf{17.00}\\
19 & \textbf{2.00} & 0.50 & 1.50\\
20 & 13.50 & 15.00 & \textbf{20.00}\\
\bottomrule
\end{tabular*}

\end{table}}

\newcommand{\StageSubjectSpreadTable}{%
\begin{table}[!htbp]
\centering
\small
\setlength{\tabcolsep}{2pt}
\caption{\textbf{Between-subject variability.} Mean and sample standard deviation of Top-1/Top-5 (\%). EEG2 averages its three source seeds within each subject first. MEG and Alljoined use seed 0; n is the number of subjects.}
\label{tab:supp_spread}
\begin{tabular*}{\columnwidth}{@{\extracolsep{\fill}}lllrr@{}}
\toprule
Dataset & \(n\) & Ranking & Top-1 & Top-5\\
\midrule
EEG2 & 10 & Direct & 58.63 \(\pm\) 7.20 & 85.18 \(\pm\) 5.77\\
EEG2 & 10 & CSLS & 63.15 \(\pm\) 8.43 & 87.67 \(\pm\) 5.58\\
EEG2 & 10 & Coupling & 68.25 \(\pm\) 10.65 & 90.47 \(\pm\) 5.20\\
MEG & 4 & Direct & 5.00 \(\pm\) 3.14 & 16.75 \(\pm\) 8.86\\
MEG & 4 & CSLS & 6.13 \(\pm\) 2.66 & 19.63 \(\pm\) 8.54\\
MEG & 4 & Coupling & 7.25 \(\pm\) 1.32 & 20.38 \(\pm\) 9.76\\
AJ & 20 & Direct & 10.65 \(\pm\) 5.56 & 30.18 \(\pm\) 12.21\\
AJ & 20 & CSLS & 11.80 \(\pm\) 6.28 & 32.83 \(\pm\) 13.66\\
AJ & 20 & Coupling & 13.25 \(\pm\) 7.45 & 35.00 \(\pm\) 14.86\\
\bottomrule
\end{tabular*}

\end{table}}

\newcommand{\StageNormalizationsTable}{%
\begin{table}[!htbp]
\centering
\small
\setlength{\tabcolsep}{2pt}
\caption{\textbf{Ranking methods at fixed costs.} Top-1/Top-5 (\%). Batch IS computes normalizers from the current query batch. The matched inverse temperature is \(1/0.035\).}
\label{tab:supp_related_normalizations}
\begin{tabular*}{\columnwidth}{@{\extracolsep{\fill}}lrrr@{}}
\toprule
Ranking & EEG2 & MEG & AJ\\
\midrule
Direct & 58.63/85.18 & 5.00/16.75 & 10.65/30.18\\
CSLS & 63.15/87.67 & 6.13/19.63 & 11.80/32.83\\
Batch IS, \(\beta=20\) & 63.52/88.43 & 7.13/\textbf{20.75} & 12.58/33.60\\
IS, \(1/\varepsilon\) & 63.17/88.30 & 6.88/20.00 & 12.48/33.53\\
GES & \textbf{68.25}/\textbf{90.47} & \textbf{7.25}/20.38 & \textbf{13.25}/\textbf{35.00}\\
\bottomrule
\end{tabular*}

\end{table}}

\newcommand{\StageGridTable}{%
\begin{table*}[!htbp]
\centering
\small
\setlength{\tabcolsep}{2pt}
\caption{\textbf{Effects of column constraints and entropy at fixed costs.} Top-1/Top-5 (\%). Columns use penalty ratio \(\xi/\eta=0,1,10,\infty\); the last enforces uniform column mass exactly. Each setting is evaluated on the same 54 subject--seed cases.}
\label{tab:supp_e1grid}
\begin{tabular*}{\textwidth}{@{\extracolsep{\fill}}llrrrr@{}}
\toprule
Dataset & \(\eta\) & Row only & Ratio 1 & Ratio 10 & Exact balance\\
\midrule
EEG2 & 0.0175 & 58.63/85.18 & 61.52/87.50 & 67.38/89.63 & \textbf{68.25}/\textbf{90.03}\\
EEG2 & 0.035 & 58.63/85.18 & 63.20/88.07 & 67.95/90.17 & \textbf{68.25}/\textbf{90.47}\\
EEG2 & 0.07 & 58.63/85.18 & 64.22/88.02 & 66.92/89.83 & \textbf{67.27}/\textbf{89.97}\\
MEG & 0.0175 & 5.00/16.75 & 5.63/18.63 & 6.88/20.00 & \textbf{7.00}/\textbf{20.13}\\
MEG & 0.035 & 5.00/16.75 & 6.00/19.00 & 7.13/20.25 & \textbf{7.25}/\textbf{20.38}\\
MEG & 0.07 & 5.00/16.75 & 6.00/19.13 & 7.00/\textbf{20.88} & \textbf{7.25}/20.63\\
AJ & 0.0175 & 10.65/30.18 & 11.53/32.10 & 12.88/\textbf{34.33} & \textbf{13.18}/34.18\\
AJ & 0.035 & 10.65/30.18 & 11.68/32.58 & 13.05/\textbf{35.15} & \textbf{13.25}/35.00\\
AJ & 0.07 & 10.65/30.18 & 12.03/32.88 & 13.23/35.10 & \textbf{13.40}/\textbf{35.63}\\
\bottomrule
\end{tabular*}

\end{table*}}

\newcommand{\StageTransitionsTable}{%
\begin{table}[!htbp]
\centering
\small
\setlength{\tabcolsep}{2pt}
\caption{\textbf{Prediction changes with coupling ranking.} Counts include all queries and both corrected and newly incorrect predictions. EEG2 totals cover ten subjects, each evaluated with three source seeds.}
\label{tab:supp_correction_transitions}
\begin{tabular*}{\columnwidth}{@{\extracolsep{\fill}}llrrrrr@{}}
\toprule
Data & From & Corr. & Harm & Both+ & Both-- & Net\\
\midrule
EEG2 & Direct & 795 & 218 & 3300 & 1687 & 9.62\\
EEG2 & CSLS & 495 & 189 & 3600 & 1716 & 5.10\\
MEG & Direct & 32 & 14 & 26 & 728 & 2.25\\
MEG & CSLS & 18 & 9 & 40 & 733 & 1.13\\
AJ & Direct & 266 & 162 & 264 & 3308 & 2.60\\
AJ & CSLS & 185 & 127 & 345 & 3343 & 1.45\\
\bottomrule
\end{tabular*}

\end{table}}

\newcommand{\StageMatchedTable}{%
\begin{table}[!htbp]
\centering
\small
\setlength{\tabcolsep}{2pt}
\caption{\textbf{Matched initialization constructions on NICE EEG2.} Final Top-1/Top-5 (\%) after the same K12 initialization and 40 updates. Each intervention replays the reference query-confidence weights at every round and uses the same regularized Procrustes estimator. Only correspondence construction changes. Bold is per ranking rule and metric across constructions.}
\label{tab:supp_matched_init}
\begin{tabular*}{\columnwidth}{@{\extracolsep{\fill}}lrrr@{}}
\toprule
Pairs & Direct & CSLS & GES\\
\midrule
Row-hard & 58.37/84.38 & 62.32/86.77 & 66.53/89.90\\
Assignment & 56.30/83.42 & 60.35/86.07 & 64.42/89.08\\
Row-softmax & 48.38/79.50 & 57.35/85.48 & 65.78/89.42\\
Soft coupling & \textbf{58.63}/\textbf{85.18} & \textbf{63.15}/\textbf{87.67} & \textbf{68.25}/\textbf{90.47}\\
\bottomrule
\end{tabular*}

\end{table}}

\newcommand{\StagePairedTable}{%
\begin{table}[!htbp]
\centering
\small
\setlength{\tabcolsep}{2pt}
\caption{\textbf{Soft-initialization paired differences.} Final coupled Top-1 change (percentage points); descriptive 95\% subject-bootstrap interval. Signs are over ten subjects after averaging their three source seeds.}
\label{tab:supp_matched_paired}
\begin{tabular*}{\columnwidth}{@{\extracolsep{\fill}}lrrr@{}}
\toprule
Control & Mean & 95\% interval & + / = / -\\
\midrule
Row-hard & 1.72 & [0.85, 2.65] & 9 / 1 / 0\\
Assignment & 3.83 & [2.15, 5.45] & 9 / 0 / 1\\
Row-softmax & 2.47 & [1.62, 3.63] & 10 / 0 / 0\\
\bottomrule
\end{tabular*}

\end{table}}

\newcommand{\StageEntropyTable}{%
\begin{table}[!htbp]
\centering
\small
\setlength{\tabcolsep}{2pt}
\caption{\textbf{Entropy sensitivity at fixed EEG2 costs.} Accuracy and argmax coverage (\%). Cost is \(\langle P,C\rangle/\sum P\). The hard assignment has no defined Top-5 ordering, so that entry is omitted.}
\label{tab:supp_r03eps}
\begin{tabular*}{\columnwidth}{@{\extracolsep{\fill}}lrrrr@{}}
\toprule
\(\eta\) & Top-1 & Top-5 & Coupling cost & Coverage\\
\midrule
0 & 67.32 & --- & 0.4915 & 100.00\\
0.001 & 67.18 & 89.63 & 0.4915 & 99.83\\
0.005 & 67.28 & 89.87 & 0.4919 & 98.73\\
0.014 & 68.18 & 89.98 & 0.4938 & 96.22\\
0.02 & \textbf{68.37} & 90.18 & 0.4962 & 94.87\\
0.035 & 68.25 & \textbf{90.47} & 0.5081 & 91.95\\
0.07 & 67.27 & 89.97 & 0.5744 & 86.90\\
0.14 & 64.32 & 88.38 & 0.7350 & 81.12\\
0.35 & 57.53 & 84.38 & 0.9620 & 71.78\\
1.0 & 47.12 & 77.40 & 1.1321 & 59.97\\
\bottomrule
\end{tabular*}

\end{table}}

\newcommand{\StageSupportTable}{%
\begin{table}[!htbp]
\centering
\small
\setlength{\tabcolsep}{2pt}
\caption{\textbf{Preferred-edge controls on NICE EEG2.} Accuracy and edge density (\%). Edges outside the preferred set retain kernel multiplier \(10^{-6}\); outside mass is a fraction of total coupling mass. These are fixed-cost support interventions.}
\label{tab:supp_r11support}
\begin{tabular*}{\columnwidth}{@{\extracolsep{\fill}}lrrrr@{}}
\toprule
Preferred set & Top-1 & Top-5 & Edges & Outside\\
\midrule
Full & 68.25 & \textbf{90.47} & 100.00 & 0.00e+00\\
MNN & 63.22 & 83.65 & 0.27 & 4.70e-01\\
Top-5 union & 67.78 & 88.03 & 3.68 & 8.43e-07\\
Top-10 union & 68.28 & 89.95 & 7.19 & 2.83e-08\\
Top-20 union & \textbf{68.30} & 90.42 & 13.81 & 3.49e-09\\
Top-50 union & 68.25 & \textbf{90.47} & 32.19 & 7.55e-11\\
\bottomrule
\end{tabular*}

\end{table}}

\newcommand{\StageWeightsTable}{%
\begin{table}[!htbp]
\centering
\small
\setlength{\tabcolsep}{2pt}
\caption{\textbf{Query-confidence sources at initialization.} NICE EEG2 final coupled accuracy (\%). The coupling matrix is retained in all four arms; only the additional row-confidence weights change.}
\label{tab:supp_r01weights}
\begin{tabular*}{\columnwidth}{@{\extracolsep{\fill}}lrr@{}}
\toprule
Row-confidence source & Top-1 & Top-5\\
\midrule
Coupling entropy & \textbf{68.25} & \textbf{90.47}\\
Uniform & 68.02 & 90.28\\
Similarity row & 68.03 & 90.32\\
Row-softmax entropy & 68.02 & 90.32\\
\bottomrule
\end{tabular*}

\end{table}}

\newcommand{\StageObjectivesTable}{%
\begin{table}[!htbp]
\centering
\small
\setlength{\tabcolsep}{2pt}
\caption{\textbf{Objectives with a shared teacher construction.} NICE EEG2 accuracy (\%); parentheses give update counts. The teacher recipe is unweighted and identical across arms; refreshed teacher values follow each arm's feature trajectory. All use K12 initialization and final coupling ranking.}
\label{tab:supp_objectives_nice}
\begin{tabular*}{\columnwidth}{@{\extracolsep{\fill}}lrr@{}}
\toprule
Objective & Top-1 & Top-5\\
\midrule
Initialization only (0) & 65.68 & 89.68\\
Scaled attraction (40) & 65.03 & 88.55\\
Row cross-entropy (40) & 67.75 & 89.47\\
Coupling matching (40) & \textbf{68.25} & \textbf{90.47}\\
\bottomrule
\end{tabular*}

\end{table}}

\newcommand{\StageWeightedObjectivesTable}{%
\begin{table}[!htbp]
\centering
\small
\setlength{\tabcolsep}{2pt}
\caption{\textbf{Confidence-weighted teacher references.} NICE EEG2 accuracy (\%); parentheses give updates. These arms multiply each teacher row by its entropy-confidence weight before unit-mass normalization, comparing complete weighted strategies alongside Table~\ref{tab:supp_objectives_nice}.}
\label{tab:supp_weighted_objectives}
\begin{tabular*}{\columnwidth}{@{\extracolsep{\fill}}lrr@{}}
\toprule
Weighted objective & Top-1 & Top-5\\
\midrule
Attraction (40) & 66.62 & 89.62\\
Attraction (80) & 66.32 & \textbf{89.67}\\
Row cross-entropy (40) & \textbf{67.50} & 89.48\\
Row cross-entropy (80) & 66.58 & 88.97\\
\bottomrule
\end{tabular*}

\end{table}}

\newcommand{\StageGalleryTable}{%
\begin{table*}[!htbp]
\centering
\small
\setlength{\tabcolsep}{2pt}
\caption{\textbf{Gallery expansion with geometry fitted on 200 candidates.} NICE EEG2 Top-1/Top-5 (\%). Geometry, encoders and query set remain fixed; the ranking gallery and its associated cost normalization are recomputed. Equal mass uses column target 1/M; capacity uses upper bound 1/200.}
\label{tab:supp_gallery}
\begin{tabular*}{\textwidth}{@{\extracolsep{\fill}}lrrrr@{}}
\toprule
Candidates & Direct & CSLS & Equal mass & Capacity\\
\midrule
200 & 58.63/85.18 & 63.15/87.67 & \textbf{68.25}/\textbf{90.47} & \textbf{68.25}/\textbf{90.47}\\
700 & 50.30/74.63 & 51.62/\textbf{75.70} & 42.13/68.83 & \textbf{52.22}/75.45\\
1200 & 45.42/69.63 & 45.58/\textbf{70.22} & 34.65/61.08 & \textbf{46.30}/70.03\\
1854 & 41.55/65.78 & 41.40/65.63 & 29.18/54.43 & \textbf{42.05}/\textbf{65.93}\\
\bottomrule
\end{tabular*}

\end{table*}}

\newcommand{\StageGalleryMassTable}{%
\begin{table*}[!htbp]
\centering
\small
\setlength{\tabcolsep}{2pt}
\caption{\textbf{Matching-mass distribution as the ranking gallery expands.} Percent of total mass. Unvisited candidates receive no Direct top-1 votes and retain positive entropic coupling mass.}
\label{tab:supp_gallerydiag}
\begin{tabular*}{\textwidth}{@{\extracolsep{\fill}}llrrr@{}}
\toprule
Candidates & Column rule & True pairs & Direct winners & Unvisited\\
\midrule
200 & Equal mass & 57.17 & 59.32 & 26.23\\
200 & Capacity & 57.17 & 59.32 & 26.23\\
700 & Equal mass & 14.97 & 17.45 & 76.36\\
700 & Capacity & 38.75 & 53.51 & 32.64\\
1200 & Equal mass & 8.71 & 10.59 & 85.63\\
1200 & Capacity & 31.98 & 47.63 & 40.24\\
1854 & Equal mass & 5.61 & 7.08 & 90.37\\
1854 & Capacity & 27.30 & 42.95 & 46.67\\
\bottomrule
\end{tabular*}

\end{table*}}

\newcommand{\StageTransferTable}{%
\begin{table*}[!htbp]
\centering
\small
\setlength{\tabcolsep}{2pt}
\caption{\textbf{Complete-map transfer to disjoint queries.} NICE EEG2 Top-1/Top-5 (\%). Fit on 100 calibration identities, evaluate 100 different identities with all transform components frozen. Calibration images are excluded from the retrieval gallery; adding 1,654 training concepts gives maximum M=1,754.}
\label{tab:supp_r04split}
\begin{tabular*}{\textwidth}{@{\extracolsep{\fill}}lrrrr@{}}
\toprule
Candidates & Direct & CSLS & Equal mass & Capacity\\
\midrule
100 & 57.37/88.57 & 64.67/92.33 & \textbf{75.47}/\textbf{95.43} & \textbf{75.47}/\textbf{95.43}\\
200 & 51.07/83.03 & 54.23/\textbf{85.33} & 50.83/84.70 & \textbf{57.37}/85.07\\
700 & 37.00/67.57 & \textbf{38.50}/68.00 & 26.20/57.03 & 38.23/\textbf{68.13}\\
1754 & 25.57/53.20 & \textbf{26.03}/\textbf{53.47} & 15.57/38.40 & 25.80/53.27\\
\bottomrule
\end{tabular*}

\end{table*}}

\newcommand{\StageFullGalleryTable}{%
\begin{table}[!htbp]
\centering
\small
\caption{\textbf{Adaptation with the full expanded gallery.}
NICE EEG2 Top-1/Top-5 (\%). Both methods fit using all candidates
at the stated gallery size. STAGE uses capacity constraints throughout
fitting and ranking; SCORE uses its recovery and CSLS.
Each row averages three source seeds within each of ten subjects.}
\label{tab:supp_fullgallery_matched}
\begin{tabular*}{\columnwidth}{@{\extracolsep{\fill}}lrr@{}}
\toprule
Candidates & SCORE & STAGE\\
\midrule
700 & 27.83/\textbf{57.48} & \textbf{30.08}/53.95\\
1854 & \textbf{17.02}/\textbf{38.50} & 16.97/33.23\\
\bottomrule
\end{tabular*}

\end{table}}

\newcommand{\StageSensitivityTable}{%
\begin{table}[!htbp]
\centering
\small
\setlength{\tabcolsep}{2pt}
\caption{\textbf{One-factor sensitivity on NICE EEG2.} Final coupled accuracy (\%). Each row uses 30 cases; changing one setting leaves the remaining defaults unchanged.}
\label{tab:supp_sensitivity}
\begin{tabular*}{\columnwidth}{@{\extracolsep{\fill}}llrr@{}}
\toprule
Setting & Value & Top-1 & Top-5\\
\midrule
Default & K12 / 40 & 68.25 & 90.47\\
Refresh & 1 & 67.57 & 90.07\\
Refresh & 5 & 68.73 & 90.30\\
Refresh & 20 & 68.00 & 90.53\\
Refresh & 40 & 67.35 & 90.50\\
Anchor & 0.0 & 68.28 & 90.42\\
Anchor & 0.05 & 68.18 & 90.45\\
Anchor & 0.3 & 68.33 & 90.55\\
Anchor & 0.5 & 68.38 & 90.65\\
Rank & 4 & 66.03 & 89.77\\
Rank & 8 & 66.33 & 89.65\\
Rank & 32 & 69.30 & 89.82\\
Rank & 64 & 66.77 & 88.93\\
Ridge & 0.02 & 71.65 & 91.78\\
Ridge & 0.05 & 70.95 & 91.68\\
Ridge & 0.2 & 66.07 & 89.42\\
Ridge & 0.4 & 63.60 & 88.25\\
\bottomrule
\end{tabular*}

\end{table}}

\newcommand{\StageExtraSensitivityTable}{%
\begin{table}[!htbp]
\centering
\small
\setlength{\tabcolsep}{2pt}
\caption{\textbf{Learning rate, student entropy, and initialization budget.} NICE EEG2 coupled accuracy (\%). Each row changes one factor from K12/40; \(\dagger\) denotes the reference setting.}
\label{tab:supp_extra_sensitivity}
\begin{tabular*}{\columnwidth}{@{\extracolsep{\fill}}llrr@{}}
\toprule
Setting & Value & Top-1 & Top-5\\
\midrule
Learning rate & 0.003 & 68.78 & 90.63\\
Learning rate & 0.005 & 68.55 & 90.60\\
Learning rate & 0.01\(\dagger\) & 68.25 & 90.47\\
Learning rate & 0.02 & 68.55 & 90.62\\
Learning rate & 0.03 & 68.63 & 90.50\\
Learning rate & 0.1 & 67.50 & 89.88\\
Student \(\tau\) & 0.03 & 63.40 & 88.48\\
Student \(\tau\) & 0.05 & 67.65 & 90.33\\
Student \(\tau\) & 0.07\(\dagger\) & 68.25 & 90.47\\
Student \(\tau\) & 0.1 & 69.07 & 90.78\\
Student \(\tau\) & 0.2 & 69.12 & 90.90\\
Init. rounds & 0 & 57.72 & 85.17\\
Init. rounds & 1 & 63.67 & 88.30\\
Init. rounds & 2 & 65.80 & 89.20\\
Init. rounds & 5 & 67.73 & 90.13\\
Init. rounds & 8 & 68.12 & 90.45\\
Init. rounds & 12\(\dagger\) & 68.25 & 90.47\\
\bottomrule
\end{tabular*}

\end{table}}

\newcommand{\StageUpdateTable}{%
\begin{table}[!htbp]
\centering
\small
\setlength{\tabcolsep}{2pt}
\caption{\textbf{Initialization and longer optimization.} NICE EEG2 coupled accuracy (\%), with the same learning rate 0.01 and rank-16 transform. Each optimization budget is fitted independently.}
\label{tab:supp_noinit}
\begin{tabular*}{\columnwidth}{@{\extracolsep{\fill}}llrr@{}}
\toprule
Initialization & Updates & Top-1 & Top-5\\
\midrule
K12 & 0 & 65.68 & 89.68\\
K12 & 5 & 67.48 & 90.70\\
K12 & 10 & 67.30 & 90.52\\
K12 & 20 & 67.85 & 90.43\\
K12 & 30 & 68.20 & 90.53\\
K12 & 40 & 68.25 & 90.47\\
K12 & 80 & 68.47 & 90.07\\
K12 & 160 & 67.63 & 89.62\\
K0 & 0 & 48.08 & 77.90\\
K0 & 1 & 50.18 & 80.02\\
K0 & 40 & 57.72 & 85.17\\
K0 & 100 & 59.30 & 85.50\\
K0 & 200 & 59.42 & 84.68\\
K0 & 300 & 59.05 & 84.70\\
K0 & 600 & 58.78 & 84.37\\
\bottomrule
\end{tabular*}

\end{table}}

\newcommand{\StageEncoderTable}{%
\begin{table}[!htbp]
\centering
\small
\setlength{\tabcolsep}{2pt}
\caption{\textbf{Additional frozen encoder: SAMGA--TSConv.} Top-1/Top-5 (\%). All three datasets here use K12 and 40 updates. EEG2 averages ten subjects and three source seeds; MEG four subjects and Alljoined twenty subjects, both at seed 0. Unadapted is the original dot-product score; final ranking rules share the learned radial cost.}
\label{tab:supp_archived_stages}
\begin{tabular*}{\columnwidth}{@{\extracolsep{\fill}}lrrrr@{}}
\toprule
Data & Encoder dot & Direct & CSLS & GES\\
\midrule
EEG2 & 18.87/49.58 & 56.98/83.20 & 60.80/86.00 & \textbf{66.30}/\textbf{88.95}\\
MEG & 2.38/10.75 & 5.75/16.88 & \textbf{6.13}/\textbf{18.25} & 5.75/17.00\\
AJ & 3.50/12.75 & 10.03/27.55 & 11.10/29.15 & \textbf{11.38}/\textbf{30.83}\\
\bottomrule
\end{tabular*}

\end{table}}

\newcommand{\StageUniformSeedTable}{%
\begin{table}[!htbp]
\centering
\small
\setlength{\tabcolsep}{2pt}
\caption{\textbf{Source-seed replication at a uniform K12/40 budget.} Final coupling Top-1/Top-5 (\%), averaged over all held-out subjects per seed. NICE and TSConv use the same frozen inference protocol. The uniform budget is K12/40; the dataset-specific MEG and Alljoined budgets in the main evaluation are K2/8 and K5/13, respectively.}
\label{tab:supp_uniform_seeds}
\begin{tabular*}{\columnwidth}{@{\extracolsep{\fill}}llrrr@{}}
\toprule
Dataset & Encoder & Seed 0 & Seed 1 & Seed 2\\
\midrule
EEG2 & NICE & 68.40/90.50 & 67.80/90.60 & 68.55/90.30\\
EEG2 & TSConv & 67.65/90.80 & 66.75/87.90 & 64.50/88.15\\
MEG & NICE & 5.63/18.88 & 5.00/17.50 & 5.00/19.50\\
MEG & TSConv & 5.75/17.00 & 7.13/17.00 & 6.00/18.13\\
AJ & NICE & 12.43/33.75 & 12.03/32.20 & 12.70/34.13\\
AJ & TSConv & 11.38/30.83 & 11.65/32.55 & 11.40/30.70\\
\bottomrule
\end{tabular*}

\end{table}}

\newcommand{\StagePermutationTable}{%
\begin{table}[!htbp]
\centering
\small
\setlength{\tabcolsep}{2pt}
\caption{\textbf{Independent query/gallery row permutations.} Complete K12/40 pipeline, thirty EEG2 cases. Errors are maxima over cases and entries. Rank changes count true-image ranks over all 6,000 query--seed evaluations after inverse permutation and identity-stable tie breaking.}
\label{tab:supp_permutation_current}
\begin{tabular*}{\columnwidth}{@{\extracolsep{\fill}}lr@{}}
\toprule
Quantity & Maximum error or count\\
\midrule
Moment scale/offset & 4.44e-16\\
Initialization map/centroids & 6.63e-09\\
Final pairwise cost & 3.77e-03\\
Cases below \(10^{-7}\) & 28 / 30\\
Direct truth-rank changes & 2\\
CSLS truth-rank changes & 0\\
GES truth-rank changes & 0\\
\bottomrule
\end{tabular*}

\end{table}}

\section{Data, Evaluation Protocols, and Implementation}
\label{sec:supp-protocol}

\paragraph{Evaluation population.}
The mechanism analyses use frozen NICE brain encoders~\cite{song2024nice} and
frozen visual features under the source-training protocol of
Jiang et al.~\cite{jiang2026samga}.
THINGS-EEG2 has ten held-out subjects and three source-model seeds
(0, 1 and 2); THINGS-MEG has four subjects at seed 0; Alljoined-1.6M
has twenty subjects at seed 0. Section~\ref{sec:supp-coupling} extends
MEG and Alljoined to all three source seeds at their main stage budgets.
Each closed-set evaluation uses all
200 queries and all 200 official candidates. Tables abbreviate the
datasets as EEG2, MEG and AJ. Neither target identities
nor their pair labels enter fitting. All ranking comparisons within a
feature condition reuse its saved pairwise costs.

\paragraph{Neural inputs and query construction.}
Repeated test responses to the same image are averaged in signal space
before the frozen brain encoder. Thus, each held-out subject contributes
200 image-specific averaged responses, yielding 200 query embeddings.
Table~\ref{tab:supp_neural_inputs} specifies the signal inputs.
EEG2 averages twenty repetitions from each of four sessions; MEG averages
one response per test image from each of twelve sessions.
For Alljoined, trials marked as dropped in the released metadata are
excluded, and all remaining responses to each image are averaged across
sessions. Its repetition count therefore varies with the retained trials.
The EEG inputs retain the baseline correction and multivariate noise
normalization (MVNN) of their preprocessed data; MEG uses the released
preprocessed epochs cropped to the post-stimulus interval.

\begin{table*}[t]
\centering
\small
\setlength{\tabcolsep}{4pt}
\caption{\textbf{Neural inputs used to construct test queries.}
Shapes describe one averaged response before encoding. EEG windows retain
their preprocessed sampling grids. Every dataset provides one query for
each of the 200 test images; trial averaging precedes feature extraction.}
\label{tab:supp_neural_inputs}
\begin{tabular*}{\textwidth}{@{\extracolsep{\fill}}lcp{0.19\textwidth}p{0.21\textwidth}p{0.24\textwidth}@{}}
\toprule
Dataset & \shortstack{Channels\\$\times$ samples}
& Sampling and window & Preprocessing & Test-trial aggregation\\
\midrule
THINGS-EEG2~\cite{gifford2022things}
& $63\times250$
& 250 Hz; first post-stimulus second
& Baseline correction and session-wise MVNN
& Mean of 80 repetitions per image across four sessions\\
THINGS-MEG~\cite{hebart2023things}
& $271\times201$
& 200 Hz; 0--1 s, both endpoints included
& Released preprocessed MEG epochs; temporal crop
& Mean of 12 repetitions per image across twelve sessions\\
Alljoined-1.6M~\cite{xu2025alljoined}
& $32\times250$
& 250 Hz; first post-stimulus second
& Baseline-corrected, MVNN-preprocessed EEG; released trial-retention mask
& Mean of all retained trials per image across sessions\\
\bottomrule
\end{tabular*}

\end{table*}

\paragraph{Source-model training.}
We follow the source-training framework of SAMGA~\cite{jiang2026samga},
replacing its brain-signal encoder with NICE~\cite{song2024nice} for the
main results while retaining its visual branch and alignment procedure.
Each LOSO fold learns from paired brain responses and images of the
source subjects. The brain encoder produces a 1,024-dimensional feature,
which a linear projector maps to 512 dimensions. A shared
$512\!\rightarrow\!512$ linear map is applied to both modalities.
After source training, the brain encoder, projectors, shared map and
visual fusion weights are frozen; STAGE operates on their embeddings.

\paragraph{Visual features and fusion.}
The frozen InternViT-6B-448px-V2.5 backbone~\cite{chen2024internvl25}
provides 3,200-dimensional CLS-token features from blocks
20, 24, 28, 32 and 36 (zero-based indices).
Images are resized with bicubic interpolation to a shortest side of
448 pixels, center-cropped to $448\times448$, and normalized with
ImageNet channel statistics.
A shared $3,200\!\rightarrow\!1,024$ linear projection and five
layer-specific $1,024\!\rightarrow\!512$ projections precede SAMGA's
subject-aware layer fusion and the shared map.
The projections and fusion weights are learned on source subjects.
At test time, fusion uses its learned global weights without a
target-subject embedding. Image augmentation and text supervision
are disabled.

\paragraph{Source optimization.}
Table~\ref{tab:supp_source_training} summarizes the source-training
settings. During epochs 1--20, the loss combines maximum mean discrepancy
(MMD), which aligns the two feature distributions, with symmetric
brain--image contrastive learning, which preserves image correspondence.
The MMD weight decreases linearly from 0.9 to 0.5, with the complementary
weight assigned to the contrastive loss. MMD averages Gaussian kernels
with bandwidths $\{0.1,0.2,0.5,1,2\}$.
During epochs 21--50, the shared map is frozen and the remaining
trainable modules use the contrastive loss alone.
The contrastive logits normalize image features but leave brain-feature
norms unchanged. Source-signal augmentation applies temporal smoothing
with kernel size five and probability 0.3.
The final checkpoint at epoch 50 supplies the frozen embeddings.

\paragraph{Concept-disjoint validation and epoch selection.}
We partition only the official training set by image concept, using
random seed 42 to assign its 1,654 concepts to 1,489 training concepts
and 165 validation concepts (approximately 9:1).
All images and associated brain responses belonging to the same concept
remain in one partition, and all subjects use the same concept split.
Within each LOSO fold, only source-subject data enter this procedure.
We train on the internal training partition and use validation Top-1
to select the number of epochs for both source-training phases:
MMD plus contrastive learning, followed by contrastive learning alone.
Once the two epoch budgets are fixed, we reinitialize the trainable
source-model components and train from scratch on the complete official
training set, including the internal validation concepts, using the
fixed schedule. The final checkpoint supplies the frozen embeddings.
Official test data are excluded from source-model fitting and epoch selection;
their unlabeled embeddings are used only for the prescribed test-time
adaptation and retrieval.

\begin{table}[t]
\centering
\small
\setlength{\tabcolsep}{4pt}
\caption{\textbf{Source-model training settings.}
These settings train the frozen base before STAGE. The optimizer is
recreated at epoch 21 after freezing the shared map. The contrastive
scale follows SAMGA's fixed softplus parameterization.}
\label{tab:supp_source_training}
\begin{tabular*}{\columnwidth}{@{\extracolsep{\fill}}lp{0.48\columnwidth}@{}}
\toprule
Setting & Value\\
\midrule
Optimizer & AdamW, $\beta_1=0.9$, $\beta_2=0.999$\\
Weight decay & $10^{-4}$\\
Training epochs & 50 (20 + 30)\\
Learning rate & $10^{-4}$; $5\times10^{-5}$ from epoch 21\\
Batch size (EEG2/MEG/AJ) & 1,024 / 128 / 1,024\\
Contrastive logit scale & $\operatorname{softplus}(\log(1/0.07))$, fixed\\
Layer-fusion temperature & 1.0\\
Subject-bias dropout & 0.3\\
Source-model seeds & 0, 1, 2\\
\bottomrule
\end{tabular*}

\end{table}

\paragraph{Source-only protocol for test-time budget selection.}
We define a source-only protocol within each outer LOSO fold.
Source subjects serve in turn as simulated targets, each excluded from
the corresponding validation encoder's training.
This budget-validation split is separate from source-epoch selection:
for EEG2, seed 42 partitions the 1,654 official training concepts into
1,454 training and 200 validation concepts; other datasets use
concept-disjoint partitions of their official training data.
All subjects share the concept split. One deterministically selected
image per validation concept supplies an image-specific brain query
and a gallery candidate. The encoder remains frozen during STAGE
validation, and identities are used only to score Top-1.
The search covers initialization rounds
$K\in\{0,1,2,4,8,12,16\}$ and adaptation steps
$T\in\{0,4,8,16,24,40,60\}$, followed by a predefined search over
neighboring integer budgets.
Selection maximizes validation Top-1, averaging source-model seeds
within each simulated target and then weighting subjects equally;
ties favor fewer adaptation steps, then fewer initialization rounds.
Each fold's selected budget is fixed before target evaluation.
The final source encoder is retrained from scratch on all source
subjects' complete official training data before target calibration.
The reported results use the fixed dataset-level budgets listed below.

\paragraph{Stage budgets and configuration comparisons.}
The main results use twelve initialization rounds and forty updates for
EEG2, two rounds and eight updates for MEG, and five rounds and thirteen
updates for Alljoined, with three source seeds per dataset.
The mechanism comparisons use the same dataset-specific budgets,
retaining the source-seed populations stated above.
The remaining defaults are shared (Table~\ref{tab:supp_controls}).
Section~\ref{sec:supp-archived} separately reports a uniform K12/40
budget for additional encoder and source-seed comparisons.

\paragraph{Batch processing cost.}
Our setting collects an unlabeled response batch before calibration
and decoding. Table~\ref{tab:supp_batch_runtime} reports computation
on the resulting frozen embeddings with the dataset-specific budgets
above, excluding signal acquisition, preprocessing and encoder feature
extraction.

\begin{table}[t]
\centering
\small
\caption{\textbf{Batch processing time (seconds).}
Each batch contains 200 queries and 200 candidates. Measurements use
an Intel Xeon Gold 6458Q CPU with four threads, subject 01 and source
seed 0 for each dataset. Entries are medians of three runs after one
warm-up.}
\label{tab:supp_batch_runtime}
\begin{tabular*}{\columnwidth}{@{\extracolsep{\fill}}lrrr@{}}
\toprule
Stage & EEG2 & MEG & AJ\\
\midrule
Moment calibration & 0.07 & 0.07 & 0.07\\
Soft initialization & 5.37 & 0.36 & 0.82\\
Feature-map setup & 0.09 & 0.09 & 0.09\\
Feature adaptation & 6.99 & 0.47 & 0.95\\
Coupling ranking & 1.55 & 0.14 & 0.08\\
\midrule
Total & 14.09 & 1.13 & 2.01\\
\bottomrule
\end{tabular*}

\end{table}

\paragraph{Aggregation and subject variation.}
Accuracy is computed by matching query and gallery identities, with
Top-1 and Top-5 evaluated over the full gallery. Each subject receives
equal weight. EEG2 first averages the three source seeds within each
subject, then averages the ten subjects; MEG and Alljoined average their
four and twenty subjects, respectively. Source seeds are repeated
measurements within each subject. Subject standard deviations
use the sample denominator \(n-1\). Descriptive 95\% intervals resample
subjects 10,000 times with seed 20261111, after averaging EEG2 seeds.
Section~\ref{sec:supp-coupling} applies seed averaging to all three
datasets and uses bootstrap seed 20261112.
Displayed percentages use the unrounded aggregates and consistent
round-half-up rounding; Top-1 and Top-5 maxima are marked separately,
including ties.

\paragraph{Gallery and transfer protocols.}
Closed-set analyses jointly fit and rank the same unlabeled
200-query/200-image batch, with one query per candidate identity.
The candidate set is known from the stimulus protocol and covers all
query identities; individual query--image pairings are withheld from fitting.
Ranking-gallery expansion freezes this fitted geometry
before increasing the retrieval gallery to 1,854. Identity-disjoint
transfer instead fits on 100 identities and retrieves 100 different
identities from the same subject, excluding calibration images and
giving at most 1,754 candidates. Full-gallery fitting starts with
700 or 1,854 images in every fitting stage.
Section~\ref{sec:supp-gallery} reports these protocols separately.
Identity-disjoint transfer measures within-subject generalization across
different image identities. Section~\ref{sec:supp-repeated}
instead keeps 200 query rows and 200 candidates while changing identity
frequencies through synthetic embedding repetition.

\paragraph{Reproducibility.}
The source-model seeds above identify different frozen encoders.
The feature-adaptation seed is separately fixed to 1701. All target
updates use float64, zero-initialized residual parameters, the prescribed
update count and its final state. Encoders and image features remain fixed
throughout. Sensitivity plots mark the reference configuration alongside
one-factor variations.

\paragraph{Matched SCORE comparison.}
The NICE-based SCORE row in the main table runs its complete target-side
pipeline: moment calibration, MNN landmark selection, regularized
orthogonal recovery and cosine-CSLS ranking. SCORE and STAGE use the
same frozen brain and image embeddings, all 200 official queries and
candidates, all held-out subjects, and source seeds 0, 1 and 2. We
average seeds within each subject before averaging subjects.
SCORE uses \(k=10\) for CSLS, at most 160 landmarks ranked by matching
margin, a minimum of twelve landmarks for recovery, ridge coefficient
\(0.1\), and numerical floors \(10^{-6}\).
This comparison isolates target-side processing on the shared
source-trained base and does not include SCORE's recovery-aware
source retraining.

\paragraph{Robustness to input order.}
The complete K12/40 pipeline is rerun for all thirty EEG2 cases after
independently permuting query and gallery rows with seed 20261111.
After inverse permutation, Table~\ref{tab:supp_permutation_current}
reports numerical errors and true-image rank changes.
Moment statistics differ by at most \(4.44\times10^{-16}\);
initialization maps and centroids by \(6.63\times10^{-9}\).
Cost and transform differences remain below \(10^{-7}\) in 28 of 30 cases;
the remaining two exceed this threshold, with a maximum final-cost
difference of \(3.77\times10^{-3}\).
With gallery-identity tie breaking, GES and CSLS preserve every
true-image rank and all Top-1/Top-5 scores. Direct changes two
true-image ranks, with a maximum per-case accuracy change of
0.5 percentage points. Equation~\eqref{eq:supp_perm} gives the corresponding
algebraic equivariance of the transport operator.

\StagePermutationTable

\section{Mathematical Details and Complete Procedure}
\label{sec:supp-notation}

\paragraph{Notation.}
At test time, the method receives an unlabeled target batch
\(\mathcal{B}=\{b_i\}_{i=1}^{N}\) and a candidate gallery
\(\mathcal{V}=\{v_j\}_{j=1}^{M}\), where the gallery is the image set of the
target dataset~\cite{gifford2022things}, and two frozen encoders map them to
\begin{equation}
Z_b=f_b(\mathcal{B})\in\mathbb{R}^{N\times d},
\qquad
Z_v=f_v(\mathcal{V})\in\mathbb{R}^{M\times d}.
\label{eq:supp_encoders}
\end{equation}
Table~\ref{tab:supp_notation} lists the symbols used throughout the paper and
this document.

\begin{table*}[t]
\centering
\small
\setlength{\tabcolsep}{4pt}
\caption{\textbf{Notation and definitions.}}
\label{tab:supp_notation}
\begin{tabular*}{\textwidth}{@{\extracolsep{\fill}}llll@{}}
\toprule
Symbol & Meaning & Symbol & Meaning\\
\midrule
\(N,M\) & number of queries and candidates (200 each in the closed set)
& \(T,T^{(k)},T^\star\) & teacher, initialization, and final couplings\\
\(d\) & feature dimension (512)
& \(Z_b^0\) & features after calibration and initialization\\
\(a=\mathbf1_N/N\) & row marginal of every coupling
& \(w_i\) & confidence weight of query \(i\)\\
\(c_j\) & upper bound on the mass a candidate may receive
& \(R\) & orthogonal map fitted at initialization\\
\(\Pi(a,c)\) & feasible set of couplings
& \(\rho\) & ridge coefficient of that fit\\
\(P\) & primal transport variable
& \(\lambda\) & weight of the anchoring term\\
\(C_{ij}\) & cost of matching query \(i\) to candidate \(j\)
& \(K\) & number of initialization rounds\\
\(H(P)\) & negative entropy of \(P\)
& \(r\) & rank of the adaptation subspace\\
\(\varepsilon,\tau\) & entropy coefficient, teacher and student
& \(\kappa_r,\kappa_p,\kappa_c\) & gains of the bounded radial representation\\
\(u_i,\nu_j\) & row and column scaling variables
& \(\sigma_s,\sigma_t\) & learned scale and translation gains\\
\(\alpha,\beta\) & coordinatewise moment scale and offset
& \(\gamma_0,\gamma_1\) & gains of the confidence weight\\
& & \(\varepsilon_m\) & stabilizer of the moment calibration\\
& & \(\delta,\delta_C\) & norm/anchor and cost-scale floors\\
& & \(\mu,\sigma\) & per-dimension mean and standard deviation\\
\(\mathrm{val}_\eta,T_\eta\) & optimal value and optimal coupling
& \(T_\theta\) & current student coupling\\
\(\Phi(Z)\) & row-wise radial map with a set-specific median
& \(E_0=Z_b^0B\) & neural subspace coordinates\\
\(f,g\) & row dual potential and column capacity multiplier
& \(A\) & confidence-weighted pair matrix\\
& & \(D\) & cross covariance\\
\bottomrule
\end{tabular*}

\end{table*}

\paragraph{Radial representation and cost.}
Before a cost is formed, each feature set is passed through a bounded radial
representation that keeps the direction and a bounded relative norm. For a
row \(z\) of \(Z\), let \(\widehat n_z=\max(\|z\|_2,\delta)\) and
\(m_Z=\operatorname{med}_{z\in Z}\widehat n_z\), using the lower median:
\begin{equation}
\phi(z;m_Z)=
\frac{z}{\widehat n_z}\,
\kappa_r\exp\!\left[
\kappa_p\,\operatorname{clip}\!\left(
\log\frac{\widehat n_z}{m_Z},-\kappa_c,\kappa_c\right)\right].
\label{eq:supp_radial}
\end{equation}
For represented features \(X\) and \(Y\), with rows \(x_i\) and \(y_j\),
the pair cost is
\begin{equation}
C_{ij}(X,Y)=
\frac{\|x_i-y_j\|_2^2}
{\max\!\left(\operatorname{med}_{p,q}\|y_p-y_q\|_2^2,\delta_C\right)} .
\label{eq:supp_cost}
\end{equation}
All distances are Euclidean, and the scale is the lower median over all gallery
pairs. It stays fixed with the gallery, and each modality has its own
reference norm. The row-wise map is denoted by \(\Phi(Z)\); its median is
computed once per represented feature set. The numerical guards are
\(\delta=10^{-12}\) and \(\delta_C=10^{-8}\), matching the norm/anchor floor
and gallery-distance floor in the implementation.

\paragraph{The feasible set.}
The couplings used throughout satisfy
\begin{equation}
\begin{aligned}
\Pi(a,c)&=\{P\geq0:\ P\mathbf1_M=a,\ P^\top\mathbf1_N\leq c\},\\
a&=\frac{\mathbf1_N}{N},\qquad c_j=\frac{1}{\min(N,M)},
\end{aligned}
\label{eq:supp_marginals}
\end{equation}
so each query sends \(1/N\); for \(M\geq N\), each candidate receives at
most that amount. When \(M>N\), total capacity exceeds one, so
columns need not have equal mass. This permits candidates to remain
below capacity while retaining soft weights.
Numerical comparisons are reported in Section~\ref{sec:supp-gallery}.
Equation~\eqref{eq:supp_e1control} uses an equal-mass column target
to measure the effect of requiring every candidate to receive a share.

\paragraph{Optimal value and coupling.}
For \(\eta>0\), define the negative entropy and the scalar optimum by
\begin{align}
H(P)&=\sum_{ij}P_{ij}(\log P_{ij}-1),\qquad 0\log0=0,\\
\mathrm{val}_\eta(C)&=\min_{P\in\Pi(a,c)}
  \langle C,P\rangle+\eta H(P),\label{eq:supp_value}\\
T_\eta(C)&=\argmin_{P\in\Pi(a,c)}
  \langle C,P\rangle+\eta H(P).\label{eq:supp_plan}
\end{align}
Here \(\langle C,P\rangle=\sum_{ij}C_{ij}P_{ij}\).
The value is a scalar, whereas \(T_\eta(C)\) is the unique optimal
matrix. The cost favors compatible pairs; the negative entropy spreads
their weights. Initialization, teachers and final ranking use
\(\eta=\varepsilon\); students use \(\eta=\tau>\varepsilon\).

\paragraph{Moment calibration.}
With \(\sigma\) and \(\mu\) the per-dimension standard deviation and mean, the
calibration of the target batch to the gallery is
\begin{equation}
\begin{aligned}
\alpha&=\frac{\sigma(Z_v)+\varepsilon_m}{\sigma(Z_b)+\varepsilon_m},
&
\beta&=\mu(Z_v)-\alpha\odot\mu(Z_b),\\
\bar Z_b&=Z_b\odot\alpha+\beta .
\end{aligned}
\label{eq:supp_moment}
\end{equation}
Statistics are computed over samples, and division is elementwise.

\paragraph{Confidence-weighted soft initialization.}
Set \(Q^{(0)}=\bar Z_b\). At round \(k\), use
\(T=T^{(k)}=T_\varepsilon(C(\Phi(Q^{(k)}),\Phi(Z_v)))\) to
estimate pair weights before refitting the original calibrated features.
The initialization uses the row-conditional transport
\(p_{ij}=T_{ij}/\sum_kT_{ik}\).
Let \(e_i=-\sum_jp_{ij}\log p_{ij}\) measure query uncertainty, and define
\(\widetilde w_i=0.25+0.75\max(0,1-e_i/\log M)\) and
\(w_i=\widetilde w_i/\sum_k\widetilde w_k\).
Lower-entropy queries receive larger confidence weights.
Set the confidence-weighted pair matrix \(A_{ij}=w_iT_{ij}\), compute its weighted neural and image centroids
\(x_c,y_c\), and form
\begin{equation}
D=(\bar Z_b-\mathbf1_Nx_c^\top)^\top
A(Z_v-\mathbf1_My_c^\top).
\end{equation}
For \(U\Sigma V^\top=\operatorname{SVD}(D+\rho\|D\|_2 I)\), the regularized
orthogonal solution is
\begin{equation}
R=UV^\top ,
\label{eq:supp_rotation}
\end{equation}
with \(\rho=0.1\); no determinant correction is applied, so \(R\) is a general
orthogonal map rather than a rotation with \(\det R=+1\).
Here \(A\in\mathbb R^{N\times M}\),
\(\omega=\sum_{ij}A_{ij}\),
\(x_c=\bar Z_b^\top A\mathbf1_M/\omega\), and
\(y_c=Z_v^\top A^\top\mathbf1_N/\omega\).
The identity term is scaled by the covariance spectral norm so that
\(\rho\) controls its strength relative to the fitted pair evidence.
Equivalently, \(R\) maximizes
\(\operatorname{tr}(R^\top D)+\rho\|D\|_2\operatorname{tr}(R)\) over
\(R^\top R=I_d\), where \(\operatorname{tr}\) is the matrix trace.
Apply \(R\) to the original \(\bar Z_b\), using the weighted centroids.
Recompute soft correspondences from the resulting current features and repeat
for the dataset-specific number of rounds. Transport costs during these rounds use the same bounded
radial representation as Eq.~\eqref{eq:supp_radial}.
Confidence weights adjust the initialization fit; adaptation uses the
unweighted teacher--student couplings.

\paragraph{Constrained feature map.}
Cache \(Z_b^0=\Phi(Q^{(K)})\) and \(Z_v^0=\Phi(Z_v)\) after initialization.
Let \(B\in\mathbb R^{d\times r}\) contain the leading right singular
vectors of \([Z_b^0;Z_v^0]\), fixed throughout adaptation, and set
\(E_0=Z_b^0B\). The trainable parameters are
\(\theta=(W,\ell,h)\in\mathbb R^{r\times r}\times\mathbb R\times\mathbb R^d\):
\begin{align}
R_\theta&=\exp(W-W^\top),&
s_\theta&=\exp(\sigma_s\tanh\ell),\nonumber\\
t_\theta&=\sigma_t\tanh h,\label{eq:supp_mapparams}\\
Z_b^\theta&=s_\theta[Z_b^0+E_0(R_\theta-I_r)B^\top]\nonumber\\
&\quad+\mathbf1_Nt_\theta^\top.\label{eq:supp_map}
\end{align}
The matrix exponential rotates only the fixed subspace. The shared scale
and bounded translation complete the similarity transform without shear or
coordinatewise stretching; zero parameters recover \(Z_b^0\).
The image side \(Z_v^0\) and the basis remain fixed.
For the squared-distance cost, translation adds only row- and column-dependent
terms, whose contributions are constant under fixed marginals. It therefore
leaves the closed-set coupling unchanged, although it remains part of the
feature-anchor penalty.

\paragraph{Teacher, student, and anchoring.}
Write \(C_\theta=C(Z_b^\theta,Z_v^0)\).
At each refresh state \(\theta_{\rm ref}\), compute
\(T=T_\varepsilon(C_{\theta_{\rm ref}})\), normalize it to
\(\widehat T=T/\sum_{ij}T_{ij}\), and stop its gradient.
The unit-mass normalization guards numerical mass error.
The current student is \(T_\theta=T_\tau(C_\theta)\).
Their Fenchel--Young gap is
\begin{equation}
F_\tau=\langle C_\theta,\widehat T\rangle+
       \tau H(\widehat T)-\mathrm{val}_\tau(C_\theta).
\label{eq:supp_gap}
\end{equation}
The adaptation objective adds an anchoring term that penalizes the normalized
deviation of the features from their initial state, with weight
\(\lambda=0.15\):
\begin{equation}
\mathcal{L}=\frac{F_\tau}{\tau}
+\lambda\,
\frac{\frac1N\sum_i\|Z_{b,i}^\theta-Z_{b,i}^0\|_2^2}
{\max(\frac1N\sum_i\|Z_{b,i}^0\|_2^2,\delta)} .
\label{eq:supp_objective}
\end{equation}
The anchor denominator uses \(\delta=10^{-12}\).
Teacher and student must belong to the same feasible set for the gap to
compare feasible couplings. At a refresh, the detached teacher is computed from
\(C_{\theta_{\mathrm{ref}}}\); it remains fixed while the student is
recomputed from the current \(C_\theta\). The teacher supplies a stable
feasible target between refreshes. Main gradient adaptation uses the square
closed-set problem.

\paragraph{One transport construction, distinct uses.}
Table~\ref{tab:supp_calls} summarizes the shared operator. Initialization
recomputes represented neural features to estimate correspondences, then fits
a full-dimensional orthogonal map in the original moment-calibrated space.
Gradient adaptation operates on the cached radial features with a rank-16
subspace rotation, scale, and translation. Ranking reuses the final
represented features and changes the coupling alone.

\paragraph{Complete procedure.}
Given unlabeled embeddings, a budget \(K,L\), and refresh interval
\(K_{\rm ref}\), the computation is:
\begin{enumerate}
\item Compute \(\alpha,\beta,\bar Z_b\) by
Eq.~\eqref{eq:supp_moment}; set \(Q^{(0)}=\bar Z_b\).
\item For \(k=0,\ldots,K-1\), solve
\(T^{(k)}=T_\varepsilon(C(\Phi(Q^{(k)}),\Phi(Z_v)))\).
Compute row confidence, centroids and \(D\); obtain \(R\) by
Eq.~\eqref{eq:supp_rotation} and set
\(Q^{(k+1)}=(\bar Z_b-\mathbf1_Nx_c^\top)R+\mathbf1_Ny_c^\top\).
Every round refits the original calibrated features.
\item Cache \(Z_b^0,Z_v^0,B,E_0\) and initialize \(\theta=0\).
For \(t=0,\ldots,L-1\), refresh and detach \(\widehat T\) when
\(t\bmod K_{\rm ref}=0\); solve \(T_\theta\) at the current cost.
Backpropagate \((\widehat T-T_\theta)/\tau\) and the anchor through
the feature map, clip the parameter gradient and take one Adam step.
\item Use the state after update \(L\). Hold its features fixed, compute
\(T^\star=T_\varepsilon(C(Z_b^\star,Z_v^0))\), and rank each row by
\(\log\max(T^\star_{ij},10^{-300})\).
\end{enumerate}

\begin{table}[t]
\centering
\small
\caption{\textbf{Transport solver evaluations for EEG2.}
The teacher is refreshed at steps \(0,10,20,30\); each student solve
contributes to one feature update. MEG uses two initialization solves,
eight student solves and one teacher refresh; Alljoined uses five
initialization solves, thirteen student solves and two teacher refreshes.}
\label{tab:supp_calls}
\begin{tabular*}{\columnwidth}{@{\extracolsep{\fill}}lrrl@{}}
\toprule
Use & Solves & Coefficient & Updated object\\
\midrule
Initialization & 12 & \(\varepsilon=0.035\) & Procrustes map\\
Teacher refresh & 4 & \(\varepsilon=0.035\) & Detached target\\
Student & 40 & \(\tau=0.070\) & Feature gradient\\
Final ranking & 1 & \(\varepsilon=0.035\) & Retrieval scores\\
\bottomrule
\end{tabular*}

\end{table}

\begin{figure*}[t]
\centering
\includegraphics[width=\textwidth]{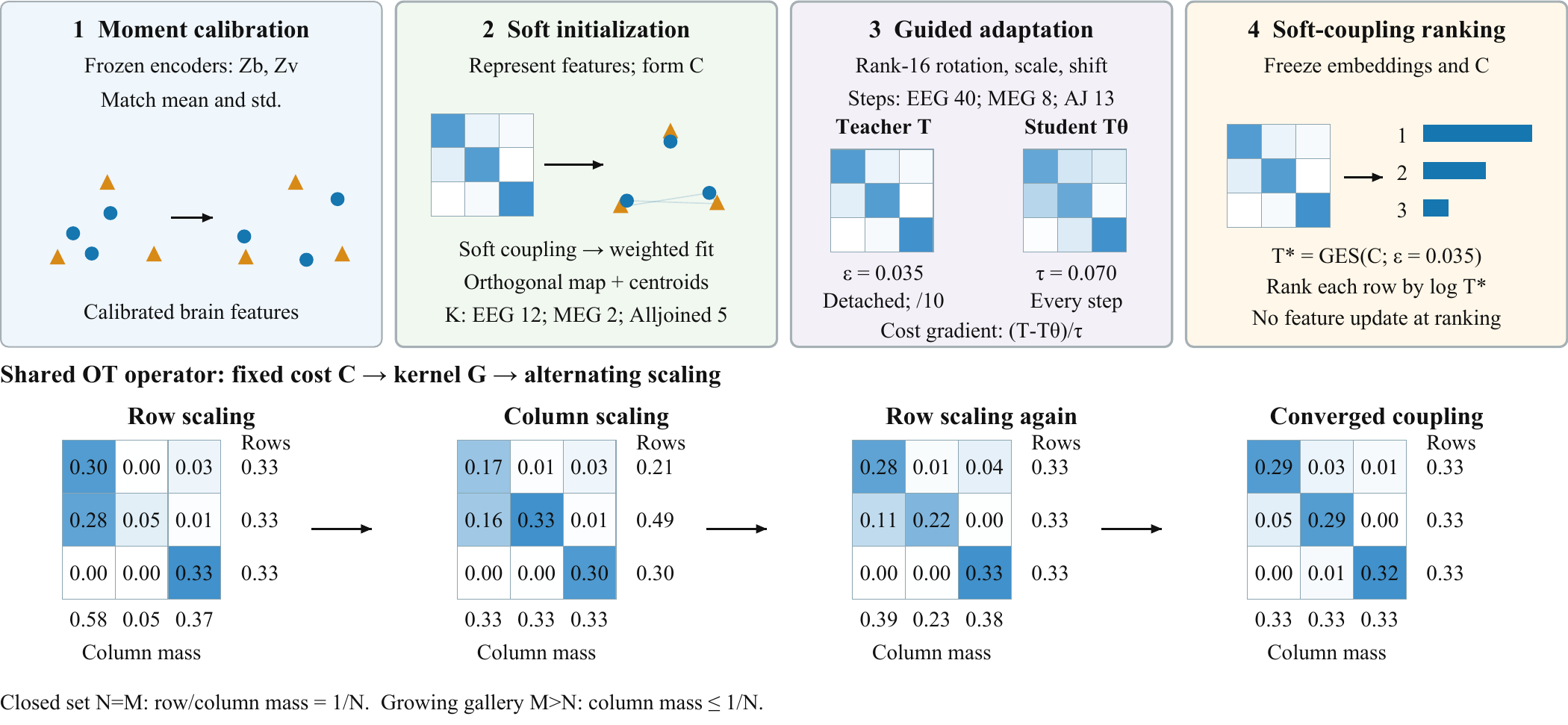}
\caption{\textbf{Coupling-based alignment, adaptation, and ranking.}
Top: the EEG2 route uses moment calibration, twelve rounds of
coupling-weighted initialization, 40 steps of query-only subspace adaptation,
and fixed-feature ranking. MEG uses two initialization rounds and eight
adaptation steps; Alljoined uses five rounds and thirteen steps.
At each refresh, teacher and student use the same cost at coefficients
0.035 and 0.070; between refreshes the teacher remains fixed while the
student follows the current cost. The teacher is detached and refreshed
every ten steps. The cost gradient is
\((\widehat T-T_\theta)/\tau\), with a separate feature-anchoring term.
Bottom: alternating scaling of one fixed synthetic \(3\times3\) cost.
Candidate 1 initially receives mass 0.58; convergence gives row and column
mass \(1/3\) and changes query 2's top-ranked image from 1 to 2.
The closed-set example illustrates soft mass reallocation; for \(M>N\),
the column constraint is a capacity. Cell values and marginals are rounded.}
\label{fig:supp_mechanism}
\end{figure*}

\paragraph{The fixed synthetic matching example.}
Fig.~\ref{fig:supp_mechanism} uses image coordinates
\((0,0),(1,0),(0.5,0.866)\) and query coordinates
\((0.4,0.3),(0.47,0.23),(0.52,0.45)\).
Squared distances use the gallery-pair lower-median normalization and
\(\varepsilon=0.035\).
Independent row normalization yields row-wise argmax \((1,1,3)\);
the balanced coupling yields \((1,2,3)\).
The four scaling panels keep these coordinates and costs fixed.
The upper row summarizes the test-time sequence; its point sketches show the
roles of neural features and fixed visual anchors rather than fitted dataset
coordinates.

\paragraph{Numerical implementation.}
Table~\ref{tab:supp_controls} lists the representation, optimization and
solver settings used here. Learned
target updates use float64 feature computations; the final ranking uses
\(\log\max(T,10^{-300})\); marginal checks accompany the stable
transport solve. Independent standard-SN checks use a \(10^{-10}\)
log-marginal residual threshold.
The base encoders are frozen NICE models~\cite{song2024nice},
trained under the source protocol of Jiang et al.~\cite{jiang2026samga}.
The envelope expression differentiates the optimal-value objective
directly with respect to its costs.

\paragraph{How the coupling is computed.}
The coupling solves the entropy-dual problem in
Section~\ref{sec:supp-dual}. The balanced implementation runs alternating scaling
following
Cuturi~\cite{cuturi2013sinkhorn} in the log domain. If its numerical checks
require a restart, it solves the same dual with L-BFGS-B and, if necessary,
Newton refinement. Capacity ranking uses the bound-constrained entropy dual.

\begin{table*}[t]
\centering
\small
\setlength{\tabcolsep}{4pt}
\caption{\textbf{Representation, optimization, and solver settings.}
These are defaults; individual controls specify changed values and stage
switches. Constants that carry
a symbol in the paper are listed with it; the remaining rows are named by
their role.}
\label{tab:supp_controls}
\begin{tabular*}{\textwidth}{@{\extracolsep{\fill}}llc@{}}
\toprule
Symbol & Meaning & Value\\
\midrule
\(K\) & initialization rounds (EEG2 / MEG / Alljoined) & 12 / 2 / 5\\
\(\varepsilon\) & entropy coefficient, teacher/ranking & 0.035\\
\(\tau\) & entropy coefficient, student & 0.07\\
\(r\) & rank of the adaptation subspace & 16\\
\(\lambda\) & anchoring weight & 0.15\\
\(\rho\) & ridge of the initialization fit & 0.1\\
\(\kappa_r,\kappa_p,\kappa_c\) & radial representation gains & 0.5, 0.25, 1.0\\
\(\sigma_s,\sigma_t\) & learned scale and translation gains & 0.4, 0.1\\
\(\gamma_0,\gamma_1\) & confidence-weight gains & 0.25, 0.75\\
\(\varepsilon_m\) & calibration stabilizer & \(10^{-6}\)\\
adaptation steps & feature updates (EEG2 / MEG / Alljoined) & 40 / 8 / 13\\
learning rate & step size of feature updates & 0.01\\
TTA seed & feature-adaptation seed, distinct from encoder seed & 1701\\
Adam & first/second moment coefficients; optimizer stabilizer &
\((0.9,0.999);\ 10^{-8}\)\\
gradient clipping & maximum parameter-gradient norm & 1.0\\
refresh interval & steps between teacher refreshes & 10\\
\(\delta,\delta_C\) & norm/anchor floor and cost-scale floor & \(10^{-12},10^{-8}\)\\
ranking floor & floor on coupling entries before \(\log\) & \(10^{-300}\)\\
SN tolerance & log-marginal residual threshold & \(10^{-10}\)\\
main solve & solver tolerance; marginal tolerance & \(10^{-8};\ 2\times10^{-7}\)\\
student solve & solver tolerance; marginal tolerance & \(10^{-12};\ 2\times10^{-10}\)\\
iteration limits & main scaling; SN comparison & 20,000; 500,000\\
dual fallback & L-BFGS-B iterations; Newton steps & 5,000; 80\\
precision & dtype of the learned target updates & float64\\
solver & balanced / capacity numerical route & log scaling / entropy dual\\
cost scale & scale of Eq.~\eqref{eq:supp_cost} & lower median of gallery pairs\\
\bottomrule
\end{tabular*}

\end{table*}

\subsection{Optimality, Dual Form, and Gradient Identity}
\label{sec:supp-dual}

\paragraph{Dual form.}
For finite costs, positive row masses and column capacities, a feasible set,
and \(\eta>0\), the entropic objective has a unique positive minimizer.
Use \(f\in\mathbb{R}^{N}\) for the free row potentials and
\(g\in\mathbb{R}_{+}^{M}\) for the column-capacity multipliers. These symbols
are distinct from the moment-calibration scale \(\alpha\) and offset \(\beta\).
The Lagrangian is
\begin{align}
\mathcal{J}(P,f,g)
={}&\langle C,P\rangle+\eta H(P) \nonumber\\
&+f^\top(a-P\mathbf1_M)
+g^\top(P^\top\mathbf1_N-c).
\end{align}
Stationarity gives
\(C_{ij}+\eta\log P_{ij}-f_i+g_j=0\), hence
\begin{equation}
\begin{aligned}
u_i&=\exp(f_i/\eta), & \nu_j&=\exp(-g_j/\eta),\\
P_{ij}&=u_i\exp(-C_{ij}/\eta)\nu_j .
\end{aligned}
\label{eq:supp_dualscaling}
\end{equation}
The nonnegative multiplier gives \(0<\nu_j\leq1\).
Complementary slackness requires
\(g_j(c_j-\sum_iP_{ij})=0\): an inactive column has \(g_j=0\) and
\(\nu_j=1\), while a positive multiplier makes its bound tight.
The row constraint and column optimality condition therefore give the
capacity-aware scaling
\begin{equation}
u\leftarrow a\oslash(G\nu),\qquad
\nu\leftarrow\min\{\mathbf1_M,c\oslash(G^\top u)\},
\label{eq:supp_scaling}
\end{equation}
where \(G_{ij}=\exp(-C_{ij}/\eta)\) and division and minimum are
elementwise.
Equivalently, maximize the concave dual
\[
f^\top a-g^\top c-\eta\sum_{ij}
\exp\bigl((f_i-g_j-C_{ij})/\eta\bigr),\qquad g\geq0.
\]
Eliminating \(f\) enforces row mass exactly and yields the column-potential
objective used by the capacity solver.

\paragraph{Scaling variables and their rescaling.}
The primal variable is \(P\); \(u,\nu\) parameterize its optimum and must be
solved for. Algebraically, \((u,\nu)\mapsto(su,\nu/s)\) for \(s>0\) leaves
the represented matrix unchanged. With equality marginals this is the usual
dual gauge. With capacity constraints, the factors must additionally satisfy
\(\nu_j\leq1\) and complementary slackness, so an inactive column fixes its
factor at one. Arbitrary rescaling need not preserve dual optimality.
Equation~\eqref{eq:supp_dualscaling} represents the unique optimizer
through scaling factors determined iteratively from the constraints.

\paragraph{Optimality representation and the ranking rule.}
At the optimum \(P=T_\eta(C)\), Equation~\eqref{eq:supp_dualscaling} implies that the optimal coupling
solves
\(P_{ij}=\exp((f_i-g_j-C_{ij})/\eta)\).
Its log is a pairwise cost term plus a query potential and minus a shared
candidate-capacity potential. Final scores use
\begin{equation}
S_{ij}=\log T^\star_{ij}
=-C^\star_{ij}/\varepsilon+\log u_i+\log\nu_j .
\label{eq:supp_ranking}
\end{equation}
The row term \(\log u_i\) is constant within a row
and cannot change the ordering; the shared candidate term \(\log\nu_j\) is
determined by all queries through the column constraint. The coupling itself is
the retrieval score.

\paragraph{Why coupling matching gives the stated gradient.}
Let \(T_\theta=T_\tau(C_\theta)\) and hold the normalized teacher
\(\widehat T\in\Pi(a,c)\) fixed. The envelope theorem gives
\(\partial\mathrm{val}_\tau(C_\theta)/\partial C_\theta=T_\theta\).
The remaining cost-dependent term differentiates to \(\widehat T\), yielding
\((\widehat T-T_\theta)/\tau\) for the scaled gap in
Eq.~\eqref{eq:supp_objective}. The identity holds for each fixed feasible
teacher target between refreshes.

The implementation supplies the envelope cost
gradient and backpropagates through the feature map; it does not unroll
Sinkhorn or differentiate the detached teacher.

\paragraph{Permutation invariance.}
For row and column permutation matrices \(J_N,J_M\), uniqueness gives
\begin{equation}
T_\eta(J_NCJ_M^\top)=J_NT_\eta(C)J_M^\top ,
\label{eq:supp_perm}
\end{equation}
so permuting the inputs relabels the optimal coupling: the transport operator
is equivariant. A relabeling of the gallery relabels the coupling.
For the weighted Procrustes sufficient statistics, simultaneous relabeling
of features, coupling and row weights cancels in centroids and cross covariance.
Subspace degeneracy, finite solver tolerances and score ties also affect
a concrete implementation. Table~\ref{tab:supp_permutation_current}
reports the numerical input-order comparison of the complete pipeline.
This comparison resolves ties by gallery identity after inverse permutation;
displayed retrieval cases use stable gallery-index tie breaking and
identity-based truth matching.

\paragraph{Uniform marginals and hard predictions.}
For total mass one, the negative entropy used here satisfies \(H(P)\leq-1\);
spreading mass makes it more negative. Uniform marginals prescribe row and
column mass, not paired identities and not distinct row-wise maxima. In the
square uniform case, an unregularized transport problem admits an optimal coupling
equal to a permutation matrix divided by \(N\), and degeneracies can admit
other optimal couplings. The Hungarian assignment therefore provides a hard Top-1
control, while the positive-entropy solution provides full candidate rankings.

\FloatBarrier
\section{Mechanism Configurations and Subject Variation}
\label{sec:supp-results}

This section separates feature improvements from the final retrieval
decision, then reports subject and source-seed variation under the
protocols of Section~\ref{sec:supp-protocol}.

\StageDatasetTable
\StageProgressiveTable

\paragraph{Feature changes and the final decision.}
Table~\ref{tab:supp_ladder} separates successive feature stages from the
final ranking. On EEG2, soft initialization raises Direct Top-1 from
30.80\% after calibration to 53.03\%; adaptation raises it to 58.63\%.
Keeping those final features and costs fixed, the coupled ranking
reaches 68.25\%/90.47\% Top-1/Top-5.
The final same-cost results for all three datasets are given in
Table~\ref{tab:supp_datasets}.
Table~\ref{tab:supp_ladder} includes the encoder's cosine and
unnormalized dot-product scores as unadapted references.
For a comparison within the method's cost definition, the unadapted
radial baseline is 26.80\%/58.45\% on EEG2. Its change to
30.80\%/63.10\% after moment calibration compares the same radial
distance rule before and after calibration.

\StageSubjectTable
\StageSeedTable
\StageMegSubjectTable
\StageAlljoinedSubjectTable
\StageSubjectSpreadTable

\paragraph{Consistency across subjects.}
At the final EEG2 cost, coupling improves Top-1 over Direct for all ten
subjects, by a mean of 9.62 percentage points
(descriptive interval [7.03, 12.05]).
Against same-cost CSLS the gain is 5.10 points
([3.58, 6.62]), also positive for all ten subjects.
Tables~\ref{tab:supp_subject}--\ref{tab:supp_spread} give all subject
scores, source-seed means and between-subject standard deviations.
MEG and Alljoined subject differences are reported separately.

\FloatBarrier
\section{Correspondences for Geometric Initialization}
\label{sec:supp-landmark}

\begin{figure*}[!tbp]
\centering
\includegraphics[width=\textwidth]{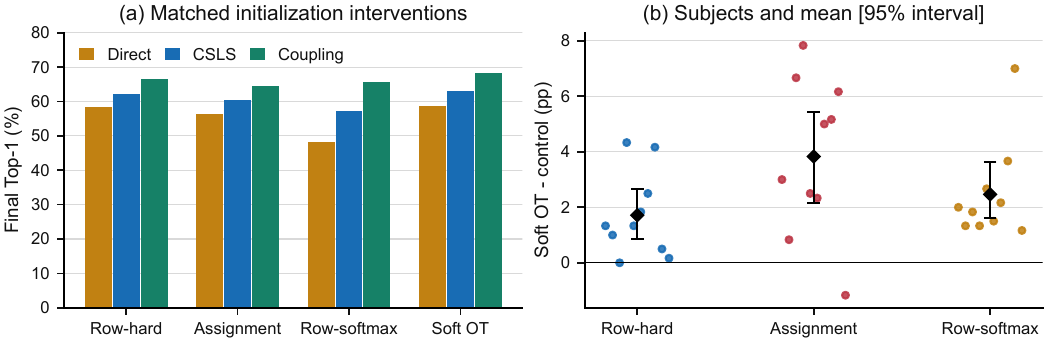}
\caption{\textbf{Effects of soft correspondences on geometric alignment.}
(a) Final Top-1 for the three ranking rules after changing initialization
correspondences, keeping reference row-confidence weights, estimator
and later optimization fixed. (b) Soft-OT minus control Top-1:
dots are subjects after averaging source seeds; diamonds and whiskers
are means and descriptive 95\% subject-bootstrap intervals.
All arms evaluate the complete 200-candidate gallery.}
\label{fig:supp_matched_initialization}
\end{figure*}

\paragraph{Matched correspondence constructions.}
Table~\ref{tab:supp_matched_init} changes only the correspondence
construction within the regularized Procrustes initialization.
The alternatives are row-wise hard matches, a global hard assignment,
and dense row-softmax weights, compared with entropic soft coupling.
Every arm uses twelve rounds, the same original calibrated features,
the same ridge, the same reference query-confidence weights at each
round, and the same subsequent 40-update adaptation.
At each round, all variants use the query-confidence weights of the
soft-coupling reference, isolating the effect of the correspondence matrix.

\StageMatchedTable
\StagePairedTable

\paragraph{Result of retaining soft correspondence evidence.}
Final coupled Top-1 is 68.25\% with soft OT, versus 66.53\% with row-hard,
64.42\% with hard assignment and 65.78\% with dense row-softmax.
Table~\ref{tab:supp_matched_paired} and
Fig.~\ref{fig:supp_matched_initialization} show subject-level differences.
Soft OT exceeds row-hard for nine subjects with one tie, exceeds
assignment for nine subjects, and exceeds row-softmax for all ten.
The comparison isolates the correspondence construction within this
shared estimator and update budget.

\paragraph{Additional confidence weighting.}
Table~\ref{tab:supp_r01weights} retains the full coupling and varies only
the extra query-confidence weights. Uniform weights give 68.02\% Top-1,
compared with 68.25\% for coupling-entropy confidence.
This small difference distinguishes confidence weighting from the
larger intervention on the correspondence matrix itself.

\StageWeightsTable

\FloatBarrier
\section{Feature Adaptation and Sensitivity}
\label{sec:supp-adaptation}

\paragraph{Initialization and optimization budget.}
Table~\ref{tab:supp_noinit} compares complete update budgets with and
without soft initialization, at the same learning rate 0.01.
With K12, coupled Top-1 is 65.68\% before adaptation and 68.25\% after
40 updates. Without initialization, it is 48.08\% before updates,
57.72\% after 40, and 59.42\% after 200.
Initialization provides a stronger starting geometry, with K12/40
remaining ahead of the tested initialization-free budgets.

\StageUpdateTable
\StageObjectivesTable
\StageWeightedObjectivesTable

\paragraph{Objective comparisons.}
Tables~\ref{tab:supp_objectives_nice}
and~\ref{tab:supp_weighted_objectives} separate two target constructions.
All arms retain K12 initialization, the same constrained feature map,
learning rate, anchor, refresh interval, and final coupled ranking.
Let \(Q_{ij}=\exp(-C_{ij}/\tau)/\sum_k\exp(-C_{ik}/\tau)\).
The first group uses the same construction of the detached unit-mass
teacher \(\widehat T\), without additional confidence weights:
\begin{align}
L_{\rm attract}&=\langle C_\theta,\widehat T\rangle/\tau,\nonumber\\
L_{\rm row}&=-\sum_{ij}\widehat T_{ij}\log Q_{ij}.
\label{eq:supp_students}
\end{align}
All objectives include the same feature anchor. Scaled attraction
removes the student term from the coupling-gap gradient, leaving
\(\widehat T/\tau\); this keeps its attraction coefficient matched to
\((\widehat T-T_\theta)/\tau\).
Row cross-entropy enforces row normalization only.
Coupling matching uses the envelope cost gradient.
Scaled attraction gives 65.03\%/88.55\%, row cross-entropy gives
67.75\%/89.47\%, and coupling matching gives 68.25\%/90.47\%.
Thus retaining the student-dependent term improves on attraction alone
under this shared optimization budget.
Teachers are refreshed from each arm's current features: their construction
is shared, while their numerical values can diverge across trajectories.

To examine teacher confidence weighting, at each refresh we form
\(\widetilde T_{ij}=w_iT_{ij}/\sum_{pk}w_pT_{pk}\) using the initialization
confidence weights and compare the following objectives:
\begin{equation}
L_{\rm attraction}=\langle C_\theta,\widetilde T\rangle,
\qquad
L_{\rm NCE}=-\sum_{ij}\widetilde T_{ij}\log Q_{ij}.
\label{eq:supp_weightedlosses}
\end{equation}
The student temperature is \(\tau=0.07\), with 40 or 80 updates as indicated.
Reweighting changes the teacher marginals, so these comparisons measure
the combined effect of teacher weighting and the objective.

\begin{figure*}[!t]
\centering
\includegraphics[width=\textwidth]{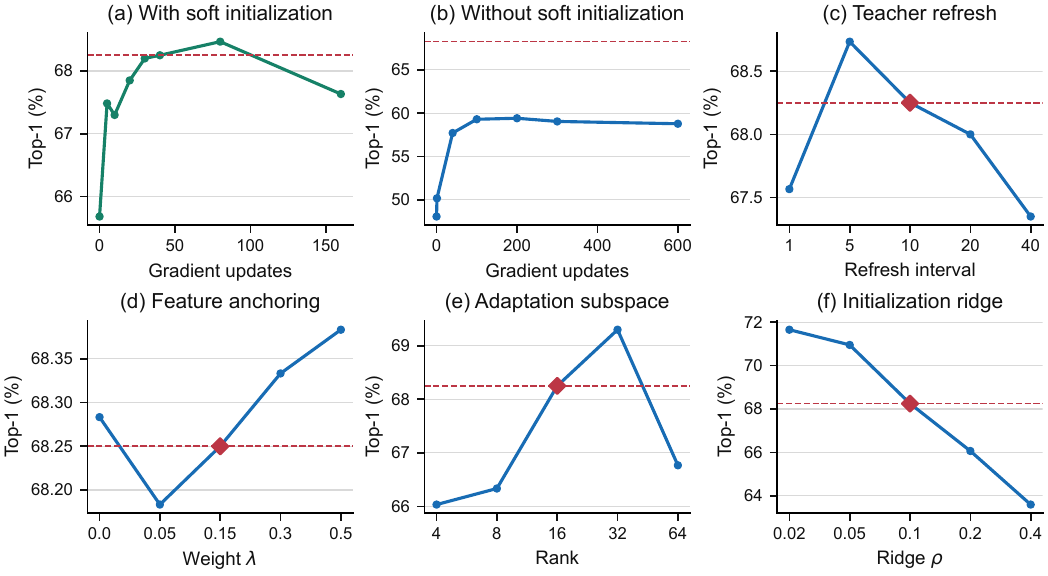}
\caption{\textbf{NICE EEG2 optimization and configuration sensitivity.}
(a,b) Sensitivity to adaptation steps with K12 and K0 initialization,
respectively. (c--f) Sensitivity to teacher refresh interval, feature-anchor
weight, subspace rank and initialization ridge.
The dashed line marks the reported 68.25\% reference; red diamonds
mark its setting in each sweep. Every point averages the same thirty
subject--seed cases.}
\label{fig:supp_sensitivity}
\end{figure*}

\paragraph{Refresh, anchoring and transform size.}
Table~\ref{tab:supp_sensitivity} and Fig.~\ref{fig:supp_sensitivity}
show one-factor changes around the reference configuration.
Refresh intervals from one to forty updates yield 67.35--68.73\%
Top-1. Anchor weights from zero to 0.5 yield 68.18--68.38\%.
Rank and initialization ridge produce larger changes.
The reference uses K12/40, rank 16 and ridge 0.1.

\StageSensitivityTable
\StageExtraSensitivityTable

Table~\ref{tab:supp_extra_sensitivity} gives the complementary learning-rate,
student-temperature, and initialization-budget sweeps, with their reference
settings marked explicitly. These compare feature fitting; the entropy
scan in Section~\ref{sec:supp-fixed} changes only the final ranking.

\FloatBarrier
\section{Candidate Competition at Fixed Features}
\label{sec:supp-fixed}

\begin{figure*}[!tbp]
\centering
\includegraphics[width=\textwidth]{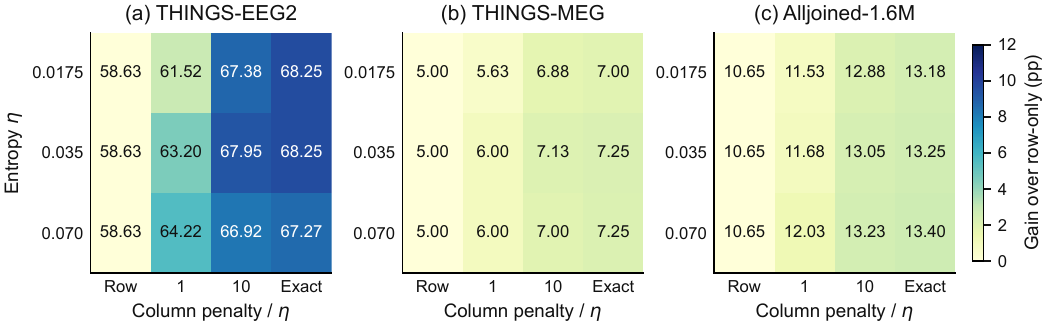}
\caption{\textbf{Effects of column constraints and entropy at fixed costs.}
Each cell prints Top-1 (\%). Color gives the gain over the row-only
cell at the same entropy, in percentage points. Columns vary only
the penalty ratio in Eq.~\eqref{eq:supp_e1control}; Exact denotes
uniform column balance. All costs use the stated dataset-specific budgets.}
\label{fig:supp_constraint_entropy}
\end{figure*}

\paragraph{Constraint and entropy as separate controls.}
Every comparison in this section freezes the final feature map and
pairwise cost. Let \(q=P^\top\mathbf1_N\), \(b=\mathbf1_M/M\), and solve
\begin{equation}
\begin{split}
\min_{P\geq0,\;P\mathbf1_M=a}\ 
&\langle C,P\rangle+\eta H(P)\\
&+\xi\,\operatorname{KL}(q\|b).
\end{split}
\label{eq:supp_e1control}
\end{equation}
Both \(q\) and \(b\) have total mass one, so
\(\operatorname{KL}(q\|b)=\sum_jq_j\log(q_j/b_j)\).
The entropy coefficient is \(\eta\); \(\xi/\eta\) controls the column
penalty. Ratio zero gives row-only normalization, ratios one and ten
penalize imbalance, and the infinite-ratio limit enforces \(q=b\).
Table~\ref{tab:supp_e1grid} and
Fig.~\ref{fig:supp_constraint_entropy} compare three entropy coefficients
and four column-penalty settings across the three datasets.

\StageGridTable

\paragraph{Corrections and introduced errors.}
Table~\ref{tab:supp_correction_transitions} includes every correctness
transition. Across the 6,000 repeated EEG2 query--seed evaluations,
GES corrects 795 Direct errors and introduces 218, giving a net
9.62-point gain. Against CSLS it corrects 495 and introduces 189.
Counts cover ten subjects, each evaluated with three source seeds;
subject-level comparisons average these seeds first.

\StageTransitionsTable

\begin{figure*}[!tbp]
\centering
\includegraphics[width=\textwidth]{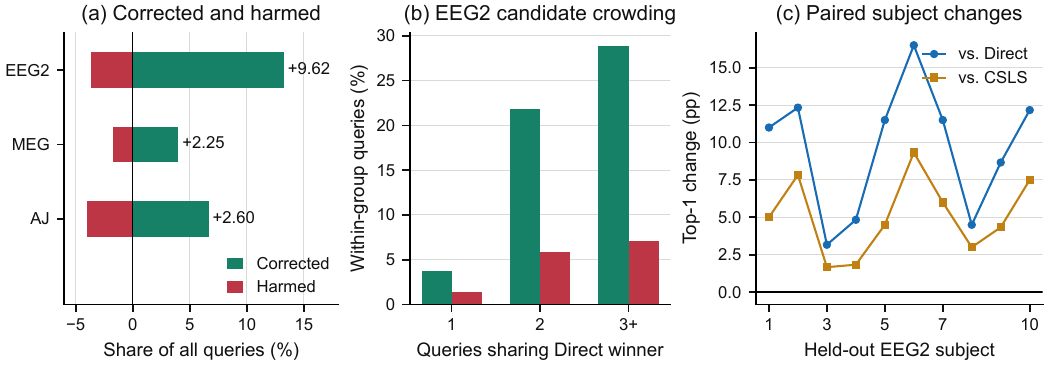}
\caption{\textbf{Prediction changes under joint coupling ranking.}
(a) Corrected and harmed predictions relative to Direct, as percentages
of all query evaluations; labels give net Top-1 changes.
(b) EEG2 changes within groups whose Direct candidate attracts one,
two, or at least three queries. (c) EEG2 subject changes after averaging
three source seeds. Features and costs remain fixed.}
\label{fig:supp_correction_mechanism}
\end{figure*}

\paragraph{Coverage and crowding.}
Write \(n_j=\#\{i:j=\argmax_k(-C_{ik})\}\).
Argmax coverage is \(\#\{j:n_j>0\}/M\), and the crowded-query rate is
\(\#\{i:n_{\argmax_k(-C_{ik})}\geq2\}/N\).
Fig.~\ref{fig:supp_correction_mechanism} groups queries by
their Direct winner's \(n_j=1,2,\text{or}\ \geq3\).
Panel (a) divides transitions by all queries; panel (b) divides each
transition count by the queries within that crowding group.
Panel (c) averages source seeds within subjects.

\paragraph{Related normalizations.}
Table~\ref{tab:supp_related_normalizations} compares Direct, CSLS,
batch-local inverted softmax (IS), and GES on identical costs.
IS estimates its column normalizers from
the current query batch at inverse temperature 20 or \(1/\varepsilon\).
For \(G_{ij}=\exp(-C_{ij}/\varepsilon)\), column-only normalization ranks by
\begin{equation}
S^{\rm col}_{ij}=-C_{ij}/\varepsilon
-\log\sum_k\exp(-C_{kj}/\varepsilon).
\label{eq:supp_columnonly}
\end{equation}
The matched-temperature batch IS control gives 63.17\%/88.30\% on EEG2.
Row-only normalization preserves Direct rankings because its
normalizer is constant within each query.

\StageNormalizationsTable

\paragraph{The hard endpoint and transport cost.}
Table~\ref{tab:supp_r03eps} varies entropy at fixed costs.
Hard assignment gives 67.32\% Top-1; \(\eta=0.035\) gives
68.25\%/90.47\% Top-1/Top-5. Zero scores tie all non-assigned
candidates under hard assignment, so its Top-5 ordering is undefined.
Its unit-mass coupling cost is 0.4915 versus 0.5081 for the soft coupling.
The soft coupling's mean row-winner cost, 0.4854, measures a different
object: row-wise winners need not be a feasible balanced assignment.

\StageEntropyTable

\paragraph{Preferred edges without deleting alternatives.}
For a preferred set \(\mathcal E\), use the reweighted kernel
\begin{equation}
\widetilde G_{ij}=\exp(-C_{ij}/\eta)
\begin{cases}
1,&(i,j)\in\mathcal E,\\
10^{-6},&(i,j)\notin\mathcal E .
\end{cases}
\label{eq:supp_e2kernel}
\end{equation}
Preferred sets are MNN pairs or unions of query-to-gallery and
gallery-to-query top-\(k\) neighborhoods.
Every edge stays positive and all candidates are ranked.
The MNN-preferred condition gives 63.22\%/83.65\%, with about
47\% of mass still outside its preferred edges
(Table~\ref{tab:supp_r11support}).
Wider unions approach the full-kernel 68.25\%/90.47\%.
This evaluates access to alternative edges at fixed costs, separately
from the SCORE geometric fitting pipeline.

\StageSupportTable

\FloatBarrier
\section{Frozen Geometry, Disjoint Queries, and Larger Galleries}
\label{sec:supp-gallery}

\paragraph{Expanding only the ranking gallery.}
The first experiment fits geometry on the complete 200-query,
200-candidate batch, freezes the transform, and adds distractors only
at ranking. Table~\ref{tab:supp_gallery} reports Direct, CSLS,
equal-mass coupling and capacity coupling.
At 1,854 candidates, capacity gives 42.05\% Top-1, compared with
41.55\% for Direct and 29.18\% for equal mass.
At the largest size, capacity ranking improves on Direct by 0.50 points.
Its descriptive subject-bootstrap interval is [0.12, 0.87] points.
The fitted transform and query set stay fixed. The represented image
gallery and its lower-median distance scale are recomputed for each
gallery size.

\StageGalleryTable
\StageGalleryMassTable

\paragraph{What the column rule reallocates.}
Table~\ref{tab:supp_gallerydiag} measures mass on true pairs,
Direct-selected winner edges and candidates without any Direct winner.
At 1,854 candidates, equal columns retain 5.61\% of mass on true pairs,
whereas capacities retain 27.30\%.
Unvisited candidates receive no Direct top-1 votes; the entropic coupling
retains positive mass on them.
These measurements explain why forcing equal mass onto all distractors
differs from allowing columns to remain below capacity.

\paragraph{Complete-map transfer to different query identities.}
For each EEG2 case, a deterministic identity-sorted alternating split
uses 100 identities for unlabeled calibration and the other 100 for
retrieval. The forward map freezes the moment scale and offset,
initialization rotation and centroid translation, radial reference
norms, and learned residual transform.
Neither retrieval EEG features nor their pair labels enter this fitting.
Ranking still uses the retrieval query batch jointly.
Calibration images are excluded from the retrieval gallery; its largest
size is therefore \(100+1654=1754\).
This evaluates identity-disjoint transfer within a subject.

\StageTransferTable

\paragraph{Retrieval with the frozen complete map.}
The complete map reaches 75.47\% Top-1 on the disjoint 100-candidate
retrieval set and 25.80\% with capacity ranking at 1,754 candidates
(Table~\ref{tab:supp_r04split}).
These results support reuse of the calibrated map for query identities
excluded from fitting.
At 1,754 candidates, the full-map capacity advantage over equal-mass
ranking is 10.23 points ([7.73, 13.03]).

\begin{figure*}[!t]
\centering
\includegraphics[width=\textwidth]{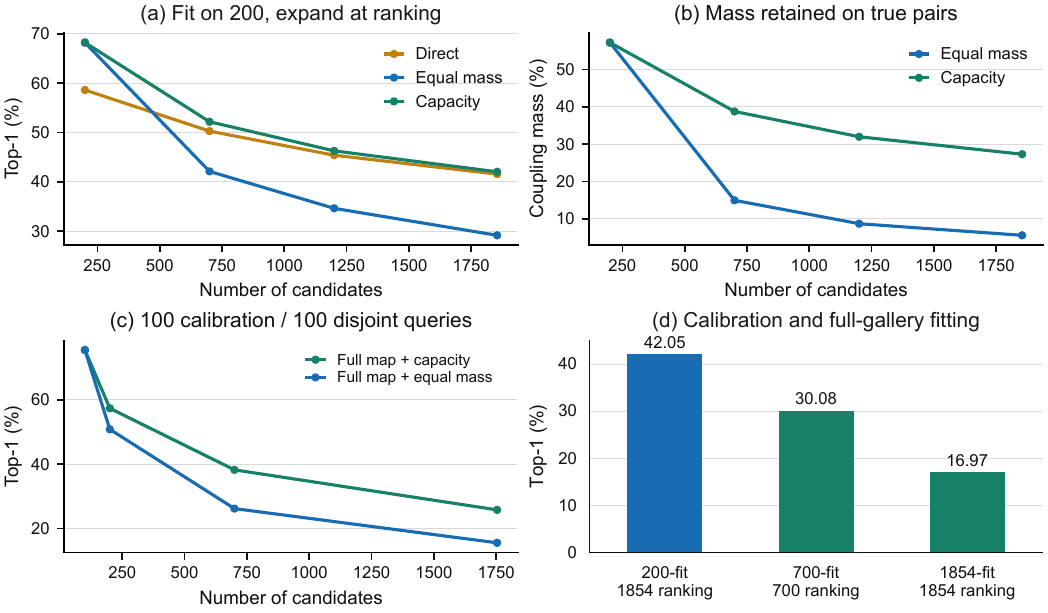}
\caption{\textbf{Gallery expansion and transfer to disjoint queries.}
(a,b) Fit geometry on 200 queries/200 images, then expand only the ranking gallery:
Top-1 and mass retained on true pairs.
(c) Fit on 100 identities and transfer the complete frozen map to
100 different query identities, using equal-mass or capacity ranking;
calibration images are excluded,
giving maximum gallery size 1,754.
(d) The frozen 200-fit geometry evaluated at 1,854 is compared with
full-gallery capacity fitting at 700 and 1,854.
All three bars use capacity ranking.
Every point or bar uses all thirty NICE EEG2 cases.}
\label{fig:supp_gallery_transfer}
\end{figure*}

\paragraph{Fitting with the full expanded gallery.}
Table~\ref{tab:supp_fullgallery_matched} instead starts every fitting
stage from the full 700- or 1,854-candidate gallery.
STAGE reaches 30.08\% and 16.97\% Top-1, compared with SCORE's
27.83\% and 17.02\%, respectively.
The 42.05\% result instead fits geometry on the matched 200-image
calibration gallery, freezes it, and expands only the ranking gallery.
It therefore evaluates gallery expansion after calibration on a known
candidate set (Figure~\ref{fig:supp_gallery_transfer}d).

\StageFullGalleryTable

\begin{table}[!htbp]
\centering
\small
\caption{\textbf{Small-batch cross-subject retrieval on THINGS-EEG2.}
Top-1 accuracy (\%) with a 200-image gallery.}
\label{tab:supp_smallbatch}
\begin{tabular*}{\columnwidth}{@{\extracolsep{\fill}}lrr@{}}
\toprule
Query batch size & SCORE & STAGE\\
\midrule
10 & \textbf{31.37} & 28.68\\
25 & \textbf{38.03} & 36.87\\
50 & 41.22 & \textbf{41.67}\\
100 & 44.48 & \textbf{48.82}\\
\bottomrule
\end{tabular*}

\end{table}

\subsection{Repeated Identities and Candidate Capacity}
\label{sec:supp-repeated}

To test the relation between candidate capacity and query composition,
we hold both query count and gallery size at 200 and retain either
150 or 100 distinct query identities. We select identities without
replacement, retain one row for each selected identity, and draw the
remaining rows with replacement using inverse-rank sampling weights.
Two fixed resampling seeds (1701 and 1702) are used for each of the
thirty EEG2 cases. These are synthetic repetitions of frozen embeddings;
they do not constitute independent EEG recordings.
Each method is fitted on the resulting batch, and the complete
200-candidate gallery is retained.

\begin{table}[!htbp]
\centering
\small
\caption{\textbf{Retrieval with repeated identities at fixed batch and gallery sizes.}
NICE EEG2 Top-1/Top-5 (\%). Each row averages two resamplings and three
source seeds within each of ten subjects.
STAGE includes initialization, adaptation, and capacity ranking.}
\label{tab:supp_repeated}
\begin{tabular*}{\columnwidth}{@{\extracolsep{\fill}}lrr@{}}
\toprule
Identities & SCORE & STAGE\\
\midrule
150 & 46.68/79.89 & \textbf{51.79}/\textbf{82.54}\\
100 & \textbf{42.20}/\textbf{75.83} & 40.78/71.20\\
\bottomrule
\end{tabular*}

\end{table}

STAGE reaches 51.79\% and 40.78\% Top-1 with 150 and 100 identities,
respectively; SCORE reaches 46.68\% and 42.20\%.
In this square setting, capacity constraints enforce equal column mass,
whereas repeated queries induce nonuniform identity frequencies.
The comparison relates candidate capacity to query composition
(Table~\ref{tab:supp_repeated}).

\clearpage
\makeatletter
\global\@colroom=\@colht
\global\vsize=\@colht
\makeatother
\section{Coupling Across Alignment and Ranking}
\label{sec:supp-coupling}

The following comparisons examine how coupling affects feature adaptation,
the recovery of correspondences during initialization, and ranking at fixed
costs. All use the NICE encoder and the complete 200-query/200-candidate
batch. EEG2 uses twelve initialization rounds and forty adaptation updates;
MEG uses two rounds and eight updates; Alljoined uses five rounds and
thirteen updates. We average source seeds 0, 1 and 2 within each subject,
then average subjects. Paired 95\% intervals resample subjects 10,000 times
after this seed averaging.

\subsection{Global Competition During Feature Adaptation}

\paragraph{Separating adaptation from final ranking.}
To determine whether global coupling improves the learned features, we
replace both the teacher and student couplings with row-normalized couplings,
\begin{equation}
T^{\mathrm{row}}_{\eta,ij}
=\frac{1}{N}\,
\frac{\exp(-C_{ij}/\eta)}{\sum_{k=1}^{M}\exp(-C_{ik}/\eta)} .
\label{eq:supp_row_plan}
\end{equation}
Here \(\eta=\varepsilon\) for the teacher and \(\eta=\tau\) for the
student. Each query retains the same total mass, but candidates have no
shared capacity constraint. Both arms use the same globally coupled
initialization, constrained feature map, forty updates, optimizer,
refresh interval, temperatures, and feature anchor. The row-only arm
minimizes the Fenchel--Young gap over the row-constrained feasible set;
its cost gradient remains the teacher--student coupling difference divided
by \(\tau\). Unlike the row cross-entropy comparison in
Section~\ref{sec:supp-adaptation}, this intervention also removes column
competition from the teacher.

\begin{table}[!htbp]
\centering
\small
\setlength{\tabcolsep}{2pt}
\caption{\textbf{Global competition during adaptation.}
EEG2 accuracy (\%) over ten subjects and three source seeds.
Teacher and student constraints change together; the final ranking
rule is identical within each row. Bold marks the better adaptation
arm for each metric.}
\label{tab:supp_current_adaptation}
\begin{tabular*}{\columnwidth}{@{\extracolsep{\fill}}lrrrr@{}}
\toprule
& \multicolumn{2}{c}{Row-only adaptation}
& \multicolumn{2}{c}{Global adaptation}\\
\cmidrule(lr){2-3}\cmidrule(l){4-5}
Final ranking & Top-1 & Top-5 & Top-1 & Top-5\\
\midrule
Direct & 49.10 & 77.10 & \textbf{58.63} & \textbf{85.18}\\
CSLS & 55.12 & 83.57 & \textbf{63.15} & \textbf{87.67}\\
GES & 66.25 & 89.48 & \textbf{68.25} & \textbf{90.47}\\
\bottomrule
\end{tabular*}

\end{table}

\paragraph{Feature gains persist under the same ranking rule.}
Global adaptation improves Top-1 over row-only adaptation by 9.53 points
with Direct ranking and 8.03 points with CSLS
(Table~\ref{tab:supp_current_adaptation}).
With the same GES ranking, the gain is 2.00 points
(95\% interval [0.93, 3.05]), with positive differences for eight of ten
subjects. Thus competition during feature refinement contributes
beyond the shared initialization and the final coupled ranking.

\subsection{Recovery of Initially Excluded Correspondences}

\begin{figure*}[!tbp]
\centering
\includegraphics[width=\textwidth]{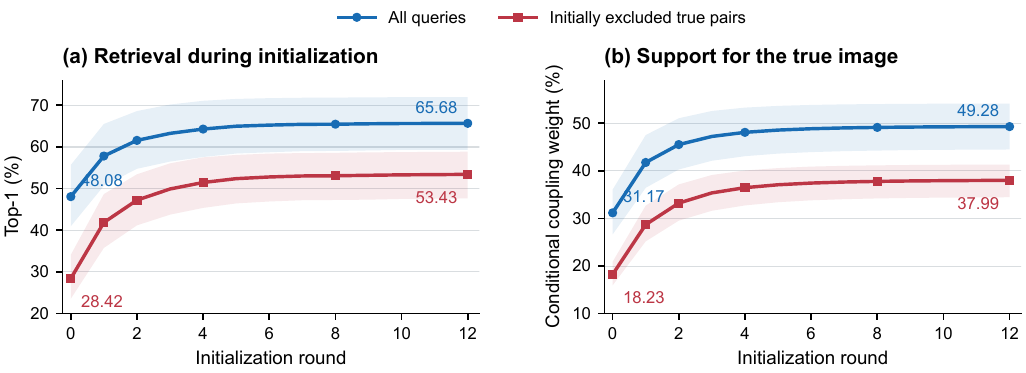}
\caption{\textbf{Correspondence recovery during alternating initialization.}
(a) GES Top-1 and (b) conditional coupling weight on the true image
through twelve initialization rounds, before feature adaptation.
The initially excluded group contains queries whose true pair is absent
from the fixed initial MNN landmark set. Curves average three source
seeds within each of ten EEG2 subjects; shaded regions are 95\%
subject-bootstrap intervals. Labels give the initial and final means.}
\label{fig:supp_init_recovery}
\end{figure*}

\paragraph{Following a fixed set of omitted true pairs.}
We record every round of the alternating coupling and Procrustes
initialization. Before these updates, we fix the SCORE-style landmark
set using cosine-CSLS with \(k=10\), mutual highest-scoring pairs, and
at most 160 landmarks ranked by matching margin. We then measure
retrieval for all queries and for queries whose true pair was absent
from this initial set. The latter group contains 139.27 queries per
batch on average; a query can have a selected landmark while its true
pair remains excluded. For true-image index \(j_i^\star\), the
conditional coupling weight is \(N T^{(k)}_{i,j_i^\star}\), since every
row has mass \(1/N\). True-pair identities define these measurements
after fitting.

\paragraph{Alternating updates restore support for omitted matches.}
Fig.~\ref{fig:supp_init_recovery} shows all-query Top-1 increasing
from 48.08\% to 65.68\%. Within the initially excluded group, Top-1
rises from 28.42\% to 53.43\%, conditional true-image weight rises
from 18.23\% to 37.99\%, and mean true-image rank improves from
6.73 to 3.45. Relative to round zero, the final round corrects 38.07
initially incorrect queries and introduces 3.07 errors per batch
within this group. The trajectory shows how correspondence support
can develop beyond the pairs admitted by the initial landmark rule;
the matched estimator comparisons in
Section~\ref{sec:supp-landmark} separately test the choice of
correspondence construction.

\subsection{Column Competition Across Datasets}

\paragraph{Comparing normalizations at fixed geometry.}
Table~\ref{tab:supp_current_normalizations} compares ranking rules on
the same fitted costs for all four MEG subjects and twenty Alljoined
subjects, each with three source seeds. Inverted softmax (IS) estimates
column normalizers from the query batch, at inverse temperature 20
or \(1/\varepsilon\).

On Alljoined, GES increases Top-1 over CSLS by 1.13 points
(95\% interval [0.51, 1.81]), with improvements for fifteen of twenty
subjects; it also exceeds matched-temperature IS by 0.68 points.
MEG shows a smaller GES--CSLS difference of 0.13 points, while the
two IS settings give the highest Top-1 and Top-5, respectively.
Candidate competition therefore has different effects across the
two embedding distributions.

\begin{table*}[!t]
\begin{minipage}[t]{\columnwidth}
\vspace{0pt}
\centering
\small
\setlength{\tabcolsep}{2pt}
\caption{\textbf{Ranking comparisons under dataset-specific adaptation budgets.}
Accuracy (\%) with frozen costs, using two initialization rounds/eight
updates for MEG and five rounds/thirteen updates for Alljoined.
Both datasets average three source seeds per subject. Bold marks
the best value for each dataset and metric.}
\label{tab:supp_current_normalizations}
\begin{tabular*}{\columnwidth}{@{\extracolsep{\fill}}lrrrr@{}}
\toprule
& \multicolumn{2}{c}{MEG} & \multicolumn{2}{c}{Alljoined}\\
\cmidrule(lr){2-3}\cmidrule(l){4-5}
Ranking & Top-1 & Top-5 & Top-1 & Top-5\\
\midrule
Direct & 5.17 & 17.13 & 10.28 & 30.03\\
CSLS & 6.13 & 19.83 & 11.76 & 32.63\\
IS ($20$) & 6.33 & \textbf{20.71} & 12.19 & 33.83\\
IS ($1/\varepsilon$) & \textbf{6.54} & 19.83 & 12.22 & 33.69\\
GES & 6.25 & 20.17 & \textbf{12.89} & \textbf{34.74}\\
\bottomrule
\end{tabular*}

\end{minipage}\hfill
\begin{minipage}[t]{\columnwidth}
\vspace{0pt}
\centering
\small
\setlength{\tabcolsep}{2pt}
\caption{\textbf{Entropy and column competition at fixed costs.}
Top-1 (\%) for Eq.~\eqref{eq:supp_e1control}, with the same subjects,
source seeds, and fitted costs as
Table~\ref{tab:supp_current_normalizations}.
Ratio zero imposes only row constraints; infinity enforces uniform
column mass. Bold marks the best column setting within each row.}
\label{tab:supp_current_grid}
\begin{tabular*}{\columnwidth}{@{\extracolsep{\fill}}llrrrr@{}}
\toprule
& & \multicolumn{4}{c}{Column penalty \(\xi/\eta\)}\\
\cmidrule(l){3-6}
Dataset & \(\eta\) & \(0\) & \(1\) & \(10\) & \(\infty\)\\
\midrule
MEG & 0.0175 & 5.17 & 5.46 & \textbf{6.04} & 6.00\\
MEG & 0.035 & 5.17 & 5.67 & 6.08 & \textbf{6.25}\\
MEG & 0.07 & 5.17 & 6.08 & 6.08 & \textbf{6.33}\\
\midrule
AJ & 0.0175 & 10.28 & 11.13 & 12.65 & \textbf{12.74}\\
AJ & 0.035 & 10.28 & 11.52 & 12.85 & \textbf{12.89}\\
AJ & 0.07 & 10.28 & 11.98 & \textbf{13.03} & 12.96\\
\bottomrule
\end{tabular*}

\end{minipage}
\end{table*}

\paragraph{Distinguishing entropy from column coordination.}
The factorial comparison in Table~\ref{tab:supp_current_grid} varies
entropy and column penalty independently. Row-only normalization
preserves Direct rankings at every temperature. Adding column
coordination raises Top-1 at all three tested entropy coefficients
on both datasets. At \(\eta=0.035\), exact column balance changes
MEG Top-1 from 5.17\% to 6.25\% and Alljoined from 10.28\% to
12.89\%. The comparison locates the ranking change in the
interaction between queries through column weights, while entropy
controls how this shared mass is distributed among pairs.

\clearpage
\begin{figure*}[p]
\begin{minipage}[t]{\textwidth}
\section{Extended Retrieval Cases with the NICE Encoder}
\label{sec:supp-retrieval-cases}

Figs.~\ref{fig:supp_retrieval_success}--\ref{fig:supp_retrieval_regression}
show recovered matches, residual errors, and introduced errors using
the same NICE source checkpoints as the numerical EEG2 results.
SCORE denotes reproduced coordinate recovery followed by cosine CSLS
on the same frozen features. \method{} ranks candidates after \(K=12\)
initialization rounds and forty adaptation steps.
Each example includes the
viewed image, both top-ten lists and the true image's rank among all 200
official THINGS-EEG2 candidates. All cases use backbone seed 0, with two cases
per subject. Green borders mark exact matches; the rightmost rank change
reads SCORE to \method{}.

\centering
\includegraphics[width=\textwidth]{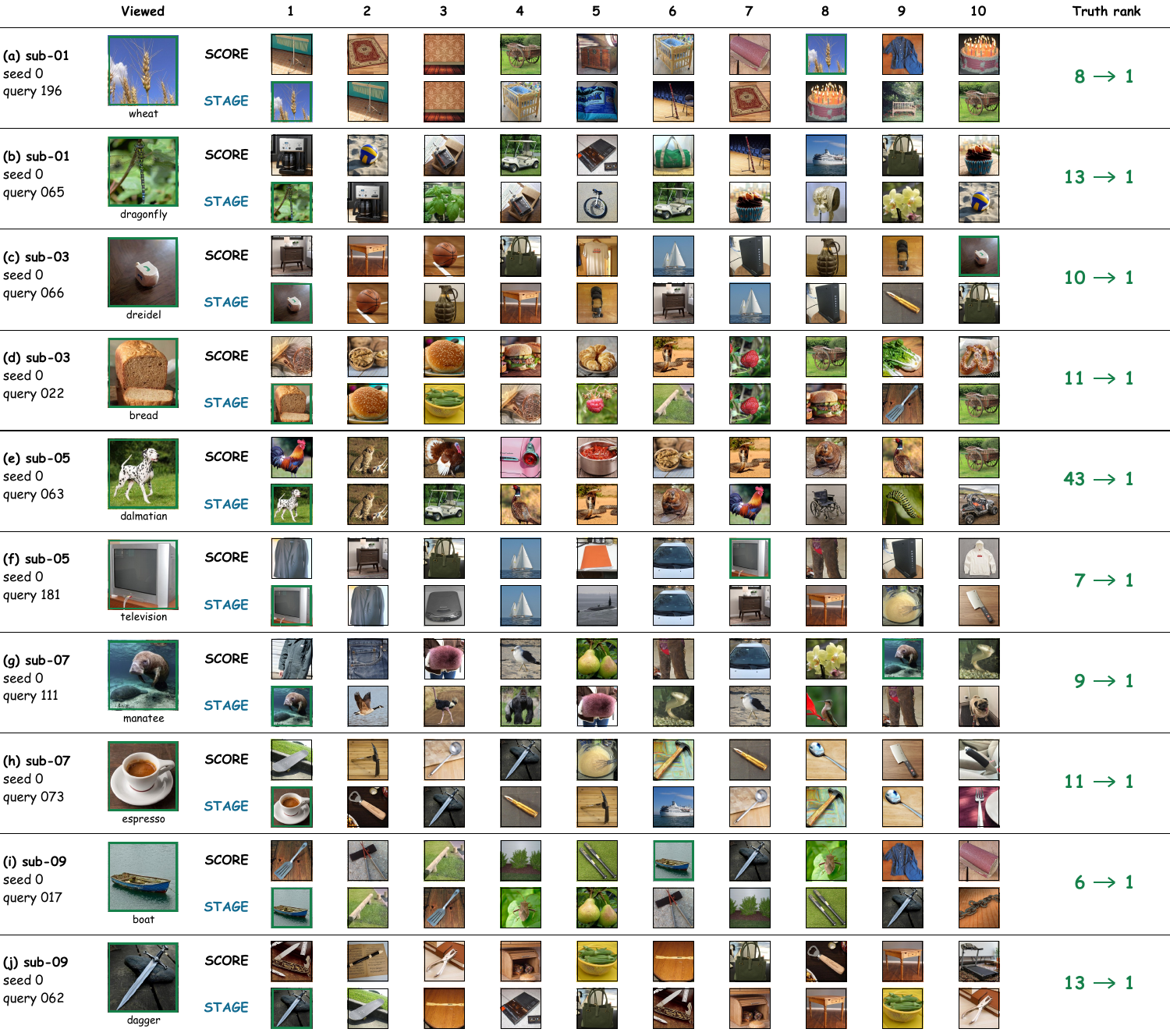}
\caption{\textbf{Recovered matches.}
Ten cases from subjects 1, 3, 5, 7, and 9, two per subject.
SCORE ranks the viewed image below fifth place and \method{} retrieves it
first. Cases are randomly sampled within each prescribed subject and outcome;
all thirty viewed images are distinct.}
\label{fig:supp_retrieval_success}
\end{minipage}
\end{figure*}

\begin{figure*}[p]
\begin{minipage}[t]{\textwidth}
\centering
\includegraphics[width=\textwidth]{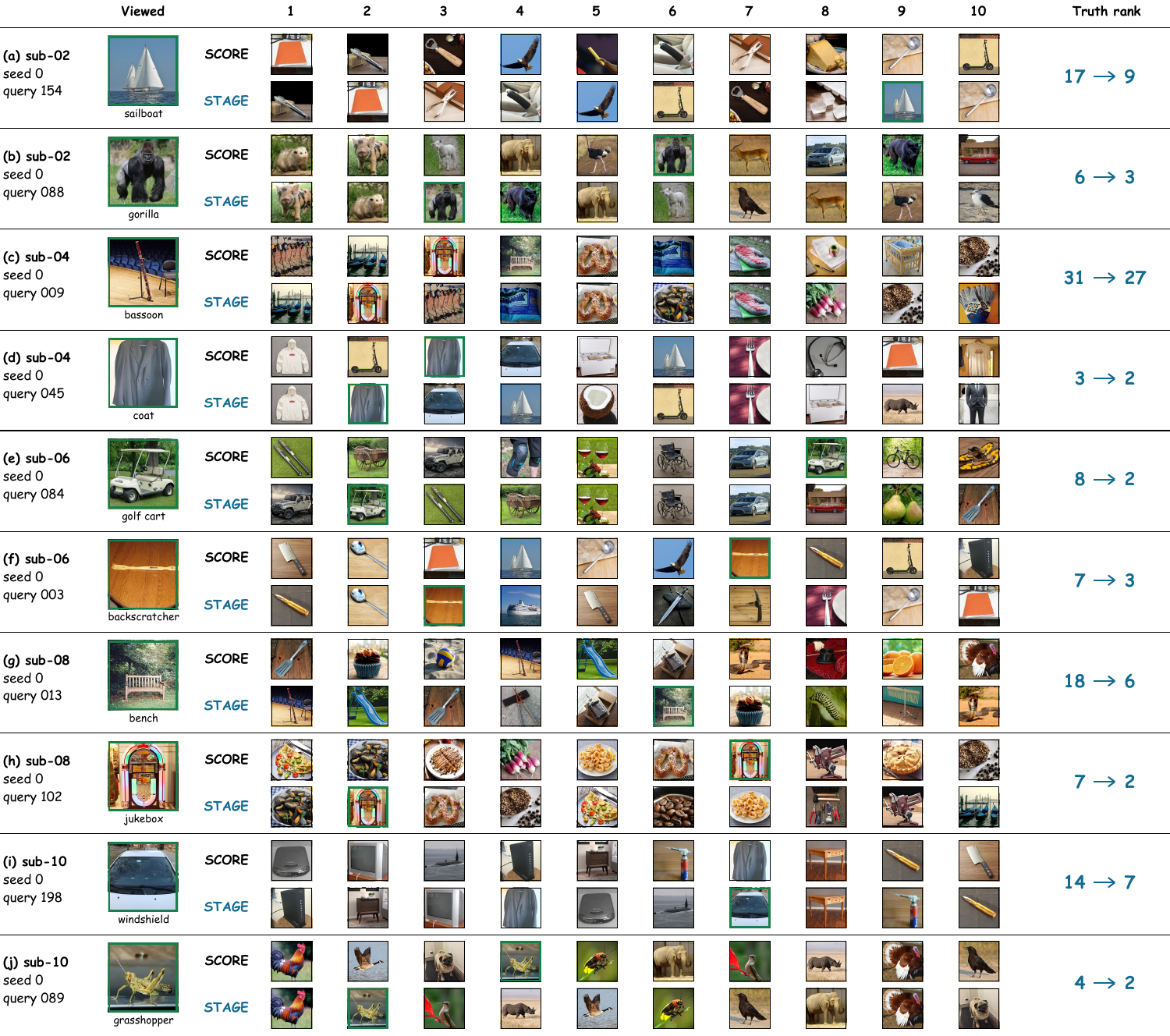}
\caption{\textbf{Residual errors despite rank improvement.}
Ten cases from subjects 2, 4, 6, 8, and 10, two per subject.
Both methods miss the viewed image at Top-1, although \method{} improves its
rank. The top-ten lists show both near misses and matches that remain below
tenth place. Labels and green borders
follow Fig.~\ref{fig:supp_retrieval_success}.}
\label{fig:supp_retrieval_residual}
\end{minipage}
\end{figure*}

\begin{figure*}[p]
\begin{minipage}[t]{\textwidth}
\centering
\includegraphics[width=\textwidth]{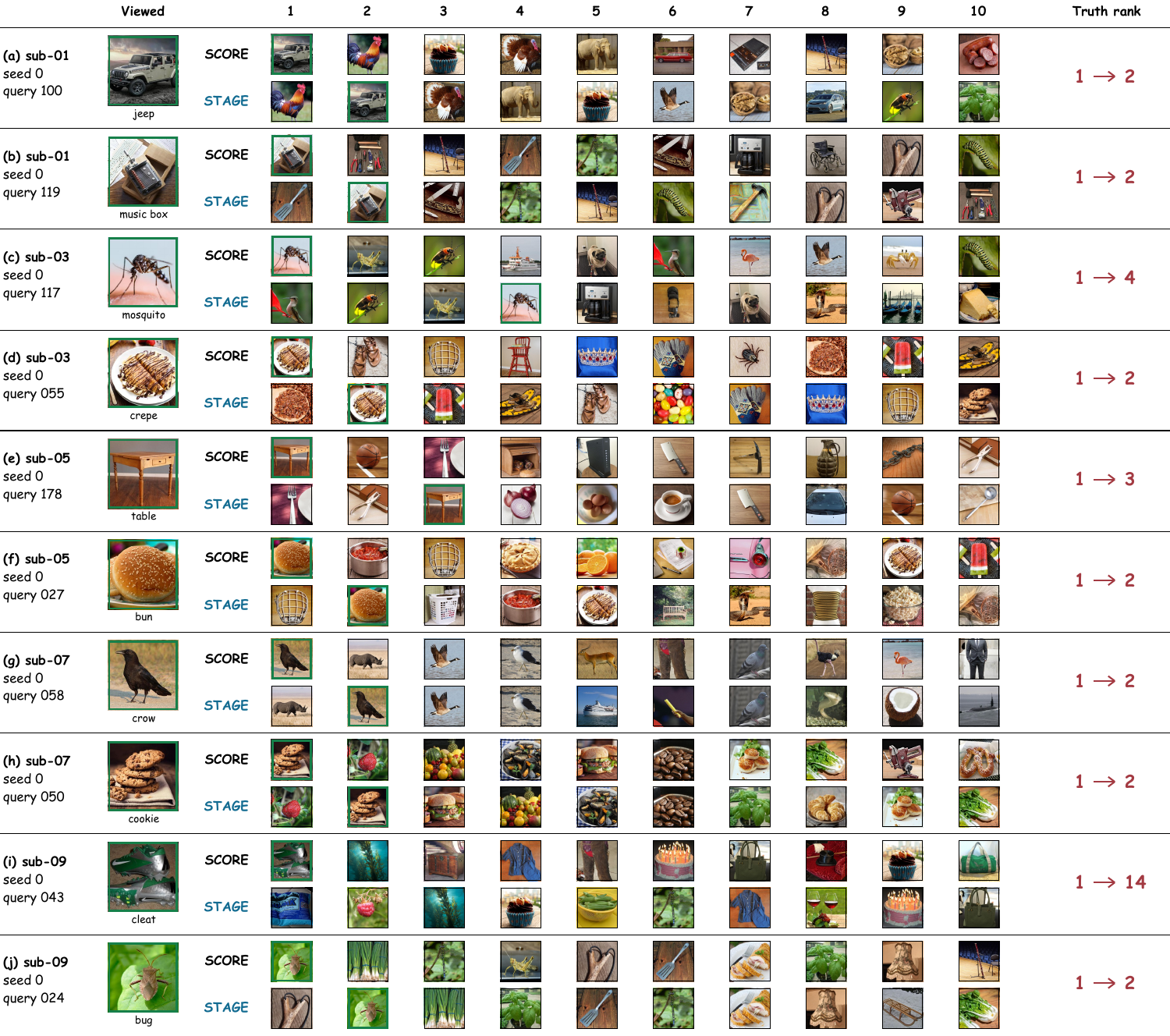}
\caption{\textbf{Errors introduced relative to SCORE.}
Ten cases from subjects 1, 3, 5, 7, and 9, two per subject.
SCORE retrieves the viewed image first; \method{} ranks it lower.
These panels include near misses and larger rank losses under the same protocol.}
\label{fig:supp_retrieval_regression}
\end{minipage}
\end{figure*}

\clearpage
\section{Additional Encoder and Source-Seed Replication}
\label{sec:supp-archived}

\paragraph{Additional frozen encoder.}
Table~\ref{tab:supp_archived_stages} applies the same alignment and
coupled ranking to SAMGA--TSConv features. All comparisons in this section
use twelve initialization rounds and forty feature updates across datasets,
with all official subjects and candidates. The final Direct, CSLS and
coupling entries share the same learned features and costs.
On EEG2, final coupling gives 66.30\%/88.95\%, compared with
56.98\%/83.20\% for Direct. This tests the method on a second frozen
brain encoder.

\StageEncoderTable

\paragraph{Source-model seed replication.}
Table~\ref{tab:supp_uniform_seeds} compares three independently trained
source seeds for NICE and TSConv under this shared budget, evaluating
ten EEG2, four MEG and twenty Alljoined subjects per seed.
The fixed budget isolates variation due to source training.

\StageUniformSeedTable

\FloatBarrier
\clearpage
\section{Alignment and Ranking on Matched Features}
\label{sec:supp-matched-system}

\subsection{Coordinate Recovery with a Shared Ranking Rule}

To separate geometric alignment from final ranking,
Table~\ref{tab:supp_shared_ranking} applies standard Sinkhorn normalization
(SN) to SCORE-recovered features, STAGE initialization, and complete
STAGE. The three radial-cost comparisons share the representation,
gallery-distance normalization, and entropy coefficient
\(\varepsilon=0.035\); their cost matrices differ with the fitted geometry.
A separate SCORE comparison retains its native cosine cost.
All comparisons use the same frozen NICE features, complete
200-query/200-candidate batches, and source seeds 0, 1 and 2.
The STAGE initialization/adaptation budgets are 12/40 for EEG2,
2/8 for MEG, and 5/13 for Alljoined.

\begin{table}[!htbp]
\centering
\small
\caption{\textbf{Geometric alignment under a shared ranking rule.}
Top-1 accuracy (\%) on identical frozen NICE features. SCORE denotes
its coordinate recovery. Below the divider,
the cost construction and standard SN ranking are shared across the three
geometries. Above it, SCORE uses its native cosine-based scores.
All SN comparisons use uniform row and column mass \(1/200\) and
\(\varepsilon=0.035\). Source seeds are averaged within each subject before
averaging subjects. Bold marks the best value in each dataset, including
ties.}
\label{tab:supp_shared_ranking}
\begin{tabular*}{\columnwidth}{@{\extracolsep{\fill}}lrrr@{}}
\toprule
Method & EEG2 & MEG & Alljoined\\
\midrule
SCORE + CSLS~\cite{cui2026score} & 49.73 & 5.00 & 10.18\\
SCORE + SN (cosine) & 53.25 & \textbf{6.25} & 10.39\\
\midrule
SCORE + SN (radial) & 58.20 & 5.63 & 11.78\\
STAGE init. + SN & 65.68 & 6.21 & 12.67\\
Complete STAGE + SN & \textbf{68.25} & \textbf{6.25}
& \textbf{12.89}\\
\bottomrule
\end{tabular*}

\end{table}

On EEG2, replacing SCORE recovery with STAGE initialization under the
shared radial-cost construction and SN ranking increases Top-1 by
7.48 percentage points (95\% interval [5.10, 9.93]).
Feature adaptation adds a further 2.57 points ([1.62, 3.63]).
These comparisons support contributions from both geometric initialization
and feature refinement beyond the shared final ranking.

\subsection{Subject-Paired System Comparisons}

Table~\ref{tab:supp_system_paired} reports paired Top-1 differences for
the same-feature system comparison in the main paper.
For each subject, we first average the three source seeds, then subtract
SCORE recovery with CSLS from complete STAGE.
Descriptive 95\% intervals resample these subject-level paired differences
10,000 times with bootstrap seed 20261114.

\begin{table}[!htbp]
\centering
\small
\caption{\textbf{Subject-paired gains over SCORE recovery with CSLS.}
Differences are in percentage points on matched frozen NICE features.
Subjects are the resampling units after averaging source seeds.
The independent subject counts are 10, 4, and 20.}
\label{tab:supp_system_paired}
\begin{tabular*}{\columnwidth}{@{\extracolsep{\fill}}lrrr@{}}
\toprule
Dataset & Subjects & \(\Delta\)Top-1 & 95\% interval\\
\midrule
EEG2 & 10 & \(+18.52\) & [16.32, 20.68]\\
MEG & 4 & \(+1.25\) & [-0.13, 2.58]\\
Alljoined & 20 & \(+2.72\) & [1.73, 3.73]\\
\bottomrule
\end{tabular*}

\end{table}

Subject-level Top-1 differences are positive for all ten EEG2 subjects,
three of four MEG subjects, and seventeen of twenty Alljoined subjects.

\clearpage
\subsection{Budget-Matched Recovery and Adaptation}
\label{sec:supp-budget-matched}

We compare soft correspondence recovery with two iterative MNN
initializers under the same transformation estimator, update budgets, and
final ranking (Table~\ref{tab:supp_matched_factorial}).
All variants use identical frozen NICE features, moment calibration,
regularized Procrustes estimation, radial-cost construction, and standard
SN ranking with \(\varepsilon=0.035\).
Each initialization update refits the transform from the original calibrated
queries to the fixed gallery.
Radial MNN selects mutual neighbors using the current radial cost;
CSLS MNN uses cosine-based CSLS scores.
Both retain at most 160 landmarks by row margin and preserve the previous
transform when fewer than 12 pairs are available.
These MNN variants repeat correspondence selection at every round.
The CSLS MNN comparator therefore evaluates iterative landmark recovery,
rather than the native SCORE pipeline.

To isolate correspondence weighting, every query has a unit multiplier
in this comparison, replacing the confidence reweighting used in
Table~\ref{tab:supp_shared_ranking}.
Selected MNN pairs receive equal weights; soft recovery uses the full
balanced coupling.
The initialization/adaptation budgets are 12/40, 2/8, and 5/13 for
EEG2, MEG, and Alljoined, respectively.
After each initializer, we compare no adaptation, row-normalized
teacher and student couplings, and globally balanced teacher and student
couplings.
The feature map, optimizer, temperatures, and update budget are identical
between the two adaptation variants.
All 200 queries and 200 candidates are retained for every subject and
source seed.

\begin{table}[!htbp]
\centering
\small
\caption{\textbf{Correspondence recovery and adaptation with matched budgets.}
Top-1 accuracy (\%) on frozen NICE features with uniform query
multipliers and shared standard SN ranking.
None, Row, and Global denote no adaptation, row-normalized
teacher/student adaptation, and globally balanced teacher/student
adaptation.
The three source seeds are averaged within each subject before averaging
10 EEG2, 4 MEG, or 20 Alljoined subjects.
Bold marks the best initializer in each adaptation column within a
dataset.}
\label{tab:supp_matched_factorial}
\begin{tabular*}{\columnwidth}{@{\extracolsep{\fill}}llrrr@{}}
\toprule
Dataset & Initialization & None & Row & Global\\
\midrule
\multirow{3}{*}{EEG2}
& Radial MNN & 58.40 & 57.22 & 63.07\\
& CSLS MNN & 62.93 & 62.90 & 66.52\\
& Soft coupling & \textbf{65.20} & \textbf{66.17} & \textbf{68.02}\\
\midrule
\multirow{3}{*}{MEG}
& Radial MNN & 5.79 & 5.58 & 5.75\\
& CSLS MNN & 5.67 & 5.63 & 5.58\\
& Soft coupling & \textbf{6.00} & \textbf{6.25} & \textbf{6.13}\\
\midrule
\multirow{3}{*}{Alljoined}
& Radial MNN & 11.21 & 11.07 & 11.55\\
& CSLS MNN & 11.87 & 11.53 & 11.96\\
& Soft coupling & \textbf{12.67} & \textbf{12.52} & \textbf{12.82}\\
\bottomrule
\end{tabular*}
\end{table}

On EEG2, soft recovery exceeds iterative CSLS MNN by
2.27 percentage points without adaptation (95\% interval [1.17, 3.33])
and by 1.50 points with the same global adaptation ([0.40, 2.65]).
Starting from soft recovery, global adaptation improves Top-1 by
2.82 points over no adaptation ([1.95, 3.72]) and by
1.85 points over row-normalized adaptation ([0.93, 2.68]).
On Alljoined, soft recovery followed by global adaptation exceeds the
corresponding CSLS MNN variant by 0.86 points ([0.28, 1.48]).
These paired intervals use 10,000 subject-bootstrap resamples after
averaging seeds, with bootstrap seed 20261115.
With the final ranking and update budgets held fixed, the EEG2
comparisons support gains from both soft correspondence recovery and
global feature refinement.

\subsection{Iterative Recovery of Initially Excluded Correspondences}
\label{sec:supp-excluded-recovery}

\begin{figure}[!htbp]
\centering
\includegraphics[width=\columnwidth]{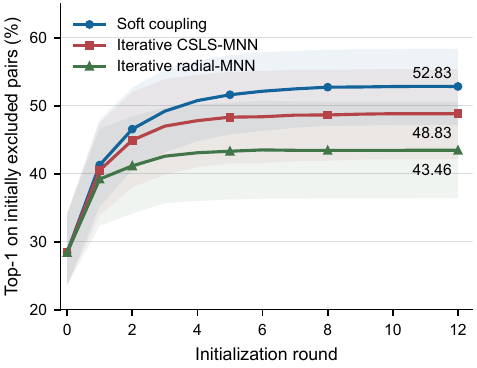}
\caption{\textbf{Recovery of initially excluded correspondences.}
Top-1 accuracy on EEG2 queries whose true pairs are excluded by the
common initial CSLS MNN selection.
Each curve uses the initialization settings in
Table~\ref{tab:supp_matched_factorial}, before feature adaptation,
with shared standard SN ranking at every round.
The query group remains fixed across initializers and rounds.
Lines average seeds within subjects and then subjects; bands are
pointwise 95\% subject-bootstrap intervals over ten subjects.}
\label{fig:supp_matched_recovery}
\end{figure}

Figure~\ref{fig:supp_matched_recovery} tracks EEG2 queries whose true
pairs are absent from the shared initial CSLS MNN landmark set.
This fixed group contains an average of 139.27 queries per
200-query case.
Ground-truth identities are used only for grouping and evaluation after
fitting.
All initializers start from the same calibrated features, with
28.42\% Top-1 on this group.
After 12 initialization rounds, soft recovery reaches 52.83\%,
compared with 48.83\% for iterative CSLS MNN.
Relative to round zero, soft recovery newly corrects 37.17 queries
per case, compared with 32.13 for CSLS MNN; the corresponding
numbers of newly incorrect predictions are 3.03 and 3.87.
Both approaches recover initially excluded matches as alignment improves;
soft recovery corrects more queries while introducing fewer errors.

\FloatBarrier
\subsection{Directional Refinement During Adaptation}
\label{sec:supp-directional-refinement}

To distinguish directional refinement from shared-scale changes, we refit
restricted versions of the adaptation transform on all ten EEG2 subjects
and three source seeds. Every variant starts from the same
coupling-weighted initialization and fixed subspace basis, with forty
updates and identical optimizer settings, teacher refreshes, temperatures,
and feature anchor. Disabled rotation, scale, and translation blocks remain
at \(R_\theta=I_r\), \(s_\theta=1\), and \(t_\theta=0\), respectively,
throughout optimization. The comparison isolates updates after
initialization; the initialization transform is shared by all variants.
Final GES ranking uses \(\varepsilon=0.035\) for the complete
200-query/200-candidate batch.

\begin{table}[!htbp]
\centering
\small
\caption{\textbf{Directional and scale updates after initialization.}
EEG2 Top-1 accuracy (\%) with identical initialization and final GES
ranking. Each row specifies the parameters learned during adaptation;
the remaining blocks stay fixed at their identity values.
Source seeds are averaged within each subject before averaging subjects.
Bold marks the highest mean.}
\label{tab:supp_directional_refinement}
\begin{tabular*}{\columnwidth}{@{\extracolsep{\fill}}lr@{}}
\toprule
Adaptation & Top-1\\
\midrule
None (initialization) & 65.68\\
Full (rotation, scale, translation) & 68.25\\
Rotation & 68.30\\
Scale & 66.00\\
Rotation + scale & \textbf{68.38}\\
Rotation + translation (fixed scale) & 68.25\\
\bottomrule
\end{tabular*}
\end{table}

Table~\ref{tab:supp_directional_refinement} shows that rotation updates
at fixed scale improve Top-1 over initialization by 2.62 percentage
points (95\% interval [1.28, 3.82]), accounting for most of the gain
obtained by full adaptation (+2.57 points).
Scale-only updates add 0.32 points ([0.12, 0.53]).
Adding scale to rotation changes Top-1 by 0.08 points
([-0.33, 0.52]). These paired intervals use 10,000 subject-bootstrap
resamples after seed averaging, with bootstrap seed 20261116.
The comparison supports directional refinement as the principal source
of the EEG2 adaptation gain under shared GES ranking.

\FloatBarrier
\subsection{Updates from a Shared Initialization}
\label{sec:supp-shared-start-updates}

To compare alignment updates and teacher correspondences,
we start every variant from the same coupling-weighted initialization
and represented query and gallery features.
Table~\ref{tab:supp_shared_start_updates} compares WP-style projected
updates~\cite{grave2019wasserstein}, hard alternating Procrustes updates,
entropic OT--Procrustes updates, and teacher-guided feature adaptation.
The three rigid controls initialize a full-space orthogonal map at identity.
WP-style updates alternate entropic transport and projected-gradient
steps, with entropy coefficient 0.05 and learning rate 500 on full
200-query/200-candidate batches.
Hard alternating updates initialize correspondences by a maximum-score
assignment and alternate exact Procrustes fitting with assignment updates.
Entropic OT--Procrustes alternates balanced transport at
\(\varepsilon=0.035\) with rigid Procrustes fitting, using uniform query
weights and no ridge regularization.
Each rigid control performs 44, 9, or 15 updates on EEG2, MEG, or Alljoined,
matching the number of teacher and student transport solves in STAGE's
40, 8, or 13 adaptation steps; the hard control also constructs its
initial assignment.
For the hard-teacher comparison, only the periodically refreshed teacher
is replaced by a minimum-cost assignment, with mass \(1/200\) on each
selected pair. The student coupling, feature map, optimizer,
temperatures, refresh schedule, anchor, and adaptation budget remain
identical to STAGE.
Final GES ranking and its cost construction are shared across variants.

\begin{table}[!htbp]
\centering
\small
\caption{\textbf{Post-initialization updates from shared features.}
Top-1 accuracy (\%) with identical STAGE initialization and final GES
ranking. The rigid controls vary the subsequent update rule;
the teacher comparison varies only the teacher's soft or hard
correspondences while learning the same constrained feature transform.
All 200 queries and candidates are retained.
Three source seeds are averaged within each subject before averaging
10 EEG2, 4 MEG, or 20 Alljoined subjects.
Bold marks the highest mean in each dataset.}
\label{tab:supp_shared_start_updates}
\begin{tabular*}{\columnwidth}{@{\extracolsep{\fill}}lrrr@{}}
\toprule
Subsequent update & EEG2 & MEG & Alljoined\\
\midrule
No further update & 65.68 & 6.21 & 12.67\\
WP-style projected updates & 50.00 & 5.92 & 12.53\\
Hard alternating Procrustes & 65.00 & 6.04 & 12.12\\
Entropic OT--Procrustes & 68.08 & 5.71 & 10.22\\
\midrule
Hard-teacher adaptation & 67.02 & 5.88 & 12.39\\
STAGE (soft teacher) & \textbf{68.25} & \textbf{6.25} & \textbf{12.89}\\
\bottomrule
\end{tabular*}
\end{table}

On EEG2, the soft teacher improves Top-1 over the hard teacher by
1.23 percentage points (95\% interval [0.12, 2.32]); the corresponding
Alljoined gain is 0.50 points ([0.13, 0.93]).
With the student and optimization settings held fixed, these comparisons
support retaining uncertain correspondences in the teacher targets.
On EEG2, STAGE also exceeds WP-style projected updates by 18.25 points
([16.07, 20.25]) and hard alternating updates by 3.25 points
([1.98, 4.48]), while improving over initialization by 2.57 points.
Continued entropic OT--Procrustes reaches 68.08\%, close to STAGE's
68.25\%, and reaches 5.71\% and 10.22\% on MEG and Alljoined.
Paired intervals use 10,000 subject-bootstrap resamples after averaging
source seeds, with bootstrap seed 20261116.

\FloatBarrier
\subsection{Retrieval Accuracy and Selected-Pair Cost}
\label{sec:supp-selected-pair-cost}

Table~\ref{tab:supp_selected_pair_cost} compares accuracy and selected-pair
costs at fixed geometry. Hard assignment improves over row-wise ranking;
soft-coupling ranking retains alternatives and reaches 68.25\% Top-1.
Ground-truth pairs cost more than row-wise selections, so minimizing
embedding cost alone does not identify true matches.

\begin{table}[t]
\centering
\small
\caption{\textbf{Retrieval accuracy and selected-pair cost.}
Top-1/Top-5 (\%) over 30 cases at fixed costs. Mean selected-pair cost is
normalized by the median squared distance within the gallery. Ground-truth
and max-cost matchings are reference controls. Dashes denote inapplicable
Top-5 for rules that return a single match without a candidate ranking.}
\label{tab:supp_selected_pair_cost}
\begin{tabular*}{\columnwidth}{@{\extracolsep{\fill}}lrrr@{}}
\toprule
Ranking rule & Top-1 & Top-5 & Selected-pair cost\\
\midrule
Ground-truth matching & 100.00 & -- & 0.5165\\
Max-cost matching & 0.00 & -- & 1.8559\\
Row-wise & 58.63 & 85.18 & 0.4681\\
Hard assignment & 67.32 & -- & 0.4915\\
Coupling (ours) & \textbf{68.25} & \textbf{90.47} & 0.4854\\
\bottomrule
\end{tabular*}

\end{table}

\begin{figure}[t]
\centering
\includegraphics[width=\columnwidth]{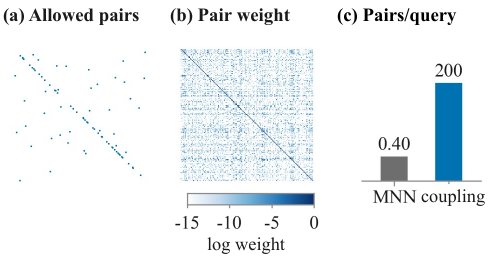}
\caption{\textbf{Correspondence support and coupling weights.} (a) Pairs the MNN rule admits,
0.40 per query and 21.5\% of them correct; (b) coupling weight per
cell; (c) available pairs per query, 0.40 for MNN and 200 for the dense
coupling. One
held-out fold (THINGS-EEG2, subject 01, seed 0, $200\times200$).}
\label{fig:supp_pair_support}
\end{figure}

\FloatBarrier
\subsection{Adaptation Objectives with a Shared Feature Map}
\label{sec:supp-direct-ot-objective}

To isolate the adaptation objective, we compare teacher-guided
refinement with directly minimizing the entropic transport value.
All variants start from identical saved initialized features and
use the same fixed subspace, constrained feature map, feature anchor,
Adam settings, clipping, update budgets, and final GES ranking.
The direct objectives are
\(\mathrm{val}_{\eta}(C_\theta)/\tau+\lambda L_{\mathrm{anchor}}\),
with \(\eta=\varepsilon=0.035\) or \(\eta=\tau=0.07\);
their cost gradients are \(T_\eta(C_\theta)/\tau\).
The shared divisor \(\tau\) keeps the anchor in the same objective
units as the teacher gap in the main method.
Both temperatures and all optimization settings are fixed across
the complete subject and source-seed populations.
Each run uses its prescribed final update.

\begin{table}[!htbp]
\centering
\small
\caption{\textbf{Adaptation objectives under matched geometry.}
Top-1 accuracy (\%) on identical initialized features, constrained
transforms and final rankings. Budgets are 40, 8 and 13 updates for
EEG2, MEG and Alljoined. Three seeds are averaged within each subject,
then across 10, 4 and 20 subjects. Bold marks the best mean.}
\label{tab:supp_direct_ot_objective}
\begin{tabular*}{\columnwidth}{@{\extracolsep{\fill}}lrrr@{}}
\toprule
Adaptation objective & EEG2 & MEG & Alljoined\\
\midrule
Direct OT (\(\eta=\varepsilon\)) & 61.48 & 5.29 & 10.93\\
Direct OT (\(\eta=\tau\)) & 52.53 & 5.58 & 10.43\\
STAGE (soft teacher) & \textbf{68.25} & \textbf{6.25} & \textbf{12.89}\\
\bottomrule
\end{tabular*}
\end{table}

Relative to direct OT at \(\eta=\varepsilon\), the teacher-guided
objective improves EEG2 by 6.77 percentage points (95\% paired interval
[3.70, 9.90]) and Alljoined by 1.96 points ([1.11, 2.83]).
The MEG difference is 0.96 points ([-0.17, 2.08]).
At \(\eta=\tau\), the corresponding gains are 15.72 points
([13.03, 18.50]), 2.46 points ([1.51, 3.50]), and 0.67 points
([-0.38, 1.63]), respectively.
Intervals use 10,000 subject-bootstrap resamples after seed averaging,
with bootstrap seed 20261116.
Under the shared optimization settings, these results support matching
soft correspondence targets during feature refinement beyond directly
reducing the transport value with the same feature map and anchor.

\FloatBarrier
\subsection{Computation of Post-Initialization Updates}
\label{sec:supp-update-timing}

Table~\ref{tab:supp_update_timing} compares computation after the shared
coupling-weighted initialization. All methods use the same initialized
features on one CPU node with four threads. We measure subject 01,
source seed 0 of each dataset, discard one warm-up, and report the
median of three repetitions with rotated execution order.
Fitting includes the prescribed updates and their transport solves;
the final GES solve is timed separately on each fitted cost matrix.
Initialization, checkpoint loading, encoder inference, trajectory
evaluation and serialization are excluded.
The rigid controls refit full 512-dimensional orthogonal maps, whereas
the teacher and direct-objective variants use the same rank-16
constrained feature map.

\begin{table*}[t]
\centering
\small
\caption{\textbf{Timing of post-initialization updates.}
Median seconds on a 200-query/200-candidate batch, four CPU threads,
subject 01 and source seed 0. Fit measures subsequent feature updates;
GES measures the final coupling solve. The ordinary OT--Procrustes
control uses uniform query weights without ridge; the confidence/ridge
variant retains STAGE's query confidence and identity regularization.
Both use 44/9/15 transport--Procrustes updates on EEG2/MEG/Alljoined.
Teacher and direct-objective variants use 40/8/13 feature updates.
Bold marks the lowest time in each column.}
\label{tab:supp_update_timing}
\begin{tabular*}{\textwidth}{@{\extracolsep{\fill}}lrrrrrr@{}}
\toprule
& \multicolumn{2}{c}{EEG2}
& \multicolumn{2}{c}{MEG}
& \multicolumn{2}{c}{Alljoined}\\
\cmidrule(lr){2-3}\cmidrule(lr){4-5}\cmidrule(l){6-7}
Subsequent update & Fit & GES & Fit & GES & Fit & GES\\
\midrule
Direct OT (\(\eta=\varepsilon\)) & 28.90 & 1.01 & 2.03 & 0.33 & 1.58 & \textbf{0.24}\\
Direct OT (\(\eta=\tau\)) & \textbf{2.58} & \textbf{0.67} & \textbf{0.31} & \textbf{0.18} & \textbf{0.71} & 0.25\\
Hard teacher & 3.37 & 4.24 & 0.32 & 0.35 & 0.75 & 0.25\\
OT--Procrustes & 425.54 & 10.11 & 46.61 & 9.21 & 60.56 & 9.69\\
OT--Procrustes + confidence/ridge & 39.22 & 4.51 & 2.89 & 0.84 & 2.12 & 0.26\\
STAGE (soft teacher) & 6.91 & 6.47 & 0.44 & 0.44 & 0.88 & 0.26\\
\bottomrule
\end{tabular*}
\end{table*}

Teacher-guided fitting takes 6.91/0.44/0.88\,s on EEG2/MEG/Alljoined,
compared with 39.22/2.89/2.12\,s for confidence-weighted, regularized
OT--Procrustes continuation.
Direct OT at \(\eta=\tau\) is faster, while teacher-guided refinement
achieves higher Top-1 accuracy under the shared optimization settings
(Table~\ref{tab:supp_direct_ot_objective}).
Together, the objective and timing comparisons characterize the
accuracy--computation trade-off of the update rules.

\FloatBarrier

\end{document}